\documentclass[10pt,final,journal]{IEEEtran}

\usepackage{amsmath}
\usepackage{amsfonts}
\usepackage{arydshln}
\usepackage{cite}
\usepackage[]{graphicx}
\usepackage{url}
\usepackage{xcolor}

\ifCLASSOPTIONcompsoc
\usepackage[caption=false,font=normalsize,labelfont=sf,textfont=sf]{subfig}
\else
\usepackage[caption=false,font=footnotesize]{subfig}
\fi

\newcommand\blfootnote[1]{%
  \begingroup
  \renewcommand\thefootnote{}\footnote{#1}%
  \addtocounter{footnote}{-1}%
  \endgroup
}

\newcommand{\achilles}{\textbf{achilles}}
\newcommand{\cian}{\textbf{cian}}
\newcommand{\hadi}{\textbf{hadi}}
\newcommand{\ipmibern}{\textbf{ipmi-bern}}
\newcommand{\ktwo}{\textbf{k2}}
\newcommand{\knight}{\textbf{knight}}
\newcommand{\lrde}{\textbf{lrde}}
\newcommand{\misp}{\textbf{misp}}
\newcommand{\neuroml}{\textbf{neuro.ml}}
\newcommand{\nicvicorob}{\textbf{nic-vicorob}}
\newcommand{\nihcidi}{\textbf{nih\_cidi}}
\newcommand{\nist}{\textbf{nist}}
\newcommand{\nlplogix}{\textbf{nlp\_logix}}
\newcommand{\scan}{\textbf{scan}}
\newcommand{\skkumedneuro}{\textbf{skkumedneuro}}
\newcommand{\sysumedia}{\textbf{sysu\_media}}
\newcommand{\textclass}{\textbf{text\_class}}
\newcommand{\tig}{\textbf{tig}}
\newcommand{\tignet}{\textbf{tignet}}
\newcommand{\upcdlmi}{\textbf{upc\_dlmi}}

\begin{document}

\title{Standardized Assessment of Automatic Segmentation of White Matter Hyperintensities and Results of the WMH Segmentation Challenge}

\author{
Hugo~J.~Kuijf, 
J.~Matthijs~Biesbroek,
Jeroen~de~Bresser,
Rutger~Heinen,
Simon~Andermatt,    
Mariana~Bento,      
Matt~Berseth,       
Mikhail~Belyaev,    
M.~Jorge~Cardoso,   
Adri\`{a}~Casamitjana, 
D.~Louis~Collins,    
Mahsa~Dadar,        
Achilleas~Georgiou, 
Mohsen~Ghafoorian,  
Dakai~Jin,          
April~Khademi,      
Jesse~Knight,       
Hongwei~Li,         
Xavier~Llad\'{o},   
Miguel~Luna,        
Qaiser~Mahmood,     
Richard~McKinley,   
Alireza~Mehrtash,   
S\'{e}bastien~Ourselin, 
Bo-yong~Park,       
Hyunjin~Park,       
Sang~Hyun~Park,     
Simon~Pezold,       
Elodie~Puybareau,   
Leticia~Rittner,    
Carole~H.~Sudre,    
Sergi~Valverde,     
Ver\'{o}nica~Vilaplana, 
Roland~Wiest,       
Yongchao~Xu,        
Ziyue~Xu,           
Guodong~Zeng,~\IEEEmembership{Student Member, IEEE,} 
Jianguo~Zhang,      
Guoyan~Zheng,~\IEEEmembership{Member, IEEE,}         
Christopher~Chen,
Wiesje~van~der~Flier,
Frederik~Barkhof,
Max~A.~Viergever,~\IEEEmembership{Fellow,~IEEE,}
and Geert~Jan~Biessels%
\thanks{%
H.J.~Kuijf and M.A.~Viergever are with the Image Sciences Institute, UMC Utrecht, Utrecht University, the Netherlands.
J.M.~Biesbroek, R.~Heinen, and G.J.~Biessels are with the Brain Center Rudolf Magnus, UMC Utrecht, Utrecht University, the Netherlands.
J.~de Bresser is with the Department of Radiology, UMC Utrecht, the Netherlands and with the Department of Radiology, LUMC, Leiden, the Netherlands.
S.~Andermatt and S.~Pezold are with the Department of Biomedical Engineering, University of Basel, Allschwil, Switzerland.
M.~Bento is with the Radiology and Clinical Neuroscience, Hotchkiss Brain Institute, University of Calgary, AB, Canada.
M.~Berseth is with NLP Logix.
M.~Belyaev is with the Skolkovo Institute of Science and Technology.
M.J.~Cardoso is with the School of Biomedical Engineering and Imaging Sciences, King's College London and the Centre for Medical Image Computing, University College London.
A.~Casamitjana and V.~Vilaplana are with the Signal Theory and Communications Department, Universitat Polit\`{e}cnica de Catalunya, BarcelonaTech, Barcelona, Spain.
D.L.~Collins and M.~Dadar are with the McGill University, Canada.
A.~Georgiou is with the Computational Statistics and Machine Learning MSc, University College London.
M.~Ghafoorian is with TomTom, Amsterdam, the Netherlands.
D.~Jin and Z.~Xu are with the Department of Radiology and Imaging Science, National Institutes of Health, USA.
A.~Khademi is with the Ryerson University, Canada.
J.~Knight is with the University of Guelph, Canada.
H.~Li is with the Sun Yat-sen University, University of Dundee, and Technical University of Munich.
X.~Llad\'{o} and S.~Valverde are with the Research institute of Computer Vision and Robotics, University of Girona, Spain.
M.~Luna and S.H.~Park are with the Department of Robotics Engineering, Daegu Gyeongbuk Institute of Science and Technology, Daegu, South Korea.
Q.~Mahmood is with the Pakistan Institute of Nuclear Science and Technology.
R.~McKinley and R.~Wiest are with the Support Center for Advanced Neuroimaging, Institute for Diagnostic and Interventional Neuroradiology, Inselspital, University of Bern, Switzerland.
A.~Mehrtash is with the Electrical and Computer Engineering Department, University of British Columbia, Vancouver, BC and the Department of Radiology, Brigham and Women's Hospital, Harvard Medical School, Boston, MA.
S.~Ourselin is with the is with the School of Biomedical Engineering and Imaging Sciences, King's College London.
B.~Park is with the Department of Electronic, Electrical and Computer Engineering, Sungkyunkwan University and the Center for Neuroscience Imaging Research, Institute for Basic Science (IBS), Suwon, Korea.
H.~Park is with the Center for Neuroscience Imaging Research, Institute for Basic Science (IBS) and the School of Electronic and Electrical Engineering, Sungkyunkwan University, Suwon, Korea.
E.~Puybareau is with the EPITA Research and Development Laboratory (LRDE), France.
L.~Rittner is with the School of Electrical and Computer Engineering, University of Campinas, SP, Brazil.
C.H.~Sudre is with the School of Biomedical Engineering and Imaging Sciences, King's College London; and the Centre for Medical Image Computing and the Dementia Research Centre, Institute of Neurology, University College London..
Yongchao~Xu is with the EPITA Research and Development Laboratory (LRDE) and the LTCI, T\'{e}l\'{e}com ParisTech, Universit\'{e} Paris-Saclay, France, and the Huazhong University of Science and Technology, China.
G.~Zeng and G.~Zheng are with the Institute for Surgical Technology \& Biomechanics, University of Bern, Bern, Switzerland.
J.~Zhang is with the University of Dundee.
C.~Chen is with the Memory Aging and Cognition Center, NUHS, Singapore.
W.~van der Flier is with the Alzheimer Center, VU Amsterdam, the Netherlands.
F.~Barkhof is with the Department of Radiology \& Nuclear Medicine, VU University Medical Center, Amsterdam, the Netherlands and the UCL institutes of Neurology and Healthcare Engineering, London, UK.
}}

\maketitle

\begin{abstract}
Quantification of cerebral white matter hyperintensities (WMH) of presumed vascular origin is of key importance in many neurological research studies. Currently, measurements are often still obtained from manual segmentations on brain MR images, which is a laborious procedure. Automatic WMH segmentation methods exist, but a standardized comparison of the performance of such methods is lacking. We organized a scientific challenge, in which developers could evaluate their method on a standardized multi-center/-scanner image dataset, giving an objective comparison: the WMH Segmentation Challenge (\url{https://wmh.isi.uu.nl/}).

Sixty T1+FLAIR images from three MR scanners were released with manual WMH segmentations for training. A test set of 110 images from five MR scanners was used for evaluation. Segmentation methods had to be containerized and submitted to the challenge organizers. Five evaluation metrics were used to rank the methods: (1) Dice similarity coefficient, (2) modified Hausdorff distance (95th percentile), (3) absolute log-transformed volume difference, (4) sensitivity for detecting individual lesions, and (5) F1-score for individual lesions. Additionally, methods were ranked on their inter-scanner robustness.

Twenty participants submitted their method for evaluation. This paper provides a detailed analysis of the results. In brief, there is a cluster of four methods that rank significantly better than the other methods, with one clear winner. The inter-scanner robustness ranking shows that not all methods generalize to unseen scanners.

The challenge remains open for future submissions and provides a public platform for method evaluation.
\end{abstract}

\blfootnote{Copyright (c) 2019 IEEE. Personal use of this material is permitted. However, permission to use this material for any other purposes must be obtained from the IEEE by sending a request to pubs-permissions@ieee.org.}

\IEEEpubidadjcol
\section{Introduction}
\label{sec:introduction}

\begin{figure*}[!t]
\centering
\subfloat[T1-weighted image]{\includegraphics[width=2in]{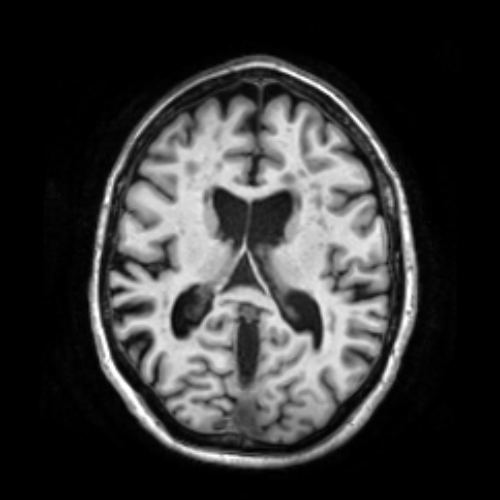}
\label{fig:example:t1}}
\hfil
\subfloat[FLAIR image]{\includegraphics[width=2in]{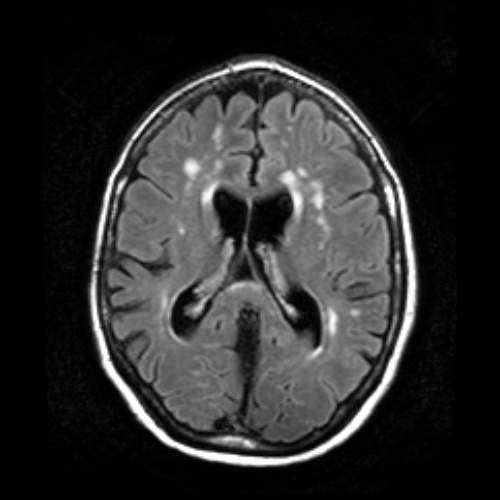}
\label{fig:example:flair}}
\hfil
\subfloat[Manual WMH segmentation]{\includegraphics[width=2in]{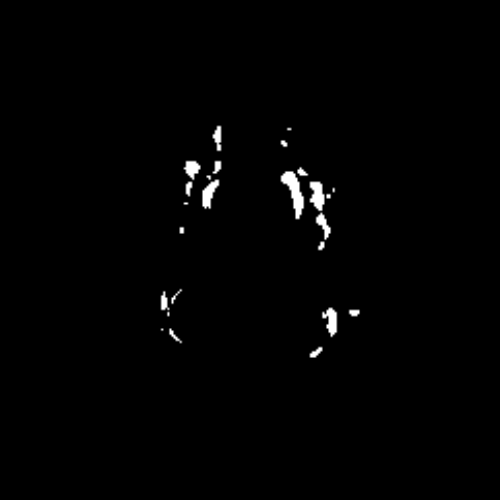}
\label{fig:example:wmh}}
\caption{Example brain MR images of a subject with white matter hyperintensities (WMH) of presumed vascular origin. On the T1-weighted image \protect\subref{fig:example:t1}, WMH show as hypointense regions within the white matter. On the FLAIR image \protect\subref{fig:example:flair}, WMH are clearly visible as hyperintense regions within the white matter. The corresponding manual WMH segmentation is shown in \protect\subref{fig:example:wmh}. }
\label{fig:example}
\end{figure*}

White matter hyperintensities (WMH) of presumed vascular origin are one of the main manifestations of cerebral small vessel disease and play a key role in stroke, dementia, and ageing \cite{Pantoni2010,Prins2015}. On T2-weighted and fluid-attenuated inversion recovery (FLAIR) brain MR images, WMH are clearly visible as hyperintense regions within the white matter\cite{Wardlaw2013}. An example image is shown in Figure~\ref{fig:example}, with the manual segmentation shown in Figure~\ref{fig:example}\subref{fig:example:wmh}. 

Quantification of WMH is of importance in clinical research studies, where measures of WMH volume, shape, and location are obtained from detailed segmentations. These measures are associated with the presence and severity of clinical symptoms, such as cognitive impairment and gait disturbances, and are likely to find their way into daily clinical practice, supporting diagnosis, prognosis, and treatment monitoring\cite{Prins2015,Biesbroek2017}. However, manual delineation of WMH is a time-consuming and observer-dependent procedure.

Automatic WMH segmentation methods have been developed, but a review by Caligiuri et al.\cite{Caligiuri2015} revealed a key issue: it is hard to compare the various methods that are described in the literature. Each proposed segmentation method has been evaluated on a different ground truth (different number of subjects, different experts, different protocols), using different evaluation criteria.

A further challenge of automatic WMH segmentation methods is the deployment of such a method within a new institute that might have different scanners or imaging protocols. Many (deep) machine learning methods require some form of transfer learning or fine-tuning on the target images\cite{Greenspan2016}, which in practice is not always feasible.

These issues are not unique to the task of automatic WMH segmentation, but occur in many medical image analysis tasks. Organizing a scientific challenge is a way to address this, having a number of competing methods perform the same task on the same data. This has been successfully applied to various tasks, such as liver segmentation\cite{Heimann2009}, image registration\cite{Murphy2011}, coronary calcium scoring\cite{Wolterink2016}, or gland segmentation in histology images\cite{Sirinukunwattana2017}. In the past, a number of challenges have been organized that included abnormalities on brain MR images, such as the multiple sclerosis (MS) lesions\cite{Styner2008,Commowick2018}, tumour\cite{Menze2015}, or tissue\cite{Mendrik2015} segmentation challenges\footnote{For a more complete overview, visit: \url{https://grand-challenge.org/challenges/}}. However, none of these challenges focuses on WMH of presumed vascular origin (although MS lesions share some characteristics with such WMH; and the brain tissue segmentation challenge included WMH lesions, but not as a separate task). 

The WMH Segmentation Challenge described in this paper provides a standardized assessment of automatic methods for the segmentation of WMH. The task for the challenge was defined as: ``the segmentation of white matter hyperintensities of presumed vascular origin on brain MR images''\footnote{\url{https://wmh.isi.uu.nl/details/}} \cite{Wardlaw2013}. Key features of the challenge include: Participants have to submit their method to the organizers for independent evaluation on a test set. The test set includes data from two additional scanners not in the training data, to evaluate generalizability of segmentation methods across scanners. The dataset was derived from patients with various degrees of ageing related degenerative and vascular pathologies, which is important for the generalizability since segmentation methods should be able to deal with this variation. Evaluation is performed using five different metrics and participants are ranked relative to each other.

In this paper, the organization of the challenge, its results, and a detailed evaluation are presented.

\section{Methods}
\label{sec:methods}
\subsection{Training and test data}
A total of 60 training and 110 test images were used in this challenge. Imaging data was acquired from five different scanners, from three different vendors, in three different institutes: the University Medical Center (UMC) Utrecht, VU University Medical Centre (VU) Amsterdam, both in the Netherlands, and the National University Health System (NUHS) in Singapore. For each subject, a 3D T1-weighted and a 2D multi-slice FLAIR image were provided.

\begin{table}[!t]
\renewcommand{\arraystretch}{1.3}
\caption{Overview of the number of images available for training (Tr.) and test (Te.).}
\label{tab:data}
\centering
\begin{tabular}{llll}
\hline
\textbf{Institute} & \textbf{Scanner} & \textbf{Tr.} & \textbf{Te.} \\
\hline\hline
UMC Utrecht    & 3~T Philips Achieva   & 20 & 30 \\
NUHS Singapore & 3~T Siemens TrioTim   & 20 & 30 \\
VU Amsterdam   & 3~T GE Signa HDxt     & 20 & 30 \\
               & 1.5~T GE Signa HDxt   &  0 & 10 \\
               & 3~T Philips Ingenuity (PET/MR) &  0 & 10 \\
\hline
\end{tabular}
\end{table}

The training data consisted of sixty images: twenty 3~T images of a single scanner of each institute. The test set included ninety images (three times thirty) of those same scanners and additionally twenty images (two times ten) of scanners that were not in the training data set. An overview of the data set is given in Table \ref{tab:data}.

Subjects included from UMC Utrecht and VU Amsterdam were selected from the memory clinic patients of both institutes\cite{Boomsma2017}.

Subjects included from the NUHS Singapore were selected from the Memory Ageing and Cognition Centre Cohort recruited from the memory clinics of the National University Hospital and St. Luke's Hospital in Singapore \cite{VanVeluw2015b}. 

For each scanner, subjects were randomly picked from all subjects and randomly placed into the training or test sets.

\subsubsection{MRI parameters}\label{sec:mri}
All 3D sequences were acquired in the sagittal direction and all 2D multi-slice sequences in the transversal direction.

\textbf{UMC Utrecht, 3~T Philips Achieva:} 3D T1-weighted sequence (192~slices, voxel size: $1.00\times 1.00\times 1.00$~mm$^3$, repetition time (TR)/echo time (TE): $7.9/4.5$~ms), 2D FLAIR sequence (48~slices, voxel size: $0.96\times 0.95\times 3.00$~mm$^3$, TR/TE/inversion time (TI): $11,000/125/2,800$~ms)

\textbf{NUHS Singapore, 3~T Siemens TrioTim:} 3D T1-weighted sequence (voxel size: $1.00\times 1.00\times 1.00$~mm$^3$, TR/TE/TI: $2,300/1.9/900$~ms), 2D FLAIR sequence (voxel size: $1.00\times 1.00\times 3.00$~mm$^3$, TR/TE/TI: $9,000/82/2,500$ ms)

\textbf{VU Amsterdam, 3~T GE Signa HDxt:} 3D T1-weighted sequence (176~slices, voxel size: $0.94\times 0.94\times 1.00$~mm$^3$, TR/TE: $7.8/3.0$~ms), 3D FLAIR sequence (132~slices, voxel size: $0.98\times 0.98\times 1.20$~mm$^3$, TR/TE/TI: $8,000/126/2,340$~ms)

\textbf{VU Amsterdam, 1.5~T GE Signa HDxt:} 3D T1-weighted sequence (172~slices, voxel size: $0.98\times 0.98\times 1.50$~mm$^3$, TR/TE: $12.3/5.2$~ms), 3D FLAIR sequence (128~slices, voxel size: $1.21\times 1.21\times 1.30$~mm$^3$, TR/TE/TI: $6,500/117/1,987$~ms)

\textbf{VU Amsterdam, 3~T Philips Ingenuity (PET/MR):} 3D T1-weighted sequence (180~slices, voxel size: $0.87\times 0.87\times 1.00$~mm$^3$, TR/TE: $9.9/4.6$~ms), 3D FLAIR sequence (321~slices, voxel size: $1.04\times 1.04\times 0.56$~mm$^3$, TR/TE/TI: $4,800/279/1,650$~ms)

All 3D FLAIR sequences were resampled into the transversal direction with slices of 3~mm thickness for two reasons: (1) to save time on the manual annotation of WMH and (2) to become more similar to the 2D multi-slice sequences.

An example FLAIR image of each scanner is shown in Appendix~\ref{app:images} Figure~\ref{app:extra_figures}\footnote{Available in the supplementary files / multimedia tab.}.

\subsubsection{Data pre-processing}
All images were bias-corrected using SPM12 \cite{Ashburner2000}. Using the \texttt{elastix} toolbox for image registration \cite{Klein2010}, the 3D T1-weighted images were aligned with the (resampled) FLAIR images. The transformation parameters were provided with the data. The faces of the subjects were manually removed from all sequences and the masks used for that were provided as well.

Data before and after preprocessing is provided on the challenge website for registered participants: \url{https://wmh.isi.uu.nl/data/}.

\subsubsection{Manual reference standard}
WMH and other pathologies (i.e. lacunes and non-lacunar infarcts, (micro)\allowbreak hemorrhages) were manually segmented in accordance with the STandards for ReportIng Vascular changes on nEuroimaging (STRIVE) criteria \cite{Wardlaw2013}. The outline of WMH and other pathology was delineated using a contour drawing technique by an expert observer (O1). This observer had extensive prior experience with the manual segmentation of WMH and had segmented 1000+ cases before this dataset. Manual delineations were peer-reviewed by a second expert observer (O2) with eleven years of experience in quantitative neuroimaging and clinical neuroradiology. In case of mistakes, errors, or delineations that were not according to the STRIVE criteria, O1 corrected the manual segmentation in a consensus meeting with O2. Hence, the provided reference standard is the corrected segmentation of O1, after peer review by O2.

The contours were converted to binary masks, whereby all voxels whose volume was within the manual delineation for \textgreater50~\%, were considered WMH. Background received label~0 and WMH label~1. Other pathology was converted to binary masks as well, receiving label~2. These masks were dilated by 1~pixel in-plane (with a $3\times 3\times 1$~voxel kernel). In case of overlap between labels 1 and 2 (after dilation), label 1 was assigned.

Two additional observers segmented the sixty training images to obtain inter-observer agreement measures. Observer O3 was trained for WMH segmentation, but had no extensive prior experience. Observer O4 was trained for WMH segmentation and had prior experience.

\subsection{Set-up of the challenge}
Participants could register on the challenge website and download the training data. Methods had to be containerized with Docker\footnote{\url{https://www.docker.com/}}\cite{Merkel2014a} and submitted for evaluation. Containerization eases deployment of methods and guarantees that the method will produce identical output when run on a different platform. To ensure this, the output of the containerized method on the first training subject was sent back to the participants for verification. 

During testing, the containerized method was run on each test subject one by one. No identifiers were present that would indicate from which of the five scanners the current subject originated. After processing a subject, the container was completely destroyed and reloaded. Full details on how the containers would be run, including a Python and MATLAB example container, were provided on the challenge website\footnote{\url{https://wmh.isi.uu.nl/methods/}}.

An NVIDIA Titan Xp GPU was available for methods that needed one.

\subsection{Participants}
Twenty teams submitted their method before the deadline and participated in the challenge. A brief summary of each method is given below, in alphabetical order. \\%
\achilles{} a neural network similar to HighResNet \cite{Li2017} and DeepLab v3 \cite{Chen2017}, utilizing atrous (dilated) convolutions, atrous spatial pyramid pooling, and residual connections. The network is trained only on the FLAIR images, taking random $71^3$ sized patches, and applying scaling and rotation augmentations \cite{Georgiou2017}. \\
\cian{} a network based on multi-dimensional gated recurrent units (MD-GRU) was trained on 3D patches using data augmentation techniques including random deformation, rotation and scaling \cite{Andermatt2016a,Andermatt2017,Andermatt2018}. \\     
\hadi{} a random forest classifier trained on multi-modal image features. These include intensities, gradient, and Hessian features of the original images, after smoothing, and of generated super-voxels \cite{Mahmood2017}. \\
\ipmibern{} a two-stage approach that uses fully convolutional neural networks to first extract the brain from the images and second identifies WMH within the brain. Both stages implement long and short skip connections. The second stage produces output at three different scales. Data augmentation was applied, including rotations and mirroring \cite{Zeng2017}. \\
\ktwo{} a 2D fully convolutional neural network with an architecture similar to U-Net \cite{Ronneberger2015}. A number of models were trained for the whole dataset, as well as for each individual scanner. During application, first the type of scanner was predicted and next that specific model was applied together with the model trained on all data \cite{Mehrtash2017}. \\ 
\knight{} a voxel-wise logistic regression model that is fitted independently for each voxel in the FLAIR image. Images were transformed to the MNI-152 standard space \cite{Fonov2011} for training and at test time the parameter maps were warped to the subject space \cite{Knight2017,Knight2018a}. \\
\lrde{} a modification of the pre-trained 16-layer VGG network \cite{Simonyan2015}, where the FLAIR, T1, and a high-pass filtered FLAIR are used as multi-channel input. The VGG network had its fully connected layers replaced by a number of convolutional layers \cite{Maninis2016,Shelhamer2017,Xu2017,Xu2018}. \\
\misp{} a 3D convolutional neural network with 18 layers using patches of $27\times 27 \times 9$ voxels. The first eight layers were trained separately for the FLAIR and T1 images and had skip-connections \cite{He2015} \cite{Luna2017}. \\
\neuroml{} a neural network using the DeepMedic \cite{Kamnitsas2017} architecture, having two parallel branches that process the images at two different scales. The network used 3D patches, which were sampled such that 60 \% of the patches contained a WMH \cite{Safiullin2017}. \\
\nicvicorob{} a 10-layer 3D convolutional neural network architecture previously used to segment multiple sclerosis lesions \cite{Valverde2017a}. A cascaded training procedure was employed, training two separate networks to first identify candidate lesion voxels and next to reduce false positive detections. A third network re-trains the last fully connected layer to perform WMH segmentation \cite{Valverde2017}. \\
\nihcidi{} a fully convolutional neural network modified from the U-Net architecture \cite{Ronneberger2015} was used to segment WMH on the FLAIR images. Next, another network was trained to segment the white matter from T1 images, and the segmented white matter mask is applied to remove false positives from the WMH segmentation results. The original U-Net architecture was trimmed to keep only three pooling layers \cite{Jin2017}. \\
\nist{} a random decision forest classifier trained on location and intensity features \cite{Dadar2017a,Dadar2017b,Dadar2017}. \\
\nlplogix{} a multiscale deep neural network similar to \cite{Ghafoorian2017}, with some minor modifications and no spatial features. The network was trained in ten folds and the three best performing checkpoints on the training data were selected. These were applied on the test set and the results averaged \cite{Berseth2017}. \\
\scan{} a densely connected convolutional network using dilated convolutions \cite{Yu2016,Huang2017}. In each dense block, the output is concatenated to the input before passing it to the next layer. Two classifiers were trained: one to apply brain extraction and the second to find lesions within the extracted brain \cite{McKinley2017}. \\
\skkumedneuro{} an intensity-based thresholding method with region growing approach to segment periventricular and deep WMH separately, and two random forest classifiers for false positive reduction. Per imaging modality, 19 texture and 100 ``multi-layer'' features were computed. The ``multi-layer'' features were computed using a feed-forward convolutional network with fixed filters (e.g. averaging, Gaussian, Laplacian); consisting of two convolutional, two max-pooling, and one fully connected layer \cite{Park2017}. \\
\sysumedia{} a fully convolutional neural network similar to U-Net \cite{Ronneberger2015}. An ensemble of three networks was trained with different initializations. Data normalization and augmentation was applied. To remove false positive detections, WMH in the first and last $\frac{1}{8}$th slices was removed \cite{Li2017a,Li2018}. \\
\textclass{} a random forest classifier trained primarily on texture features. Features include local binary pattern, structural and morphological gradients, and image intensities \cite{Bento2017,Bento2018}. \\
\tig{} a three-level Gaussian mixture model, slightly adapted from \cite{Sudre2015}. The model is iteratively modified and evaluated, until it converges. After that, candidate WMH is selected and possible false positives are pruned based on their location \cite{Sudre2017}. \\
\tignet{} a neural network with the HighResNet architecture \cite{Li2017}. The network was trained on $2,660$ images segmented using the previous method of team \tig{} \cite{Sudre2015,Sudre2017a}. \\
\upcdlmi{} a neural network modified from the V-Net architecture \cite{Milletari2016}. An additional network with convolutional layers is trained on upsampled images and then concatenated with the output of the V-Net \cite{Casamitjana2017}. \\

Detailed information on each method can be found online at \url{https://wmh.isi.uu.nl/results/results-miccai-2017/}.

\subsection{Evaluation and Ranking}
Methods were evaluated according to five criteria: (1) the Dice Similarity Coefficient (DSC), (2) a modified Hausdorff distance (95th percentile; H95), (3) the absolute percentage volume difference (AVD), (4) the sensitivity for detecting individual lesions (recall), and (5) F1-score for individual lesions (F1). For recall and F1, individual lesions are defined as 3D connected components within an image. The exact implementation of each metric was put online\footnote{\url{https://github.com/hjkuijf/wmhchallenge/blob/master/evaluation.py}} beforehand and could be used by participants for self-evaluation during development. During evaluation of the results, it was discovered that the AVD metric had a slight flaw. A method could undersegment WMH by at most $100\%$, but could oversegment WMH almost infinitely. Therefore, in this manuscript, the AVD metric was replaced by the absolute log-transformed volume difference (lAVD, (\ref{eg:lavd})).

\begin{equation}\label{eg:lavd}
\text{lAVD} = | \log\frac{\text{segmented volume}}{\text{true volume}} |
\end{equation}

The final ranking was based on the five metrics and each method received a rank relative to the performance of all methods. This was computed in a number of steps. First, each metric was averaged over all test scans per method. For each metric, the methods were sorted from best to worst. Next, the best method received a rank of $0$ and the worst method a rank of $1$; all other methods were ranked relatively in the range $(0, 1)$. Finally, the five ranks were averaged into the overall rank.

$95 \%$ confidence intervals on each individual metric and the final ranking were computed using bootstrapping. The bootstrap distribution included $2,000$ samples taken randomly from the test set with replacement. Non-overlapping confidence intervals indicate a significant difference between methods, with $\alpha=0.05$.

It is expected that methods might have more difficulties detecting and segmenting small lesions compared to large lesions. For each subject, the recall will be computed separately for individual lesions smaller than or equal to the median lesion size and for lesions larger than the median lesion size.

Additionally, a ranking was computed based solely on the inter-scanner differences. This ranking highlights which methods have the most robust performance across various scanners. For each method and scanner, the median performance of each metric was computed. Next, the standard deviation of those medians per scanner was averaged; giving a single value per metric: the standard deviation of the median per scanner. Methods were then ranked based on this value: first per metric and then averaged over all five metrics. A lower standard deviation across the median performance on all scanners (for all metrics) indicates a better inter-scanner robustness. 

Finally, the Simultaneous Truth And Performance Level Estimation (STAPLE) algorithm \cite{Warfield2004} 
was applied to all methods and to the top-ranking methods. STAPLE takes multiple segmentations as input and produces a combined segmentation, which was evaluated and ranked separately. It has been shown for other applications, e.g. brain tumour segmentation\cite{Menze2015}, that fusing the output of multiple methods can outperform all individual methods.

\section{Results}
\label{sec:results}
\begin{figure}[!t]
\centering
\includegraphics[width=3.4in]{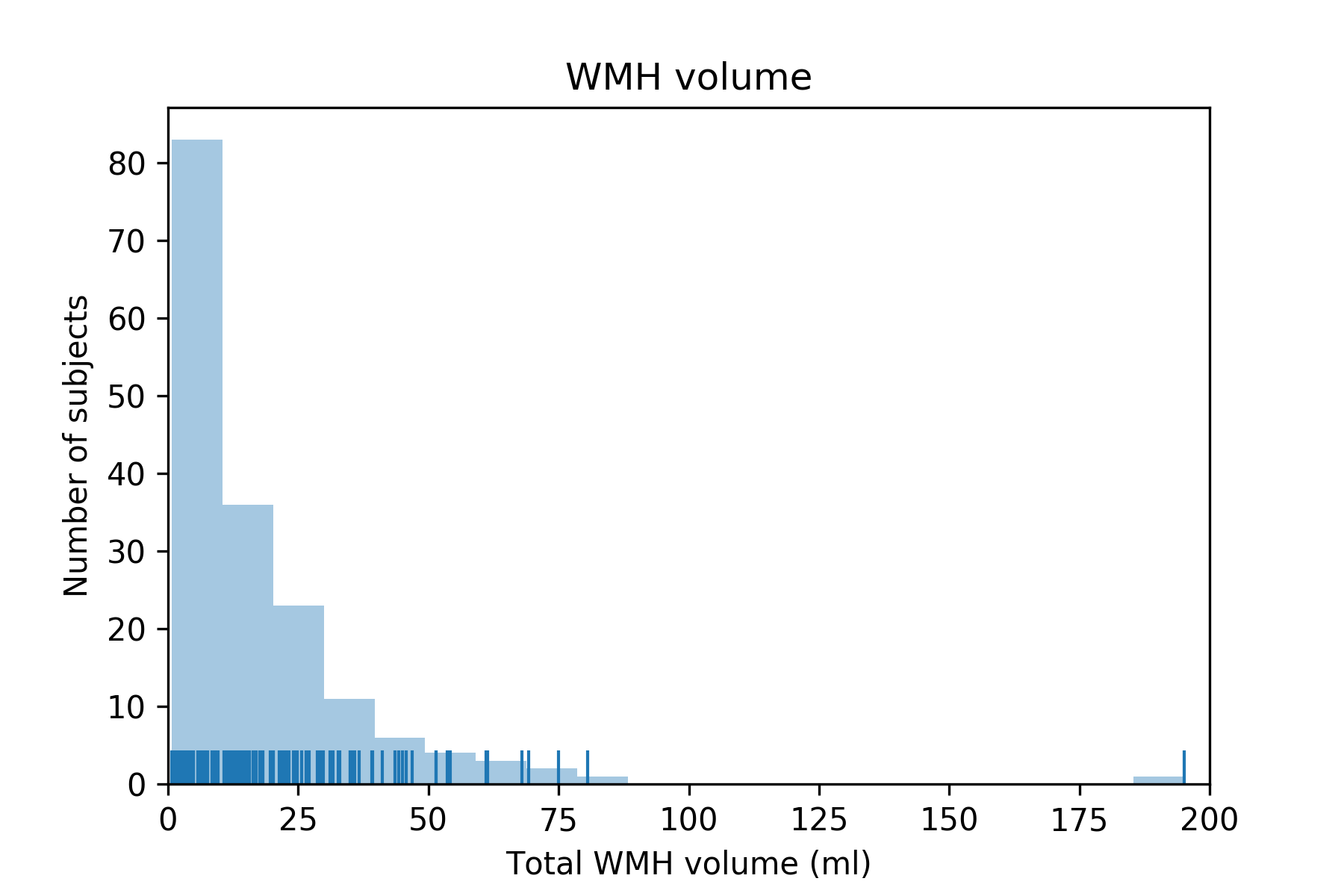}
\caption{Histogram showing the WMH volume distribution throughout the dataset. The ticks on the x-axis represent each individual subject.}
\label{fig:results_wmh_histogram}
\end{figure}

\begin{figure}[!t]
\centering
\includegraphics[width=3.4in]{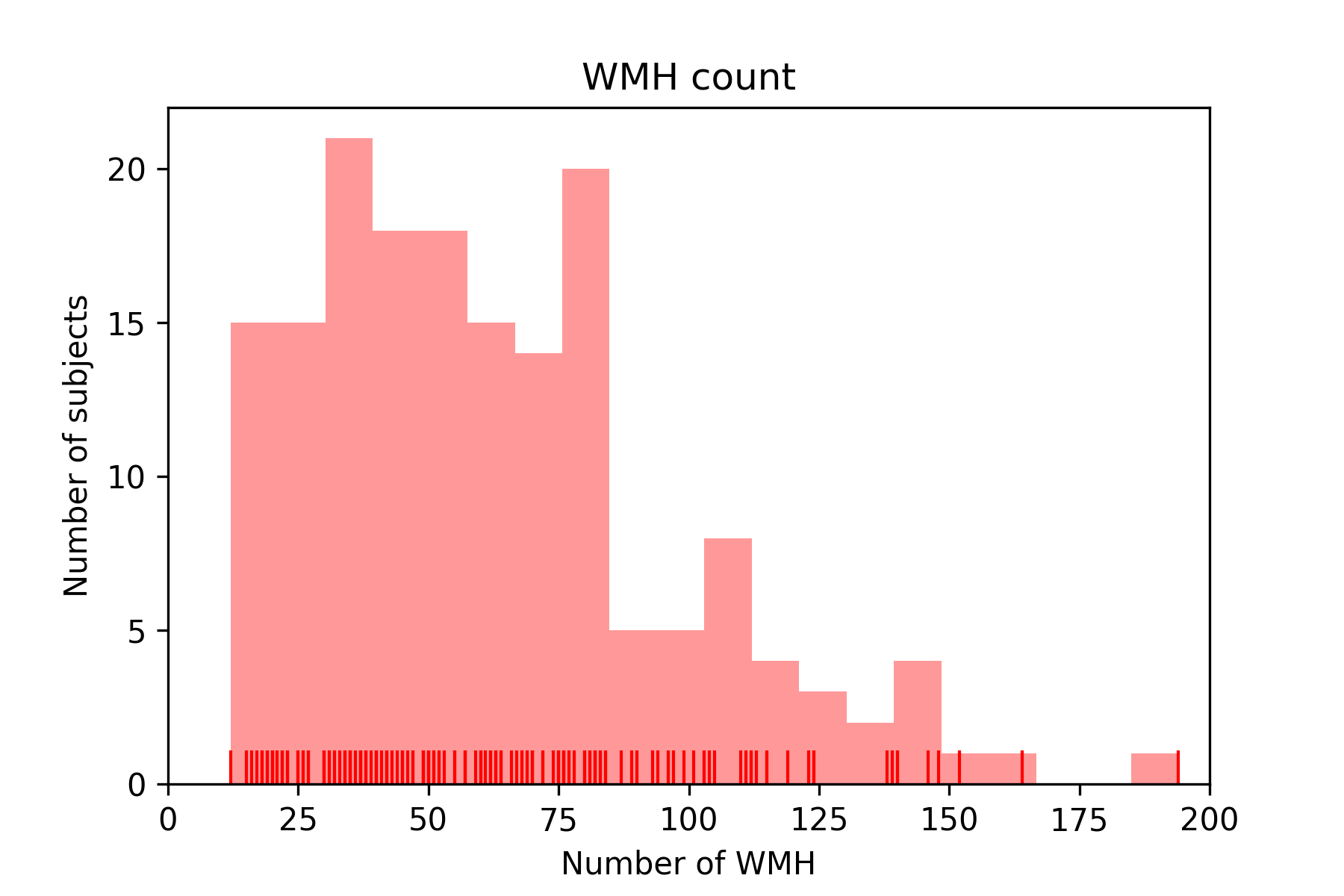}
\caption{Histogram showing the WMH count distribution throughout the dataset. The ticks on the x-axis represent each individual subject. An individual lesion is defined as a 3D connected component within an image. }
\label{fig:results_wmh_histogram_count}
\end{figure}

\begin{figure*}[!t]
\centering
\includegraphics[width=7in]{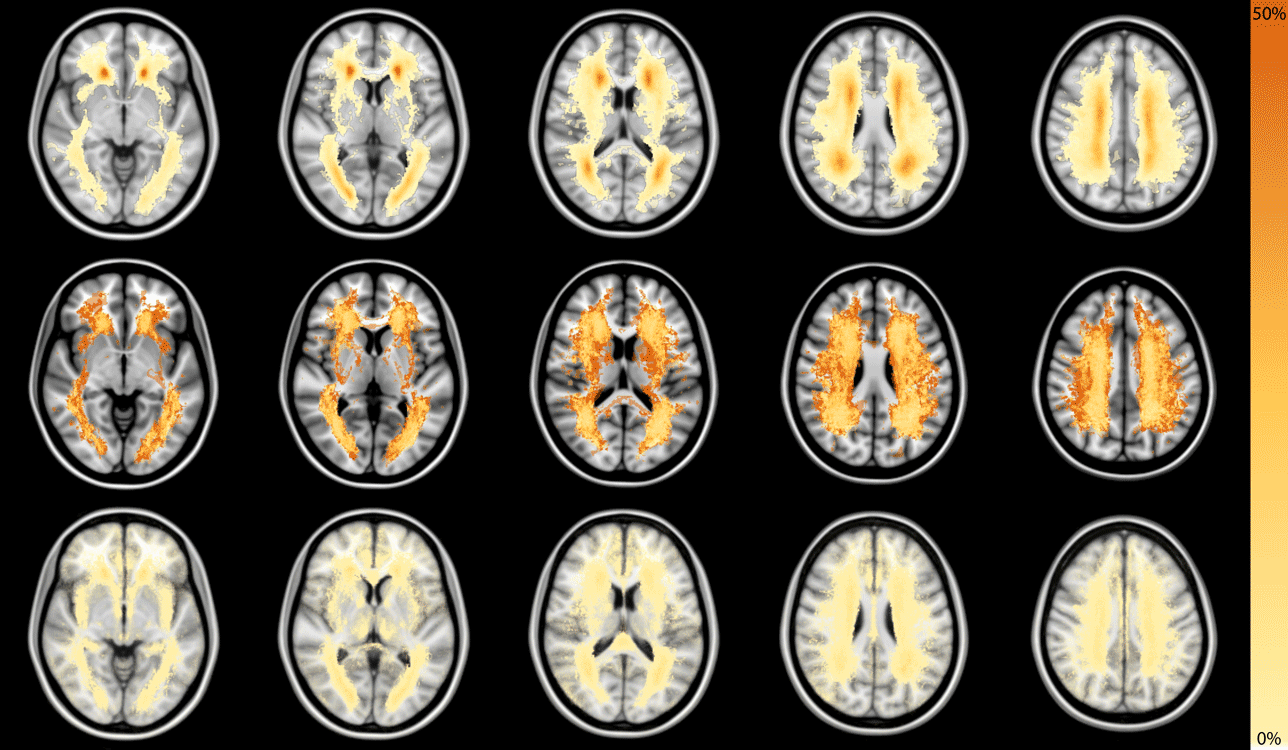}
\caption{The MNI-152 standard brain template\cite{Fonov2011}, showing different overlays. Top row: WMH distribution throughout the dataset, where the colour indicates the percentage of subjects that have a lesion in that specific voxel. Middle row: false negative rate, showing the percentage of lesions that were missed in a specific voxel. Bottom row: false positive rate, showing the percentage of false positives in a specific voxel. All voxels where only one subject has a lesion are shown half translucent.}
\label{fig:mni_lesion_overview}
\end{figure*}

The subjects included in the challenge were (mean $\pm$ sd) $70.1 \pm 9.3$~years old and 50~\% were male. 
The WMH volume in the dataset was (mean $\pm$ sd): $16.9 \pm 21.6$~ml (min: 0.78~ml, Q1: 3.24~ml, median: 11.18~ml, Q3: 23.00~ml, max: 195.15~ml; see Figure~\ref{fig:results_wmh_histogram}). 
The WMH count in the dataset was (mean $\pm$ sd): $62 \pm 35$~lesions (min: 12~lesions, Q1: 36~lesions, median: 57~lesions, Q3: 81~lesions, max: 194~lesions; see Figure~\ref{fig:results_wmh_histogram_count}). 
The distribution of lesions throughout the dataset is shown in the top row of Figure~\ref{fig:mni_lesion_overview}.
There were no significant differences between the training and test sets for age ($p=0.45$), gender ($p=0.87$), WMH volume ($p=0.74$), WMH count ($p=0.75$), the presence of lacunes ($p=0.86$), nor for the volume of other pathology ($p=0.62$). Tests for age and volumes were performed using Welch's unequal variances t-test \cite{Welch1947}. Tests for gender and presence of lacunes were performed using Fisher's exact test.

The bottom row of Table~\ref{tab:results_metrics} shows the inter-observer agreement of observers O3 and O4 compared with the manual reference standard of the sixty training images. 
Additionally, the associated positions of O3 and O4 with respect to all methods in the ranking is provided. This is the position these observers would have achieved if they had participated as method in the challenge.

\begin{figure}[!t]
\centering
\includegraphics[width=3.4in]{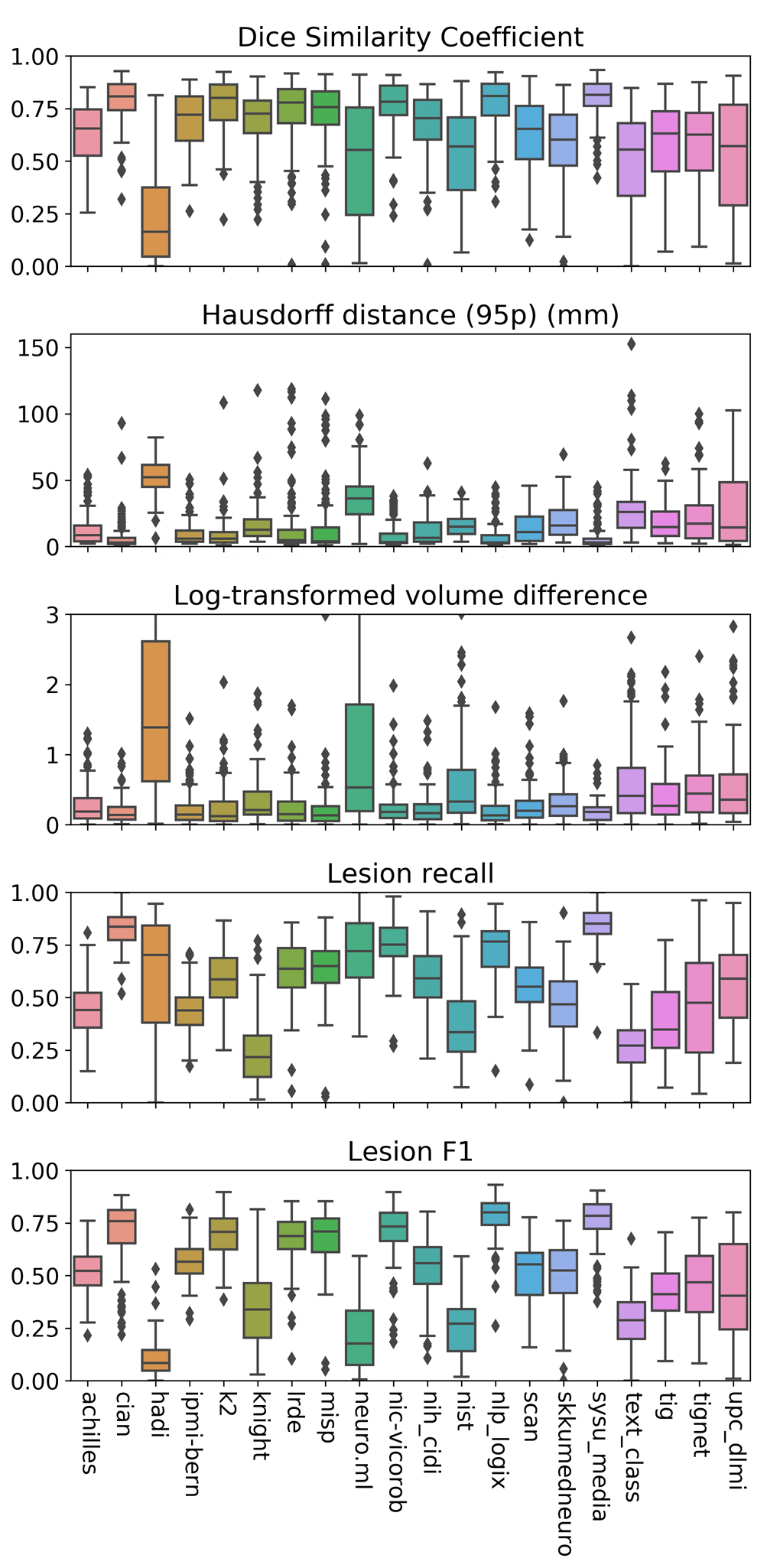}
\caption{Boxplots showing all five metrics per method. The box indicates the interquartile range (IQR) with a line at the median. The whiskers extend up to 1.5 times the IQR and the fliers indicate the remaining data points. Note for the Hausdorff distance that \hadi{} did not produce any output for 10 subjects and hence their boxplot is based on only 100 subjects (see Appendix \ref{app:summaries} Figure \ref{app:hadi} for full details). Note for the log-transformed volume difference that for visibility purposes, this figure is clipped at 3.0. Teams \hadi{}, \lrde{}, \misp{}, \neuroml{}, \nihcidi{}, \nist{}, \skkumedneuro{}, \textclass{}, and \upcdlmi{} have lAVD values above 3.0. For full details, see Appendix \ref{app:summaries} Figures \ref{app:hadi}, \ref{app:lrde}, \ref{app:misp}, \ref{app:neuro.ml}, \ref{app:nih_cidi}, \ref{app:nist}, \ref{app:skkumedneuro}, \ref{app:text_class}, and \ref{app:upc_dlmi}, respectively. }
\label{fig:results_metrics}
\end{figure}

The mean performance of each participating method on each individual metric is shown in Table~\ref{tab:results_metrics}, together with the $95\%$ confidence intervals. The method of \sysumedia{} performed best on the DSC, H95, and recall metrics. The method of \cian{} performed best on the lAVD metric. The method of \nlplogix{} performed best on the F1 metric. Figure \ref{fig:results_metrics} shows boxplots of all results of each method on each metric.

\begin{table*}[!t]
\renewcommand{\arraystretch}{1.3}
\caption{Mean performance and 95~\% confidence intervals of each participating method on each individual metric. Metrics include: (1) Dice Similarity Coefficient (DSC), (2) modified Hausdorff distance (95th percentile; H95), (3) absolute of the percentage volume difference (AVD), (4) sensitivity for detecting individual lesions (recall), and (5) F1-score for individual lesions (F1). Bold indicates that a method has the best score on that metric. Methods are sorted based on the final ranking as shown in Table~\ref{tab:results_ranking}. The bottom rows include the results of the Simultaneous Truth And Performance Level Estimation (STAPLE) algorithm applied on all methods and on the top~4 ranking methods, and the results of observers O3 and O4, together with the associated positions in the ranking if STAPLE, O3, and O4 would have participated in the challenge. Note that O3 and O4 segmented the sixty training images. }
\label{tab:results_metrics}
 \begin{center}
\begin{tabular}{rllllll}
\hline
\textbf{\#} & \textbf{Team} & \textbf{DSC}  & \textbf{H95 (mm)} & \textbf{lAVD} & \textbf{Recall} & \textbf{F1} \\
\hline\hline
 1 & sysu\_media  & \textbf{0.80 (0.78 - 0.82)} &  \textbf{6.30 (4.75 -  7.93)} &  0.193 (0.165 - 0.224) & \textbf{0.84 (0.82 - 0.86)}$^\ast$ & 0.76 (0.73 - 0.78)$^\dagger$ \\
 2 & cian         & 0.78 (0.76 - 0.80) &  6.82 (4.92 -  9.22) & \textbf{0.193 (0.162 - 0.228)} & 0.83 (0.81 - 0.84)$^\ast$ & 0.70 (0.67 - 0.73)$^\ddagger$ \\
 3 & nlp\_logix   & 0.77 (0.75 - 0.80) &  7.16 (5.61 -  8.82) &  0.219 (0.174 - 0.271) & 0.73 (0.71 - 0.76)$^{\ast\ast}$ & \textbf{0.78 (0.76 - 0.80)}$^\dagger$ \\
 4 & nic-vicorob  & 0.77 (0.74 - 0.79) &  8.28 (6.60 - 10.06) &  0.248 (0.201 - 0.303) & 0.75 (0.73 - 0.77)$^{\ast\ast}$ & 0.71 (0.68 - 0.73)$^\ddagger$ \\
 5 & k2           & 0.77 (0.74 - 0.79) &  9.79 (7.72 - 12.28) &  0.246 (0.187 - 0.310) & 0.59 (0.56 - 0.61) & 0.70 (0.68 - 0.72)$^\ddagger$ \\
 6 & misp         & 0.72 (0.69 - 0.75) & 14.88 (10.52 - 19.41) &  0.258 (0.167 - 0.388) & 0.63 (0.60 - 0.65) & 0.68 (0.65 - 0.70)$^\ddagger$ \\
 7 & lrde         & 0.73 (0.70 - 0.76) & 14.54 (10.32 - 19.31) &  0.309 (0.218 - 0.442) & 0.63 (0.60 - 0.66) & 0.67 (0.65 - 0.69)$^\ddagger$ \\
 8 & nih\_cidi    & 0.68 (0.65 - 0.70) & 12.82 (10.54 - 15.16) & 0.281 (0.200 - 0.394) & 0.59 (0.56 - 0.62) & 0.54 (0.51 - 0.57) \\
 9 & ipmi-bern    & 0.69 (0.67 - 0.72) &  9.72 (7.98 - 11.56) &  0.225 (0.178 - 0.275) & 0.44 (0.42 - 0.46) & 0.57 (0.55 - 0.58) \\
10 & scan         & 0.63 (0.59 - 0.66) & 14.34 (12.25 - 16.50) &  0.277 (0.223 - 0.336) & 0.55 (0.52 - 0.58) & 0.51 (0.48 - 0.53) \\
11 & achilles     & 0.63 (0.60 - 0.66) & 11.82 (9.80 - 13.94) &  0.276 (0.226 - 0.331) & 0.45 (0.42 - 0.47) & 0.52 (0.50 - 0.53) \\
12 & skkumedneuro & 0.58 (0.54 - 0.61) & 19.02 (16.64 - 21.58) &  0.384 (0.292 - 0.503) & 0.47 (0.44 - 0.49) & 0.51 (0.48 - 0.54) \\
13 & tignet       & 0.59 (0.56 - 0.63) & 21.58 (18.15 - 25.33) &  0.533 (0.450 - 0.623) & 0.46 (0.41 - 0.51) & 0.45 (0.42 - 0.49) \\
14 & tig          & 0.60 (0.56 - 0.63) & 17.86 (15.57 - 20.20) &  0.400 (0.333 - 0.474)& 0.38 (0.36 - 0.41) & 0.42 (0.40 - 0.44) \\
15 & knight       & 0.70 (0.67 - 0.72) & 17.03 (14.48 - 19.88) &  0.352 (0.290 - 0.427) & 0.25 (0.22 - 0.27) & 0.35 (0.32 - 0.38) \\
16 & upc\_dlmi    & 0.53 (0.48 - 0.58) & 27.01 (22.25 - 31.99) & 0.612 (0.481 - 0.762) & 0.57 (0.53 - 0.60) & 0.42 (0.38 - 0.46) \\
17 & nist         & 0.53 (0.49 - 0.57) & 15.91 (14.44 - 17.42) & 0.581 (0.469 - 0.695) & 0.37 (0.34 - 0.40) & 0.25 (0.22 - 0.27) \\
18 & neuro.ml     & 0.51 (0.45 - 0.56) & 37.36 (33.70 - 40.89) & 1.033 (0.836 - 1.241) & 0.71 (0.68 - 0.75)$^{\ast\ast}$ & 0.21 (0.19 - 0.24) \\
19 & text\_class  & 0.50 (0.45 - 0.54) & 28.23 (24.15 - 32.68) & 0.605 (0.492 - 0.724) & 0.27 (0.25 - 0.29) & 0.29 (0.26 - 0.31) \\
20 & hadi         & 0.23 (0.19 - 0.27) & 52.02 (49.25 - 54.82) & 1.685 (1.448 - 1.939) & 0.58 (0.52 - 0.63) & 0.11 (0.09 - 0.12) \\
\hline
4 & STAPLE (all)   & 0.77 (0.74 - 0.80) & \textbf{5.74 (4.26 - 7.43)} & 0.315 (0.249 - 0.393) & 0.77 (0.75 - 0.79) & 0.74 (0.71 - 0.76) \\
2 & STAPLE (top 4) & \textbf{0.80 (0.78 - 0.82)} & 6.43 (4.48 - 8.81) & \textbf{0.171 (0.144 - 0.201)} & 0.80 (0.78 - 0.82) & 0.76 (0.74 - 0.78) \\
\hline
5 & O3 & 0.77 (0.74 - 0.80) & 6.79 (5.32 - 8.54) & \textbf{0.176 (0.135 - 0.222)} & 0.65 (0.62 - 0.69) & 0.74 (0.71 - 0.76) \\
4 & O4 & 0.79 (0.76 - 0.81) & 7.22 (5.36 - 9.36) & 0.195 (0.148 - 0.245) & 0.66 (0.63 - 0.70) & 0.76 (0.73 - 0.78) \\
\hline
\end{tabular}
\end{center}
$^\ast$ \sysumedia{} and \cian{} perform significantly better on the recall metric than all other teams.\\
$^{\ast\ast}$ \nicvicorob{}, \nlplogix{}, and \neuroml{} perform significantly better on the recall metric than all remaining teams.\\
$^\dagger$ \nlplogix{} and \sysumedia{} perform significantly better on the F1 metric than all other teams.\\
$^\ddagger$ \nicvicorob{}, \ktwo{}, \cian{}, \misp{}, and \lrde{} perform significantly better on the F1 metric than all remaining teams.
\end{table*}

The final ranking is shown in Table~\ref{tab:results_ranking}, together with the $95\%$ confidence intervals.

\begin{table}[!t]
\renewcommand{\arraystretch}{1.3}
\caption{Final ranking of the methods that participated in the challenge. The column Rank shows the relative performance of each method, based on all five metrics listed in Table~\ref{tab:results_metrics}, together with the 95~\% confidence intervals. The column Inter-scanner rank shows the ranking when it is computed solely based on inter-scanner robustness. The symbols between brackets indicate whether a team is ranked on the same position ($-$), lower ($\vee$), or higher ($\wedge$) compared with the original ranking; with the new position indicated as well. Dotted lines indicate clusters of methods that rank significantly different from methods ranked above/below, because of non-overlapping confidence intervals. }
\label{tab:results_ranking}
\begin{center}
\begin{tabular}{rll|l}
\hline
\textbf{\#} & \textbf{Team} & \textbf{Rank} (95 \% CI) & \textbf{Inter-scanner rank} \\
\hline\hline
 1 & sysu\_media  & 0.0068 (0.0019 - 0.0161)$^\dagger$ & 0.0375 ($\vee$ 2)    \\
\cdashline{1-3}
 2 & cian         & 0.0357 (0.0248 - 0.0539)$^\ddagger$ & 0.0831 ($\vee$ 5) \\
 3 & nlp\_logix   & 0.0520 (0.0365 - 0.0744)$^\ddagger$ & 0.1111 ($\vee$ 7) \\
 4 & nic-vicorob  & 0.0785 (0.0577 - 0.1045)$^\ddagger$ & 0.1629 ($\vee$ 11)\\
\cdashline{1-3}
 5 & k2           & 0.1437 (0.1188 - 0.1711) & 0.1174 ($\vee$ 8) \\
 6 & misp         & 0.1740 (0.1356 - 0.2273) & 0.1915 ($\vee$ 12)\\
 7 & lrde         & 0.1782 (0.1395 - 0.2290) & 0.3510 ($\vee$ 17)\\
 8 & nih\_cidi    & 0.2376 (0.2131 - 0.2680) & 0.1570 ($\vee$ 10)\\
 9 & ipmi-bern    & 0.2537 (0.2391 - 0.2727) & 0.0345 ($\wedge$ 1)\\
10 & scan         & 0.2836 (0.2631 - 0.3099) & 0.2252 ($\vee$ 14)\\
11 & achilles     & 0.3058 (0.2896 - 0.3276) & 0.0714 ($\wedge$ 3)\\
\cdashline{1-3}
12 & skkumedneuro & 0.3649 (0.3325 - 0.4044) & 0.1105 ($\wedge$ 6)\\
13 & tignet       & 0.4090 (0.3765 - 0.4481) & 0.2969 ($\vee$ 15)\\
14 & tig          & 0.4097 (0.3795 - 0.4454) & 0.1289 ($\wedge$ 9)\\
15 & knight       & 0.4320 (0.4082 - 0.4598) & 0.0785 ($\wedge$ 4)\\
16 & upc\_dlmi    & 0.4429 (0.3903 - 0.5016) & 0.7415 ($\vee$ 20)\\
17 & nist         & 0.5040 (0.4724 - 0.5404) & 0.3052 ($\wedge$ 16)\\
18 & neuro.ml     & 0.5615 (0.5193 - 0.6084) & 0.6110 ($\vee$ 19)\\
19 & text\_class  & 0.5961 (0.5539 - 0.6430) & 0.2117 ($\wedge$ 13)\\
\cdashline{1-3}
20 & hadi         & 0.8886 (0.8687 - 0.9103) & 0.4974 ($\wedge$ 18)\\
\hline
\end{tabular}
\end{center}
$^\dagger$ sysu\_media ranks significantly higher than all other participants.\\
$^\ddagger$ cian, nlp\_logix, and nic-vicorob rank significantly higher than all remaining participants.
\end{table}

\begin{figure}[!t]
\centering
\includegraphics[width=3.4in]{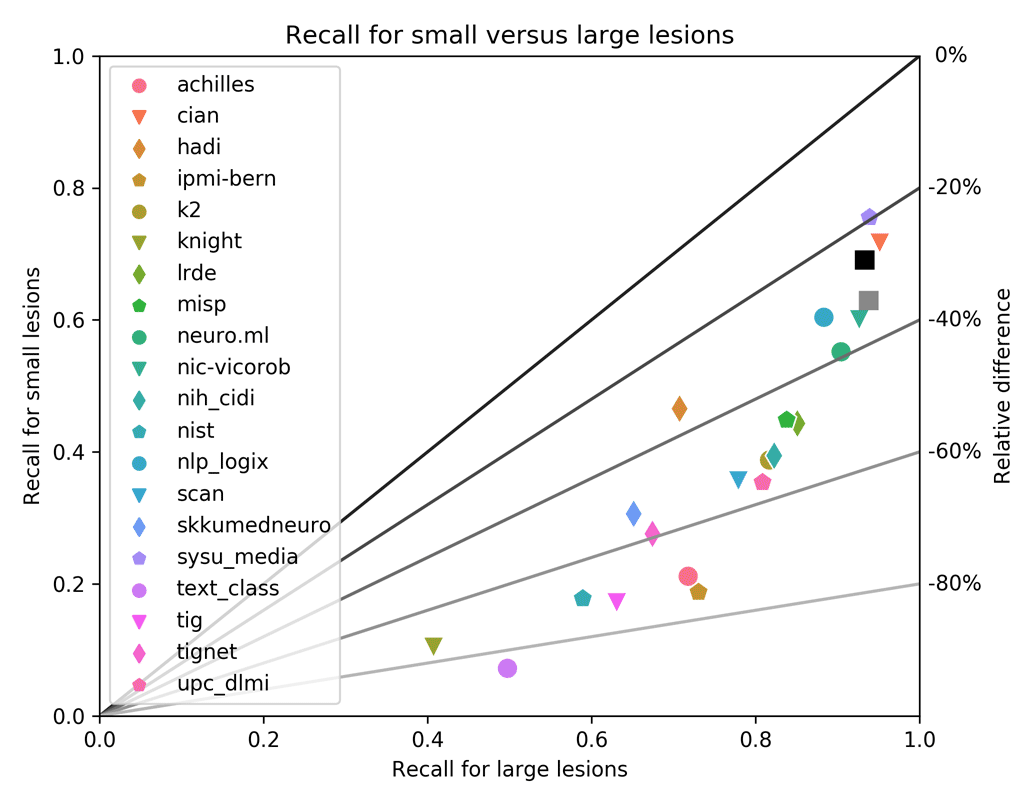}
\caption{Plot showing the recall of each method for small and large lesions. The right vertical axis indicates the relative difference for small lesions with respect to that of large lesions. Small lesions are defined as all lesions smaller than or equal to the median lesion volume per subject. Large lesions are all lesions larger than the median lesion volume per subject. The black and grey squares indicate the results of STAPLE applied on the top~4 or all methods, respectively.}
\label{fig:small_large_recall}
\end{figure}

The middle and bottom rows of Figure~\ref{fig:mni_lesion_overview} show spatial maps of the false negative rate and false positive rate, respectively, of all methods combined. Appendix~\ref{app:summaries} Figures~\ref{app:sysu_media}--\ref{app:hadi}\footnote{Available in the supplementary files / multimedia tab.} show these spatial maps per method, ordered by their final ranking.

Figure~\ref{fig:small_large_recall} shows the (relative) difference in recall between small and large lesions. All methods perform worse in recalling small lesions compared to large lesions. For example, the method of \sysumedia{} has a recall for large lesions of $94\%$ and a recall for small lesions of $76\%$, resulting in a relative difference of $-20\%$. Overall, the drop in recall ranges from $-20\%$ (\sysumedia{}) to $-87\%$ (\textclass{}), as indicated by the solid lines in the figure.

Table~\ref{tab:results_methods} highlights various properties of all methods, sorted by their final ranking. The top~11 methods all employ some form of deep learning, with a U-Net-like architecture\cite{Ronneberger2015} being the overall most common. Amongst the non-deep learning methods, the use of a random forest classifier is most common. Almost all methods apply various kinds of pre-processing, where normalizing intensities of an image to a standardized range is applied by most methods. Some methods apply post-processing techniques, mainly aimed at reducing the number of false positive detections. The H95 and F1 metrics are most sensitive to false positive detections, but the methods that apply post-processing do not have a notable better score on these metrics than nearby ranking methods. When considering only deep learning methods, most use data augmentation to generate more training samples. Scaling, rotating, and mirroring an image are quite common, but the top~2 methods also apply shearing or non-linear deformations. The last columns of Table~\ref{tab:results_methods} highlight some properties of deep learning methods, in which a few clusters can be distinguished. Top ranking methods have applied dropout during training, some form of hard negative mining, and use an ensemble of networks. Three methods use dilated convolutions, but these cluster in the middle of the ranking. Most methods that use 3D convolutions appear to rank at the bottom. Using batch normalization, multi scale approaches, or learning rate schedules does not seem to influence the ranking; and neither does the choice of loss function.

\begin{table*}[!t]%
\renewcommand{\arraystretch}{1.3}%
\setlength{\tabcolsep}{5pt}%
\caption{Overview of various properties of all methods. Methods are sorted based on the final ranking as shown in Table~\ref{tab:results_ranking}. }%
\label{tab:results_methods}%
 \begin{center}%
\begin{tabular}{rllllllllllllllll}%
\hline%
&&&&&&&&&\multicolumn{8}{l}{Neural network features$^5$}\\
\cline{10-17}
\textbf{\#} &\textbf{Team}         & \textbf{Pre}$^1$                                       & \textbf{Method}     & \textbf{Post}$^2$    & \textbf{Data}                            & \textbf{DL}$^3$        & \textbf{Aug}$^4$                            & \textbf{Loss function}  & \textbf{Dim} & \textbf{Dil}          & \textbf{BN}           & \textbf{Drop}         & \textbf{MS}           & \textbf{LR}           & \textbf{HN}           & \textbf{Ens}     \\
\hline                                                                                                                                                                                                                                                                                            \hline                                                                                                                                                                                                                                                                                            
1&sysu\_media  & \textsc{i},\textsc{r}                       & U-Net       & \textsc{Sl} & \textsc{t},\textsc{f}         & $\checkmark$  & \textsc{h},\textsc{r},\textsc{s} & DSC            & 2D  &              &              &              &              &              &              & $\checkmark$ \\
2&cian         & \textsc{f},\textsc{i}                       & MDGRU      &             & \textsc{t},\textsc{f}         & $\checkmark$  & \textsc{d},\textsc{r},\textsc{s} & multinom. log. & 2D$\dagger$ &              &              & $\checkmark$ &              &              &              &              \\
3&nlp\_logix   & \textsc{i},\textsc{s}                       & CNN        &             & \textsc{t},\textsc{f}         & $\checkmark$  &                                    & cross-entropy  & 2D  &              &              & $\checkmark$ & $\checkmark$ & $\checkmark$ & $\checkmark$ & $\checkmark$ \\
4&nic-vicorob  & $ $                                           & CNN        & \textsc{Sm} & \textsc{t},\textsc{f}         & $\checkmark$  & \textsc{m},\textsc{r}            & cross-entropy  & 3D  &              & $\checkmark$ & $\checkmark$ &              &              & $\checkmark$ &              \\
5&k2           & \textsc{i},\textsc{r},\textsc{s}            & U-Net       &             & \textsc{t},\textsc{f}         & $\checkmark$  & \textsc{m}                       & DSC            & 2D  &              &              & $\checkmark$ &              & $\checkmark$ &              & $\checkmark$ \\
6&misp         & \textsc{i},\textsc{r}                       & CNN        &             & \textsc{t},\textsc{f}         & $\checkmark$  &                                    & mean sq. error & 3D  &              & $\checkmark$ & $\checkmark$ &              &              & $\checkmark$ &              \\
7&lrde         & \textsc{f},\textsc{i}                       & VGG-16     &             & \textsc{t},\textsc{f}         & $\checkmark$  & \textsc{r},\textsc{s}            & multinom. log. & 2D  &              &              &              &              & $\checkmark$ &              &              \\
8&nih\_cidi    & \textsc{s}                                  & U-Net       & \textsc{g}  & \textsc{t},\textsc{f}         & $\checkmark$  & \textsc{m},\textsc{r}            & cross-entropy  & 2D  &              & $\checkmark$ &              &              & $\checkmark$ &              &              \\
9&ipmi-bern    & \textsc{i},\textsc{s}                       & U-Net       &             & \textsc{t},\textsc{f}         & $\checkmark$  & \textsc{m},\textsc{r}            & cross-entropy  & 2D  &              & $\checkmark$ &              & $\checkmark$ & $\checkmark$ &              &              \\
10&scan         & \textsc{s}                                  & DenseNet   &             & \textsc{t},\textsc{f}         & $\checkmark$  &                                    & cross-entropy  & 2D  & $\checkmark$ &              &              &              &              &              &              \\
11&achilles     & \textsc{i},\textsc{r}                       & HighResNet &             & \textsc{f}                    & $\checkmark$  & \textsc{r},\textsc{s}            & DSC            & 3D  & $\checkmark$ & $\checkmark$ &              & $\checkmark$ &              &              &              \\
12&skkumedneuro & \textsc{i},\textsc{s}                       & RF         &             & \textsc{t},\textsc{f}         &               &                                    &                &     &              &              &              &              &              &              &              \\
13&tignet       & \textsc{b},\textsc{i},\textsc{t}            & HighResNet &             & \textsc{t},\textsc{f}$^\ast$    & $\checkmark$  &                                    & DSC            & 3D  & $\checkmark$ & $\checkmark$ &              &              & $\checkmark$ &              &              \\
14&tig          & \textsc{b},\textsc{s},\textsc{t}            & GMM        & \textsc{fp} & \textsc{t},\textsc{f}         &               &                                    &                &     &              &              &              &              &              &              &              \\
15&knight       & \textsc{b},\textsc{i},\textsc{s},\textsc{t} & VLR        & \textsc{Sm} & \textsc{f}$^\ast$               &               & \textsc{m},\textsc{t},\textsc{y} & DSC            &     &              &              &              &              &              &              &              \\
16&upc\_dlmi    & \textsc{i}                                  & U-Net       &             & \textsc{t},\textsc{f}         & $\checkmark$  & \textsc{m}                       & DSC            & 3D  &              & $\checkmark$ &              & $\checkmark$ & $\checkmark$ &              &              \\
17&nist         & \textsc{b},\textsc{i},\textsc{t}            & RF         &             & \textsc{t},\textsc{f}         &               &                                    &                &     &              &              &              &              &              &              &              \\
18&neuro.ml     & $ $                                           & DeepMedic  &             & \textsc{t},\textsc{f}         & $\checkmark$  &                                    & cross-entropy  & 3D  &              &              &              & $\checkmark$ & $\checkmark$ &              &              \\
19&text\_class  & \textsc{i},\textsc{r}                       & RF         & \textsc{Sm} & \textsc{t},\textsc{f}         &               &                                    &                &     &              &              &              &              &              &              &              \\
20&hadi         & $ $                                           & RF         &             & \textsc{t},\textsc{f}         &               &                                    &                &     &              &              &              &              &              &              &              \\
\hline
\end{tabular}
\end{center}
$^1$ Pre-processing: \textsc{b}= bias field correction, \textsc{f}= morphological filter to enhance small lesions, \textsc{i}= intensity normalization, \textsc{r}= resizing or resampling to a predefined grid, \textsc{s}= skull stripping, and \textsc{t}= transformation to a standard space.\\
$^2$ Post-processing: \textsc{fp} location based false positive reduction, \textsc{g} graph-based segmentation refinement, \textsc{Sl} remove slices prone to false positives, and \textsc{Sm} remove small segmentation results.\\
$^3$ DL indicates whether this method uses deep learning.\\
$^4$ Augmentation of training data: \textsc{d}= non-linear deforming, \textsc{h}= shearing, \textsc{m}= mirroring, \textsc{r}= rotating, \textsc{s}= scaling, \textsc{t}= translating/moving, and \textsc{Y}= generating synthetic lesions.\\
$^5$ Features used in the neural networks. Dim: 2D or 3D convolutions. Dil: dilated convolutions. BN: batch normalisation. Drop: dropout. MS: multi scale approaches (e.g. separate paths at different resolutions). LR: use of a learing rate schedule (e.g. reducing the learning rate during training). HN: hard negative mining. Ens: an ensemble of multiple networks. \\
$^\ast$ additional data from other sources was used to train this method.\\
$\dagger$ the convolutions are 2D, but the third dimension is processed within an RNN that incorporates all dimensions.\\
\end{table*}

The inter-scanner robustness was determined as follows: Appendix \ref{app:summaries} Figures \ref{app:sysu_media}--\ref{app:hadi} show the median performance of each method per metric per scanner (the line in the individual boxplots). Per metric, the standard deviation of the median values per scanner is computed. Next, methods are ranked based on those values, where a lower standard deviation indicates better inter-scanner performance. The result of this inter-scanner ranking is shown in the last column of Table~\ref{tab:results_ranking}, together with the new position of that method in the ranking. The method of \ipmibern{} achieves the highest inter-scanner rank and \sysumedia{} is just behind on the second rank. The methods of \achilles{} and \knight{} enter the top~4 of the ranking.

\begin{figure*}[!h]%
\centering%
\includegraphics[height=19cm]{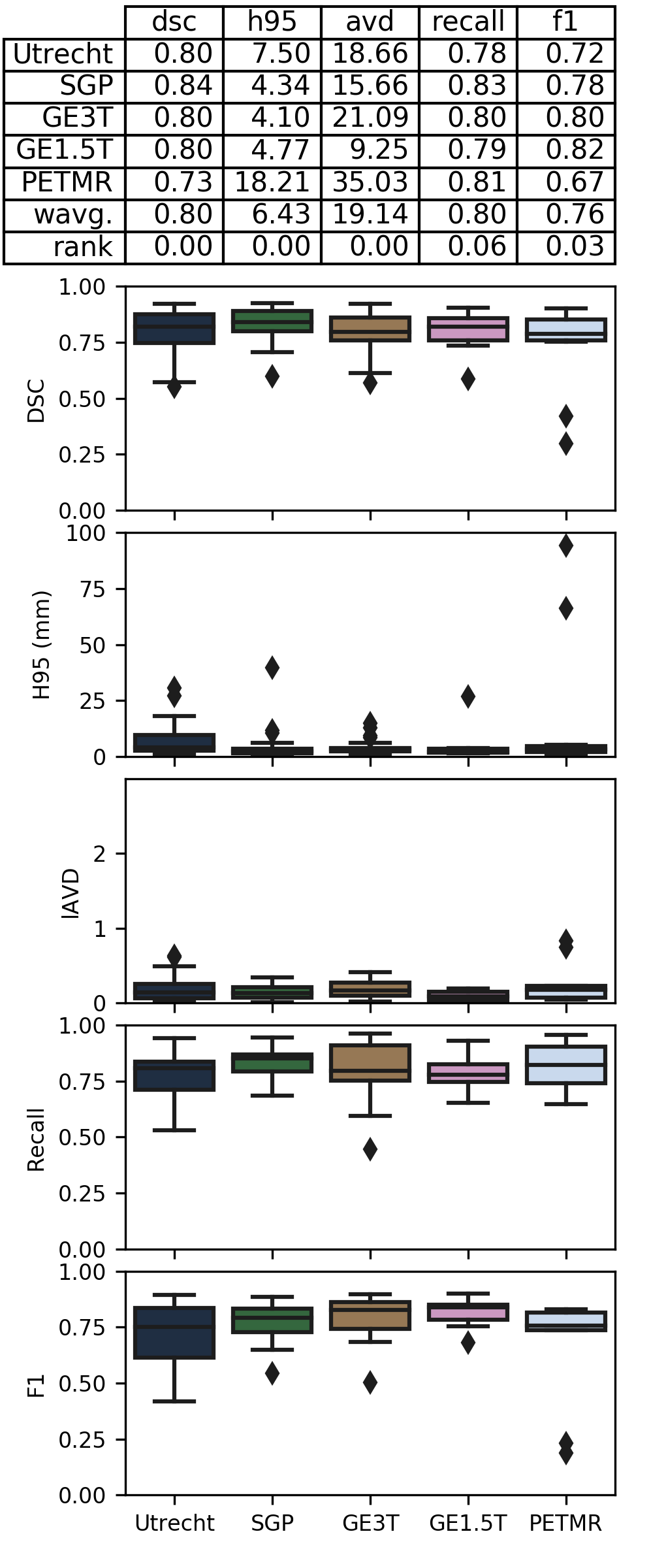}%
\includegraphics[height=19cm]{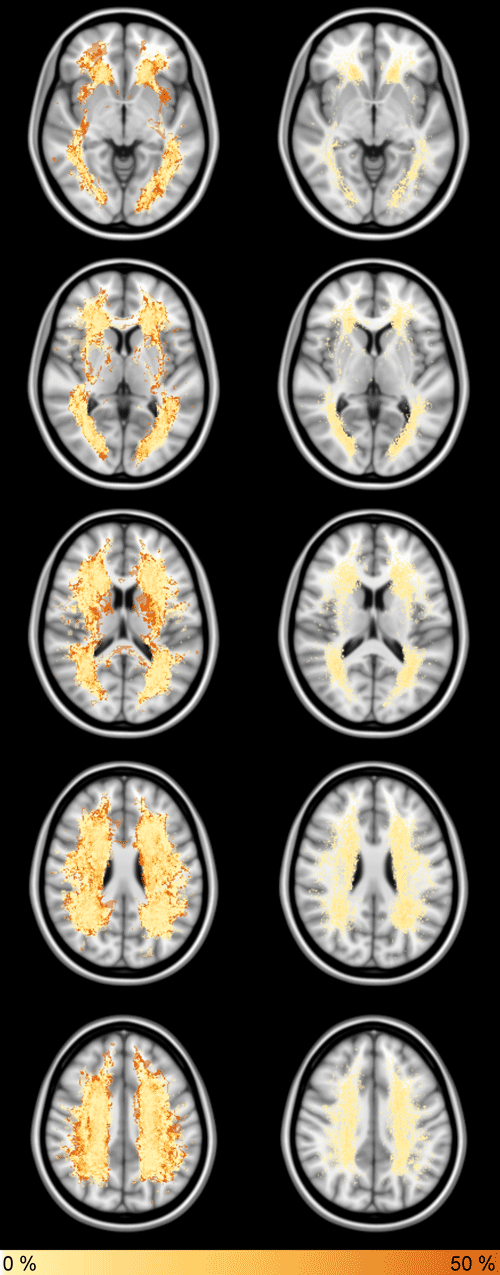}%
\caption{Detailed results of STAPLE applied on the top~4 ranking methods. The two columns on the right show the false negative rate (left) and false positive rate (right).}%
\label{app:staple4}%
\end{figure*}

STAPLE was applied on all methods and on the top~4 ranking methods, since these rank significantly higher than all other methods. The results are shown on the bottom rows of Table~\ref{tab:results_metrics} and in Appendix~\ref{app:summaries} Figures \ref{app:staple} and \ref{app:staple4}. STAPLE on all methods would rank fourth in the challenge and achieves the best H95. STAPLE on the top~4 ranking methods would rank second in the challenge and achieves the best DSC and lAVD. When re-computing the inter-scanner robustness, both STAPLE methods outperform all other methods. STAPLE compared with the top~3 methods in the inter-scanner ranking is shown separately in Table~\ref{tab:results_staple}, because the relative ranking values change when including STAPLE.

\begin{table}[!t]
\renewcommand{\arraystretch}{1.3}
\caption{The re-computed results of the inter-scanner robustness ranking when including the Simultaneous Truth And Performance Level Estimation (STAPLE) algorithm applied on all methods or on the top~4 ranking methods. STAPLE outperforms all methods and therefore the relative ranking values change. Here, STAPLE is compared to the top~3 methods in the original inter-scanner ranking in Table~\ref{tab:results_ranking}. The symbols between brackets indicate whether a team is ranked on the same position ($-$), lower ($\vee$), or higher ($\wedge$) compared to the original ranking; with the new position indicated as well.}
\label{tab:results_staple}
\begin{center}
\begin{tabular}{rlll}
\hline
\# & \textbf{Team} & \multicolumn{2}{l}{\textbf{Inter-scanner rank}} \\
\hline\hline
 & STAPLE (top~4)  &                   & 0.0152 (1) \\
 & STAPLE (all)    & 0.0390 (1)        &        \\
1 & ipmi-bern      & 0.0400 ($\vee$ 2) & 0.0402 ($\vee$ 2) \\
2 & sysu\_media    & 0.0433 ($\vee$ 3) & 0.0434 ($\vee$ 3) \\
3 & achilles       & 0.0769 ($\vee$ 4) & 0.0768 ($\vee$ 4) \\
\hline
\end{tabular}
\end{center}
\end{table}

Finally, it could be hypothesized that low ranking methods suffer from training set overfitting\cite{Recht2018} or poor generalization. This was evaluated by applying all submitted methods to the training data and comparing the performance on the training data to the performance on the test data. This analysis shows excellent correlation (R-squared: $0.94$, with $p < 0.001$), suggesting that there is no indication for overfitting of methods on the training data.

\section{Discussion}\label{sec:discussion}
We have presented a standardized assessment of automatic methods for the segmentation of white matter hyperintensities of presumed vascular origin. This assessment was performed in the context of the WMH Segmentation Challenge, hosted at the 20th International Conference on Medical Image Computing and Computer Assisted Intervention (MICCAI) in 2017, Qu\'{e}bec City, Quebec, Canada.

The manual reference standard was created in consensus by two skilled observers with extensive prior experience in WMH segmentation, which resulted in high quality WMH segmentations. Two additional observers individually segmented the sixty training images, without a consensus reading, to determine inter-observer agreement. The top-ranking methods achieve similar or superior performance as these two individual observers, which suggests that automatic methods might be able to replace individual observers in WMH segmentation. The moderate recall of the individual observers is mainly caused by not segmenting or missing small WMH. The F1 is higher than the recall, which is opposite for most automatic methods, and indicates that both O3 and O4 have hardly segmented any false positive WMH. 

The organizers have chosen not to disclose the test set, contrary to what is common in medical image analysis challenges. By keeping the test set secret, a high reliability of the results can be ensured because it obviates the possibility of (visual) self-evaluation by participants. 

The rapidly increasing popularity of (deep) neural networks as methodology of choice for analysing medical images \cite{Litjens2017} is noticeable in this challenge as well. Fourteen of the twenty submitted methods employ some form of (deep) neural networks, including all methods in the top ten. Nevertheless, the use of deep learning methodology is not a guaranteed recipe for success, since a number of low-ranking methods use it as well. 

Ensemble methods appear to do very well in this challenge. The methods of \sysumedia{} (\#~1), \nlplogix{} (\#~3), and \ktwo{} (\#~5) use an ensemble of separately trained neural networks to achieve top-ranking results. Furthermore, the STAPLE algorithm that combines all methods or the top~4 ranking methods achieves good results as well. On the inter-scanner robustness ranking, both results of the STAPLE algorithm outperform all other participating methods. Combining the results of various methods has been performed in other challenges as well, for example in the brain tumour segmentation challenge (BRATS)\cite{Menze2015}. However, in that challenge the combination of methods always outperformed all individual methods, whereas in this challenge the method of \sysumedia{} remains the winner. This seems to be mainly caused by the good performance of \sysumedia{} on the recall metric compared to the STAPLE results. Both STAPLE methods perform less well in recalling small lesions below the median size, as can be seen in Figure~\ref{fig:small_large_recall}.

The use of dropout during training is another characteristic of top-ranking methods. Random dropouts prevent units in neural networks from co-adapting too much\cite{Srivastava2014} and introduces some redundancy in the network. A larger network trained with dropout might behave like an ensemble of smaller networks; and ensemble methods also rank at the top. However, the deep learning methods trained with dropout have a considerably lower inter-scanner rank (Table~\ref{tab:results_ranking}). They drop more in the inter-scanner ranking than methods trained without dropout, suggesting that these methods might not generalize well to unseen data from unseen scanners; but instead only to unseen data from the same scanners as in the training data.

Selectively sampling WMH mimics, locations that resemble WMH but are not, or hard negative mining appears to be advantageous as well, since the three methods that apply it are amongst the top-ranking methods. When comparing the false positive maps of methods \nlplogix{} (Appendix~\ref{app:summaries} Figure~\ref{app:nlp_logix}), \nicvicorob{} (Appendix~\ref{app:summaries} Figure~\ref{app:nic-vicorob}), and \misp{} (Appendix~\ref{app:summaries} Figure~\ref{app:misp}) with that of the winner, \sysumedia{} (Appendix~\ref{app:summaries} Figure~\ref{app:sysu_media}); all three methods have less false positives (data not shown). However, this difference is not directly noticeable in any of the metrics in Table~\ref{tab:results_metrics}, so the sampling strategy might have had a minimal influence. A common location for false positive detections is the septum pellucidum, the area that separates both lateral ventricles. This can be seen in the third and fourth picture on the bottom row of Figure~\ref{fig:mni_lesion_overview}. This area appears hyperintense on FLAIR, similar to WMH, but is never part of a WMH as can be seen in the top row of Figure~\ref{fig:mni_lesion_overview}. Most top-ranking methods have no false positives in this area, whereas most lower-ranking methods do.

Implementing batch normalization, multi-scale processing, or using a learning rate schedule does not seem to influence the ranking of deep learning methods. The three methods that use dilated convolutions cluster together in the middle of the ranking, but whether that is attributable to the use of dilated convolutions or other factors is not sure.

Most deep learning methods that use 3D convolutions achieve a low ranking in the challenge. It could be that training 3D convolutional neural networks involved too many parameters, which could not be learned from the provided training data. Most FLAIR images were 2D multi-slice acquisitions (approximately $1 \times 1 \times 3$~mm voxels) with relatively few slices. Training 2D convolutional neural networks appears to work better in this case, but the methods of \cian{}, \nicvicorob{}, and \misp{} demonstrate that it was feasible to train 3D networks.

Regions with the highest false negative rates are located in regions with fewer WMH, as can be seen in the top and middle rows of Figure~\ref{fig:mni_lesion_overview}. It appears that methods have issues finding WMH of which there are fewer training examples. This holds for all methods, as can be seen in the individual maps in Appendix~\ref{app:summaries}. Furthermore, the regions with high false negative rates usually have smaller WMH, for which the recall is lower compared to larger WMH (Figure~\ref{fig:small_large_recall}). It has been noted before that smaller WMH are harder to find and the proposed solution was to develop designated methods for small WMHs\cite{Ghafoorian2016}. This has been adopted by the method of \nicvicorob{}, where a separate network reclassifies detected locations below a size of 30 voxels. Additionally, a selective sampling strategy might be used, combined with data augmentation, to provide more examples of small lesions during training. The method of \lrde{} highlights small WMH as part of the pre-processing, but does not adapt the sampling strategy. Furthermore, method developers might need to make their methods less location-sensitive: not rejecting a WMH because it is at a location with low a priori probability. This might also be a strategy to reduce the number of false positive detections. These appear to coincide with the location of true positives, suggesting that methods more easily segment a false positive at locations with high a priori probability. 

The inter-scanner robustness ranking in the last column of Table~\ref{tab:results_ranking} shows some remarkable changes in the ranking. The method of \ipmibern{} becomes first, having the best inter-scanner robustness and putting \sysumedia{} at the second place. Furthermore, the methods of \achilles{}, \knight{}, and \skkumedneuro{} rank considerably higher. Despite the somewhat moderate performance on the individual metrics, these methods generalize well to unseen scanners and have robust performance; ranking very close to the winner. The top~10 of the inter-scanner ranking shows three non-deep learning methods, whereas none is present in the final ranking. The methods of \nicvicorob{} and \lrde{} drop considerably in the inter-scanner ranking. Both methods perform less well on the images from the 3~T Philips Ingenuity (PET/MR) scanner that was not in the training data. Since only 10/110 test images originated from this scanner, it likely did not affect their overall ranking that much. The inter-scanner ranking of the \tig{} and \tignet{} methods shows a remarkable difference with the overall ranking. The \tignet{} method, a neural network trained to replicate the results of the \tig{} method, ranks close to the \tig{} method in the overall ranking. In the inter-scanner ranking, the \tignet{} method drops whereas the \tig{} method rises.

No method performs best/worst on all individual metrics. Neither on the overall rankings nor on the inter-scanner rankings in Table~\ref{tab:results_ranking}, ranking 0.0000 (overall best) nor 1.0000 (overall worst) are assigned to a method. Most room for improvement seems to be on the recall and F1 metrics. Many methods fail to achieve a good score on these, which seems to be caused by methods missing small individual lesions. Missing one or a few small lesions does not contribute to a lower DSC, H95, nor lAVD, but does have a considerable influence on the recall and F1 metrics. Recent evidence shows that the presence and shape of small WMH can be of added value to further unravel the etiology and functional impact of WMH \cite{DeBresser2018}. Furthermore, WMH location in strategic white matter tracts can explain cognitive dysfunctioning better than total WMH volume\cite{Biesbroek2017}. Hence, evaluating the recall and F1 metrics are of increasing importance for WMH segmentation methods.

Future developments in WMH segmentation might focus on improving the recall for small lesions and the inter-scanner robustness, especially on unseen data from unseen scanners. However, the current top ranking deep learning methods can already assist, or even replace, individual human observers in segmenting WMH.

After the results were presented at the MICCAI conference, a number of participants submitted an updated version of their method: \misp{}, \neuroml{}, \nihcidi{}, \sysumedia{}, and \tig{}. All methods showed an increased performance with respect to their original submission. Updated descriptions and results are available on the challenge website.

The WMH Segmentation Challenge remains open for new and updated future submissions.

\section*{Acknowledgment}
\addcontentsline{toc}{section}{Acknowledgment}
The organizers thank T.~Doeven for assisting with the manual segmentation of WMH.

H.J.~Kuijf is supported by Off Road grant 451001007 from the Netherlands Organisation for Health Research and Development (ZonMW).

S.~Andermatt was funded by the MIAC AG, Basel, Switzerland.

M.~Bento and L.~Rittner thank Hotchkiss Brain Institute and CAPES process PVE 88881.062158/2014-01.

A.~Casamitjana and V.~Vilaplana have been partially supported by the project MALEGRA TEC2016-75976-R financed by the Spanish Ministerio de Econom\'{i}a y Competitividad and the European Regional Development Fund (ERDF). A.~Casamitjana is supported by the Spanish ``Ministerio de Educacin, Cultura y Deporte'' FPU Research Fellowship.

D.~Jin and Z.~Xu were funded by the intramural research program of the National Institute of Allergy and Infectious Diseases, USA.

A.~Khademi and J.~Knight were supported in part by the Natural Science and Engineering Research Council of Canada (NSERC CGS-M) and by the Ontario Ministry of Advanced Education and Skills Development (OGS-M).

H.~Li and J.~Zhang were funded by the National Natural Science Foundation of China (No 61628212).

X.~Llad\'{o} and S.~Valverde were partially supported by TIN2014-55710-R and DPI2017-86696-R from the Ministerio de Ciencia y Tecnolog\'{i}a (Spain).

M.~Luna and S.H.~Park were supported by Basic Science Research Program through the National Research Foundation of Korea (NRF) funded by the Ministry of Education (2018R1D1A1B07044473).

R.~McKinley and R.~Wiest were funded by the Swiss Multiple Sclerosis Society.

A.~Mehrtash was supported partially by the US National Institutes of Health grants P41EB015898, Natural Sciences and Engineering Research Council (NSERC) of Canada, and the Canadian Institutes of Health Research (CIHR).

E.~Puybareau and Y.~Xu thank NVIDIA Corporation for donating a GeForce GTX 1080 Ti.

C.H.~Sudre acknowledges funding of the Alzheimer's Society Junior Research Fellowship (AS-JF-17-011).

G.~Zeng and G.~Zheng were partially supported by the Swiss National Science Foundation via project no. 205321\_163224.

F.~Barkhof is supported by the NIHR UCLH biomedical research centre.

G.J.~Biessels is supported by VICI grant 918.16.616 from the Netherlands Organisation for Scientific Research (NWO).

\bibliographystyle{IEEEtran}
\bibliography{IEEEabrv,../../../Bibliography/library}

\begin{thebibliography}{10}
\providecommand{\url}[1]{#1}
\csname url@samestyle\endcsname
\providecommand{\newblock}{\relax}
\providecommand{\bibinfo}[2]{#2}
\providecommand{\BIBentrySTDinterwordspacing}{\spaceskip=0pt\relax}
\providecommand{\BIBentryALTinterwordstretchfactor}{4}
\providecommand{\BIBentryALTinterwordspacing}{\spaceskip=\fontdimen2\font plus
\BIBentryALTinterwordstretchfactor\fontdimen3\font minus
  \fontdimen4\font\relax}
\providecommand{\BIBforeignlanguage}[2]{{%
\expandafter\ifx\csname l@#1\endcsname\relax
\typeout{** WARNING: IEEEtran.bst: No hyphenation pattern has been}%
\typeout{** loaded for the language `#1'. Using the pattern for}%
\typeout{** the default language instead.}%
\else
\language=\csname l@#1\endcsname
\fi
#2}}
\providecommand{\BIBdecl}{\relax}
\BIBdecl

\bibitem{Pantoni2010}
L.~Pantoni, ``{Cerebral small vessel disease: from pathogenesis and clinical
  characteristics to therapeutic challenges},'' \emph{The Lancet Neurology},
  vol.~9, no.~7, pp. 689--701, 2010.

\bibitem{Prins2015}
N.~D. Prins and P.~Scheltens, ``{White matter hyperintensities, cognitive
  impairment and dementia: an update},'' \emph{Nature Reviews Neurology},
  vol.~11, no.~3, pp. 157--165, 2015.

\bibitem{Wardlaw2013}
J.~M. Wardlaw, E.~E. Smith, G.~J. Biessels, C.~Cordonnier, F.~Fazekas,
  R.~Frayne, R.~I. Lindley, J.~T. O'Brien, F.~Barkhof, O.~R. Benavente, S.~E.
  Black, C.~Brayne, M.~Breteler, H.~Chabriat, C.~Decarli, F.-E. de~Leeuw,
  F.~Doubal, M.~Duering, N.~C. Fox, S.~Greenberg, V.~Hachinski, I.~Kilimann,
  V.~Mok, R.~van Oostenbrugge, L.~Pantoni, O.~Speck, B.~C.~M. Stephan,
  S.~Teipel, A.~Viswanathan, D.~Werring, C.~Chen, C.~Smith, M.~van Buchem,
  B.~Norrving, P.~B. Gorelick, and M.~Dichgans, ``{Neuroimaging standards for
  research into small vessel disease and its contribution to ageing and
  neurodegeneration.}'' \emph{The Lancet. Neurology}, vol.~12, no.~8, pp.
  822--38, 2013.

\bibitem{Biesbroek2017}
J.~M. Biesbroek, N.~A. Weaver, and G.~J. Biessels, ``{Lesion location and
  cognitive impact of cerebral small vessel disease},'' \emph{Clinical
  Science}, vol. 131, no.~8, pp. 715--728, 2017.

\bibitem{Caligiuri2015}
M.~E. Caligiuri, P.~Perrotta, A.~Augimeri, F.~Rocca, A.~Quattrone, and
  A.~Cherubini, ``{Automatic Detection of White Matter Hyperintensities in
  Healthy Aging and Pathology Using Magnetic Resonance Imaging: A Review},''
  \emph{Neuroinformatics}, vol.~13, no.~3, pp. 261--276, 2015.

\bibitem{Greenspan2016}
H.~Greenspan, B.~van Ginneken, and R.~M. Summers, ``{Guest Editorial Deep
  Learning in Medical Imaging: Overview and Future Promise of an Exciting New
  Technique},'' \emph{IEEE Transactions on Medical Imaging}, vol.~35, no.~5,
  pp. 1153--1159, may 2016.

\bibitem{Heimann2009}
T.~Heimann, B.~van Ginneken, M.~Styner, Y.~Arzhaeva, V.~Aurich, C.~Bauer,
  A.~Beck, C.~Becker, R.~Beichel, G.~Bekes, F.~Bello, G.~Binnig, H.~Bischof,
  A.~Bornik, P.~Cashman, {Ying Chi}, A.~Cordova, B.~Dawant, M.~Fidrich,
  J.~Furst, D.~Furukawa, L.~Grenacher, J.~Hornegger, D.~Kainmuller, R.~Kitney,
  H.~Kobatake, H.~Lamecker, T.~Lange, {Jeongjin Lee}, B.~Lennon, {Rui Li},
  {Senhu Li}, H.-P. Meinzer, G.~Nemeth, D.~Raicu, A.-M. Rau, E.~van Rikxoort,
  M.~Rousson, L.~Rusko, K.~Saddi, G.~Schmidt, D.~Seghers, A.~Shimizu,
  P.~Slagmolen, E.~Sorantin, G.~Soza, R.~Susomboon, J.~Waite, A.~Wimmer, and
  I.~Wolf, ``{Comparison and Evaluation of Methods for Liver Segmentation From
  CT Datasets},'' \emph{IEEE Transactions on Medical Imaging}, vol.~28, no.~8,
  pp. 1251--1265, aug 2009.

\bibitem{Murphy2011}
K.~Murphy, B.~van Ginneken, J.~M. Reinhardt, S.~Kabus, {Kai Ding}, {Xiang
  Deng}, {Kunlin Cao}, {Kaifang Du}, G.~E. Christensen, V.~Garcia,
  T.~Vercauteren, N.~Ayache, O.~Commowick, G.~Malandain, B.~Glocker,
  N.~Paragios, N.~Navab, V.~Gorbunova, J.~Sporring, M.~de~Bruijne, {Xiao Han},
  M.~P. Heinrich, J.~A. Schnabel, M.~Jenkinson, C.~Lorenz, M.~Modat, J.~R.
  McClelland, S.~Ourselin, S.~E.~A. Muenzing, M.~A. Viergever, D.~{De Nigris},
  D.~L. Collins, T.~Arbel, M.~Peroni, {Rui Li}, G.~C. Sharp,
  A.~Schmidt-Richberg, J.~Ehrhardt, R.~Werner, D.~Smeets, D.~Loeckx, {Gang
  Song}, N.~Tustison, B.~Avants, J.~C. Gee, M.~Staring, S.~Klein, B.~C. Stoel,
  M.~Urschler, M.~Werlberger, J.~Vandemeulebroucke, S.~Rit, D.~Sarrut, and
  J.~P.~W. Pluim, ``{Evaluation of Registration Methods on Thoracic CT: The
  EMPIRE10 Challenge},'' \emph{IEEE Transactions on Medical Imaging}, vol.~30,
  no.~11, pp. 1901--1920, nov 2011.

\bibitem{Wolterink2016}
J.~M. Wolterink, T.~Leiner, B.~D. de~Vos, J.-L. Coatrieux, B.~M. Kelm,
  S.~Kondo, R.~A. Salgado, R.~Shahzad, H.~Shu, M.~Snoeren, R.~A.~P. Takx, L.~J.
  van Vliet, T.~van Walsum, T.~P. Willems, G.~Yang, Y.~Zheng, M.~A. Viergever,
  and I.~I{\v{s}}gum, ``{An evaluation of automatic coronary artery calcium
  scoring methods with cardiac CT using the orCaScore framework},''
  \emph{Medical Physics}, vol.~43, no.~5, pp. 2361--2373, apr 2016.

\bibitem{Sirinukunwattana2017}
K.~Sirinukunwattana, J.~P. Pluim, H.~Chen, X.~Qi, P.-A. Heng, Y.~B. Guo, L.~Y.
  Wang, B.~J. Matuszewski, E.~Bruni, U.~Sanchez, A.~B{\"{o}}hm, O.~Ronneberger,
  B.~B. Cheikh, D.~Racoceanu, P.~Kainz, M.~Pfeiffer, M.~Urschler, D.~R. Snead,
  and N.~M. Rajpoot, ``{Gland segmentation in colon histology images: The glas
  challenge contest},'' \emph{Medical Image Analysis}, vol.~35, pp. 489--502,
  2017.

\bibitem{Styner2008}
\BIBentryALTinterwordspacing
M.~Styner, J.~Lee, B.~Chin, M.~Chin, O.~Commowick, H.~Tran, S.~Markovic-Plese,
  V.~Jewells, and S.~Warfield, ``{3D Segmentation in the Clinic: A Grand
  Challenge II: MS lesion segmentation},'' \emph{The MIDAS Journal}, 2008.
  [Online]. Available:
  \url{https://www.midasjournal.org/browse/publication/638/2}
\BIBentrySTDinterwordspacing

\bibitem{Commowick2018}
\BIBentryALTinterwordspacing
O.~Commowick, A.~Istace, M.~Kain, B.~Laurent, F.~Leray, M.~Simon, S.~C. Pop,
  P.~Girard, R.~Am{\'{e}}li, J.-C. Ferr{\'{e}}, A.~Kerbrat, T.~Tourdias,
  F.~Cervenansky, T.~Glatard, J.~Beaumont, S.~Doyle, F.~Forbes, J.~Knight,
  A.~Khademi, A.~Mahbod, C.~Wang, R.~McKinley, F.~Wagner, J.~Muschelli,
  E.~Sweeney, E.~Roura, X.~Llad{\'{o}}, M.~M. Santos, W.~P. Santos, A.~G.
  Silva-Filho, X.~Tomas-Fernandez, H.~Urien, I.~Bloch, S.~Valverde, M.~Cabezas,
  F.~J. Vera-Olmos, N.~Malpica, C.~Guttmann, S.~Vukusic, G.~Edan, M.~Dojat,
  M.~Styner, S.~K. Warfield, F.~Cotton, and C.~Barillot, ``{Objective
  Evaluation of Multiple Sclerosis Lesion Segmentation using a Data Management
  and Processing Infrastructure},'' \emph{Scientific Reports}, vol.~8, no.~1,
  p. 13650, 2018. [Online]. Available:
  \url{http://www.nature.com/articles/s41598-018-31911-7}
\BIBentrySTDinterwordspacing

\bibitem{Menze2015}
B.~H. Menze, A.~Jakab, S.~Bauer, J.~Kalpathy-Cramer, K.~Farahani, J.~Kirby,
  Y.~Burren, N.~Porz, J.~Slotboom, R.~Wiest, L.~Lanczi, E.~Gerstner, M.-A.
  Weber, T.~Arbel, B.~B. Avants, N.~Ayache, P.~Buendia, D.~L. Collins,
  N.~Cordier, J.~J. Corso, A.~Criminisi, T.~Das, H.~Delingette,
  {\c{C}}.~Demiralp, C.~R. Durst, M.~Dojat, S.~Doyle, J.~Festa, F.~Forbes,
  E.~Geremia, B.~Glocker, P.~Golland, X.~Guo, A.~Hamamci, K.~M. Iftekharuddin,
  R.~Jena, N.~M. John, E.~Konukoglu, D.~Lashkari, J.~A. Mariz, R.~Meier,
  S.~Pereira, D.~Precup, S.~J. Price, T.~R. Raviv, S.~M.~S. Reza, M.~Ryan,
  D.~Sarikaya, L.~Schwartz, H.-C. Shin, J.~Shotton, C.~A. Silva, N.~Sousa,
  N.~K. Subbanna, G.~Szekely, T.~J. Taylor, O.~M. Thomas, N.~J. Tustison,
  G.~Unal, F.~Vasseur, M.~Wintermark, D.~H. Ye, L.~Zhao, B.~Zhao, D.~Zikic,
  M.~Prastawa, M.~Reyes, and K.~{Van Leemput}, ``{The Multimodal Brain Tumor
  Image Segmentation Benchmark (BRATS).}'' \emph{IEEE transactions on medical
  imaging}, vol.~34, no.~10, pp. 1993--2024, 2015.

\bibitem{Mendrik2015}
A.~Mendrik, K.~Vincken, H.~Kuijf, M.~Breeuwer, W.~Bouvy, J.~de~Bresser,
  A.~Alansary, M.~de~Bruijne, A.~Carass, A.~El-Baz, A.~Jog, R.~Katyal, A.~Khan,
  F.~van~der Lijn, Q.~Mahmood, R.~Mukherjee, A.~van Opbroek, S.~Paneri,
  S.~Pereira, M.~Persson, M.~Rajchl, D.~Sarikaya, {\"{O}}.~Smedby, C.~Silva,
  H.~Vrooman, S.~Vyas, C.~Wang, L.~Zhao, G.~Biessels, and M.~Viergever,
  ``{MRBrainS Challenge: Online Evaluation Framework for Brain Image
  Segmentation in 3T MRI Scans},'' \emph{Computational Intelligence and
  Neuroscience}, vol. 2015, 2015.

\bibitem{Boomsma2017}
J.~M.~F. Boomsma, L.~G. Exalto, F.~Barkhof, E.~van~den Berg, J.~de~Bresser,
  R.~Heinen, H.~L. Koek, N.~D. Prins, P.~Scheltens, H.~C. Weinstein, W.~M.
  van~der Flier, and G.~J. Biessels, ``{Vascular Cognitive Impairment in a
  Memory Clinic Population: Rationale and Design of the "Utrecht-Amsterdam
  Clinical Features and Prognosis in Vascular Cognitive Impairment" (TRACE-VCI)
  Study.}'' \emph{JMIR research protocols}, vol.~6, no.~4, p. e60, 2017.

\bibitem{VanVeluw2015b}
S.~J. {Van Veluw}, S.~Hilal, H.~J. Kuijf, M.~K. Ikram, X.~Xin, T.~{Boon Yeow},
  N.~Venketasubramanian, G.~J. Biessels, and C.~Chen, ``{Cortical microinfarcts
  on 3T MRI: Clinical correlates in memory-clinic patients},''
  \emph{Alzheimer's {\&} Dementia}, vol.~11, no.~12, pp. 1500--1509, 2015.

\bibitem{Ashburner2000}
J.~Ashburner and K.~J. Friston, ``{Voxel-based Morphometry--The methods},''
  \emph{NeuroImage}, vol.~11, no. 6 Pt 1, pp. 805--21, 2000.

\bibitem{Klein2010}
S.~Klein, M.~Staring, K.~Murphy, M.~A. Viergever, and J.~P.~W. Pluim,
  ``{Elastix: a Toolbox for Intensity-Based Medical Image Registration.}''
  \emph{IEEE transactions on medical imaging}, vol.~29, no.~1, pp. 196--205,
  2010.

\bibitem{Merkel2014a}
D.~Merkel, ``{Docker: lightweight Linux containers for consistent development
  and deployment},'' \emph{Linux Journal}, vol. 2014, no. 239, p.~5, 2014.

\bibitem{Li2017}
W.~Li, G.~Wang, L.~Fidon, S.~Ourselin, M.~J. Cardoso, and T.~Vercauteren, ``{On
  the compactness, efficiency, and representation of 3D convolutional networks:
  Brain parcellation as a pretext task},'' in \emph{Information Processing in
  Medical Imaging. IPMI 2017. Lecture Notes in Computer Science}, vol. 10265
  LNCS.\hskip 1em plus 0.5em minus 0.4em\relax Springer, Cham, 2017, pp.
  348--360.

\bibitem{Chen2017}
L.-C. Chen, G.~Papandreou, F.~Schroff, and H.~Adam, ``{Rethinking Atrous
  Convolution for Semantic Image Segmentation},'' arXiv:1706.05587, Tech. Rep.,
  jun 2017.

\bibitem{Georgiou2017}
\BIBentryALTinterwordspacing
A.~Georgiou, ``{WMH segmentation challenge MICCAI 2017: Team name -
  Achilles},'' 2017. [Online]. Available:
  \url{http://wmh.isi.uu.nl/results/achilles/}
\BIBentrySTDinterwordspacing

\bibitem{Andermatt2016a}
S.~Andermatt, S.~Pezold, and P.~Cattin, ``{Multi-dimensional Gated Recurrent
  Units for the Segmentation of Biomedical 3D-Data},'' in \emph{Deep Learning
  and Data Labeling for Medical Applications}, G.~Carneiro, D.~Mateus,
  L.~Peter, A.~Bradley, J.~M. R.~S. Tavares, V.~Belagiannis, J.~P. Papa, J.~C.
  Nascimento, M.~Loog, Z.~Lu, J.~S. Cardoso, and J.~Cornebise, Eds.\hskip 1em
  plus 0.5em minus 0.4em\relax Springer, Cham, 2016, pp. 142--151.

\bibitem{Andermatt2017}
\BIBentryALTinterwordspacing
------, ``{Multi-dimensional Gated Recurrent Units for the Segmentation of
  White Matter Hyperintensites},'' 2017. [Online]. Available:
  \url{http://wmh.isi.uu.nl/results/cian/}
\BIBentrySTDinterwordspacing

\bibitem{Andermatt2018}
S.~Andermatt, S.~Pezold, and P.~C. Cattin, ``{Automated Segmentation of
  Multiple Sclerosis Lesions using Multi-Dimensional Gated Recurrent Units},''
  in \emph{Brainlesion: Glioma, Multiple Sclerosis, Stroke and Traumatic Brain
  Injuries}, A.~Crimi, S.~Bakas, H.~Kuijf, B.~Menze, and M.~Reyes, Eds.\hskip
  1em plus 0.5em minus 0.4em\relax Springer, Cham, 2018, pp. 31--42.

\bibitem{Mahmood2017}
\BIBentryALTinterwordspacing
Q.~Mahmood and A.~Basit, ``{Automated Segmentation of White Matter
  Hyperintensities in Multi-modal MRI Images Using Random Forests},'' 2017.
  [Online]. Available: \url{http://wmh.isi.uu.nl/results/hadi/}
\BIBentrySTDinterwordspacing

\bibitem{Zeng2017}
\BIBentryALTinterwordspacing
G.~Zeng and G.~Zheng, ``{Deeply Supervised Multi-Scale Fully Convolutional
  Networks for Segmentation of White Matter Hyperintensities},'' 2017.
  [Online]. Available: \url{http://wmh.isi.uu.nl/results/ipmi-bern/}
\BIBentrySTDinterwordspacing

\bibitem{Ronneberger2015}
O.~Ronneberger, {Philipp Fischer}, and T.~Brox, ``{U-Net: Convolutional
  Networks for Biomedical Image Segmentation},'' in \emph{Medical Image
  Computing and Computer-Assisted Intervention – MICCAI 2015. Lecture Notes
  in Computer Science}.\hskip 1em plus 0.5em minus 0.4em\relax Springer, Cham,
  2015, vol. 9351, pp. 234--241.

\bibitem{Mehrtash2017}
\BIBentryALTinterwordspacing
A.~Mehrtash and M.~Ghafoorian, ``{Simurgh Team Method Description},'' 2017.
  [Online]. Available: \url{http://wmh.isi.uu.nl/results/k2/}
\BIBentrySTDinterwordspacing

\bibitem{Fonov2011}
V.~Fonov, A.~C. Evans, K.~Botteron, C.~R. Almli, R.~C. McKinstry, and D.~L.
  Collins, ``{Unbiased average age-appropriate atlases for pediatric
  studies.}'' \emph{NeuroImage}, vol.~54, no.~1, pp. 313--27, 2011.

\bibitem{Knight2017}
\BIBentryALTinterwordspacing
J.~Knight, G.~Taylor, and A.~Khademi, ``{Voxel-Wise Logistic Regression for
  White Matter Hyperintensity Segmentation in FLAIR MRI},'' 2017. [Online].
  Available: \url{http://wmh.isi.uu.nl/results/knight/}
\BIBentrySTDinterwordspacing

\bibitem{Knight2018a}
------, ``{Voxel-Wise Logistic Regression and Leave-One-Source-Out Cross
  Validation for White Matter Hyperintensity Segmentation},'' \emph{Magnetic
  Resonance Imaging}, vol.~54, pp. 119--136, 2018.

\bibitem{Simonyan2015}
K.~Simonyan and A.~Zisserman, ``{Very Deep Convolutional Networks for
  Large-Scale Image Recognition},'' \emph{International Conference on Learning
  Representations (ICLR)}, pp. 1--14, 2015.

\bibitem{Maninis2016}
K.-K. Maninis, J.~Pont-Tuset, P.~Arbel{\'{a}}ez, and L.~{Van Gool}, ``{Deep
  Retinal Image Understanding},'' in \emph{Medical Image Computing and
  Computer-Assisted Intervention – MICCAI 2016. MICCAI 2016. Lecture Notes in
  Computer Science}, S.~Ourselin, L.~Joskowicz, M.~Sabuncu, G.~Unal, and
  W.~Wells, Eds.\hskip 1em plus 0.5em minus 0.4em\relax Springer, Cham, 2016.

\bibitem{Shelhamer2017}
E.~Shelhamer, J.~Long, and T.~Darrell, ``{Fully Convolutional Networks for
  Semantic Segmentation},'' \emph{IEEE Transactions on Pattern Analysis and
  Machine Intelligence}, vol.~39, no.~4, pp. 640--651, 2017.

\bibitem{Xu2017}
\BIBentryALTinterwordspacing
Y.~Xu, T.~G{\'{e}}raud, {\'{E}}.~Puybareau, I.~Bloch, and J.~Chazalon, ``{White
  Matter Hyperintensities Segmentation Using Fully Convolutional Network and
  Transfer Learning},'' 2017. [Online]. Available:
  \url{http://wmh.isi.uu.nl/results/lrde/}
\BIBentrySTDinterwordspacing

\bibitem{Xu2018}
------, ``{White Matter Hyperintensities Segmentation in a Few Seconds Using
  Fully Convolutional Network and Transfer Learning},'' in \emph{Brainlesion:
  Glioma, Multiple Sclerosis, Stroke and Traumatic Brain Injuries. BrainLes
  2017. Lecture Notes in Computer Science}, A.~Crimi, S.~Bakas, H.~Kuijf,
  B.~Menze, and M.~Reyes, Eds.\hskip 1em plus 0.5em minus 0.4em\relax Springer,
  Cham, 2018, pp. 501--514.

\bibitem{He2015}
K.~He, X.~Zhang, S.~Ren, and J.~Sun, ``{Deep Residual Learning for Image
  Recognition},'' arXiv:1512.03385v1, Tech. Rep., dec 2015.

\bibitem{Luna2017}
\BIBentryALTinterwordspacing
M.~Luna and S.~H. Park, ``{3D Convolutional Neural Network with Skip
  Connections for WMH Segmentation},'' 2017. [Online]. Available:
  \url{http://wmh.isi.uu.nl/results/misp/}
\BIBentrySTDinterwordspacing

\bibitem{Kamnitsas2017}
K.~Kamnitsas, C.~Ledig, V.~F. Newcombe, J.~P. Simpson, A.~D. Kane, D.~K. Menon,
  D.~Rueckert, and B.~Glocker, ``{Efficient multi-scale 3D CNN with fully
  connected CRF for accurate brain lesion segmentation},'' \emph{Medical Image
  Analysis}, vol.~36, pp. 61--78, 2017.

\bibitem{Safiullin2017}
\BIBentryALTinterwordspacing
A.~Safiullin, ``{NeuroML team: Brief description of the solution},'' 2017.
  [Online]. Available: \url{http://wmh.isi.uu.nl/results/neuro-ml/}
\BIBentrySTDinterwordspacing

\bibitem{Valverde2017a}
S.~Valverde, M.~Cabezas, E.~Roura, S.~Gonz{\'{a}}lez-Vill{\`{a}}, D.~Pareto,
  J.~C. Vilanova, L.~Rami{\'{o}}-Torrent{\`{a}}, {\`{A}}.~Rovira, A.~Oliver,
  and X.~Llad{\'{o}}, ``{Improving automated multiple sclerosis lesion
  segmentation with a cascaded 3D convolutional neural network approach},''
  \emph{NeuroImage}, vol. 155, pp. 159--168, 2017.

\bibitem{Valverde2017}
\BIBentryALTinterwordspacing
S.~Valverde, M.~Cabezas, J.~Bernal, K.~Kushibar, S.~Gonz{\'{a}}lez-Vill{\`{a}},
  M.~Salem, J.~Salvi, A.~Oliver, and X.~Llad{\'{o}}, ``{White matter
  hyperintensities segmentation using a cascade of three convolutional neural
  networks},'' 2017. [Online]. Available:
  \url{http://wmh.isi.uu.nl/results/nic-vicorob/}
\BIBentrySTDinterwordspacing

\bibitem{Jin2017}
\BIBentryALTinterwordspacing
D.~Jin, ``{WMH Segmentation Method Description - NIH{\_}CIDI},'' 2017.
  [Online]. Available: \url{http://wmh.isi.uu.nl/results/nih{\_}cidi/}
\BIBentrySTDinterwordspacing

\bibitem{Dadar2017a}
M.~Dadar, J.~Maranzano, K.~Misquitta, C.~J. Anor, V.~S. Fonov, M.~C. Tartaglia,
  O.~T. Carmichael, C.~Decarli, and D.~L. Collins, ``{Performance comparison of
  10 different classification techniques in segmenting white matter
  hyperintensities in aging},'' \emph{NeuroImage}, vol. 157, pp. 233--249,
  2017.

\bibitem{Dadar2017b}
M.~Dadar, T.~A. Pascoal, S.~Manitsirikul, K.~Misquitta, V.~S. Fonov, M.~C.
  Tartaglia, J.~Breitner, P.~Rosa-Neto, O.~T. Carmichael, C.~Decarli, and D.~L.
  Collins, ``{Validation of a Regression Technique for Segmentation of White
  Matter Hyperintensities in Alzheimer's Disease},'' \emph{IEEE Transactions on
  Medical Imaging}, vol.~36, no.~8, pp. 1758--1768, 2017.

\bibitem{Dadar2017}
\BIBentryALTinterwordspacing
M.~Dadar, V.~S. Fonov, and D.~L. Collins, ``{Automatic Multi-Modality
  Segmentation of White Matter Hyperintensities Using a Random Forests
  Classifier},'' 2017. [Online]. Available:
  \url{http://wmh.isi.uu.nl/results/nist/}
\BIBentrySTDinterwordspacing

\bibitem{Ghafoorian2017}
M.~Ghafoorian, N.~Karssemeijer, T.~Heskes, I.~W.~M. van Uden, C.~I. Sanchez,
  G.~Litjens, F.-E. de~Leeuw, B.~van Ginneken, E.~Marchiori, and B.~Platel,
  ``{Location Sensitive Deep Convolutional Neural Networks for Segmentation of
  White Matter Hyperintensities},'' \emph{Scientific Reports}, vol.~7, no.~1,
  p. 5110, 2017.

\bibitem{Berseth2017}
\BIBentryALTinterwordspacing
M.~Berseth, ``{WMH Segmentation Challenge, MICCAI 2017},'' 2017. [Online].
  Available: \url{http://wmh.isi.uu.nl/results/nlp{\_}logix/}
\BIBentrySTDinterwordspacing

\bibitem{Yu2016}
F.~Yu and V.~Koltun, ``{Multi-Scale Context Aggregation by Dilated
  Convolutions},'' in \emph{6th International Conference on Learning
  Representations}, 2016.

\bibitem{Huang2017}
G.~Huang, Z.~Liu, L.~van~der Maaten, and K.~Q. Weinberger, ``{Densely Connected
  Convolutional Networks},'' in \emph{Computer Vision and Pattern Recognition
  (CVPR), 2017 IEEE Conference on}, 2017.

\bibitem{McKinley2017}
\BIBentryALTinterwordspacing
R.~McKinley, A.~Jungo, R.~Wiest, and M.~Reyes, ``{Pooling-free fully
  convolutional networks with dense skip connections for semantic segmentation,
  with application to segmentation of white matter lesions},'' 2017. [Online].
  Available: \url{http://wmh.isi.uu.nl/results/scan/}
\BIBentrySTDinterwordspacing

\bibitem{Park2017}
\BIBentryALTinterwordspacing
B.-y. Park, M.~J. Lee, and H.~Park, ``{WMH segmentation challenge at MICCAI
  2017: Brief description of the method},'' 2017. [Online]. Available:
  \url{http://wmh.isi.uu.nl/results/skkumedneuro/}
\BIBentrySTDinterwordspacing

\bibitem{Li2017a}
\BIBentryALTinterwordspacing
H.~Li, G.~Jiang, L.~Zhao, R.~Wang, J.~Zhang, and W.-S. Zheng, ``{Automatic
  White Matter Hyperintensity Segmentation via Two-channel U-Net},'' 2017.
  [Online]. Available: \url{http://wmh.isi.uu.nl/results/sysu{\_}media/}
\BIBentrySTDinterwordspacing

\bibitem{Li2018}
H.~Li, G.~Jiang, J.~Zhang, R.~Wang, Z.~Wang, W.-S. Zheng, and B.~Menze,
  ``{Fully convolutional network ensembles for white matter hyperintensities
  segmentation in MR images},'' \emph{NeuroImage}, vol. 183, pp. 650--665,
  2018.

\bibitem{Bento2017}
\BIBentryALTinterwordspacing
M.~Bento, R.~de~Souza, R.~Lotufo, R.~Fraynea, and L.~Rittner, ``{WMH
  Segmentation Challenge: a Texture-based Classification Approach (ID:
  textclass)},'' 2017. [Online]. Available:
  \url{http://wmh.isi.uu.nl/results/text{\_}class/}
\BIBentrySTDinterwordspacing

\bibitem{Bento2018}
M.~Bento, R.~de~Souza, R.~Lotufo, R.~Frayne, and L.~Rittner, ``{WMH
  Segmentation Challenge: A Texture-Based Classification Approach},'' in
  \emph{Brainlesion: Glioma, Multiple Sclerosis, Stroke and Traumatic Brain
  Injuries. BrainLes 2017. Lecture Notes in Computer Science}, A.~Crimi,
  S.~Bakas, H.~Kuijf, B.~Menze, and M.~Reyes, Eds.\hskip 1em plus 0.5em minus
  0.4em\relax Springer, Cham, 2018, pp. 489--500.

\bibitem{Sudre2015}
C.~H. Sudre, M.~J. Cardoso, W.~H. Bouvy, G.~J. Biessels, J.~Barnes, and
  S.~Ourselin, ``{Bayesian Model Selection for Pathological Neuroimaging Data
  Applied to White Matter Lesion Segmentation},'' \emph{IEEE Transactions on
  Medical Imaging}, vol.~34, no.~10, pp. 2079--2102, 2015.

\bibitem{Sudre2017}
\BIBentryALTinterwordspacing
C.~Sudre, ``{Team TIG - WMH Challenge},'' 2017. [Online]. Available:
  \url{http://wmh.isi.uu.nl/results/tig/}
\BIBentrySTDinterwordspacing

\bibitem{Sudre2017a}
\BIBentryALTinterwordspacing
------, ``{TIGNet - WMH Challenge},'' 2017. [Online]. Available:
  \url{http://wmh.isi.uu.nl/results/tignet/}
\BIBentrySTDinterwordspacing

\bibitem{Milletari2016}
F.~Milletari, N.~Navab, and S.-A. Ahmadi, ``{V-Net: Fully Convolutional Neural
  Networks for Volumetric Medical Image Segmentation},'' in \emph{Fourth
  International Conference on 3D Vision (3DV)}.\hskip 1em plus 0.5em minus
  0.4em\relax IEEE, 2016, pp. 565--571.

\bibitem{Casamitjana2017}
\BIBentryALTinterwordspacing
A.~Casamitjana, M.~Combalia, I.~S{\'{a}}nchez, and V.~Vilaplana, ``{Augmented
  V-Net for White Matter Hyperintensities segmentation},'' 2017. [Online].
  Available: \url{http://wmh.isi.uu.nl/results/upc{\_}dlmi/}
\BIBentrySTDinterwordspacing

\bibitem{Warfield2004}
S.~K. Warfield, K.~H. Zou, and W.~M. Wells, ``{Simultaneous truth and
  performance level estimation (STAPLE): An algorithm for the validation of
  image segmentation},'' \emph{IEEE Transactions on Medical Imaging}, vol.~23,
  no.~7, pp. 903--921, jul 2004.

\bibitem{Welch1947}
B.~L. Welch, ``{The generalization of `Student's' problem when several
  different population variances are involved},'' \emph{Biometrika}, vol.~34,
  no. 1-2, pp. 28--35, 1947.

\bibitem{Recht2018}
\BIBentryALTinterwordspacing
B.~Recht, R.~Roelofs, L.~Schmidt, and V.~Shankar, ``{Do CIFAR-10 Classifiers
  Generalize to CIFAR-10?}'' Tech. Rep., jun 2018. [Online]. Available:
  \url{http://arxiv.org/abs/1806.00451}
\BIBentrySTDinterwordspacing

\bibitem{Litjens2017}
G.~Litjens, T.~Kooi, B.~E. Bejnordi, A.~Arindra, A.~Setio, F.~Ciompi,
  M.~Ghafoorian, J.~A. W.~M. {Van Der Laak}, B.~{Van Ginneken}, and C.~I.
  S{\'{a}}nchez, ``{A survey on deep learning in medical image analysis},''
  \emph{Medical Image Analysis}, vol.~42, pp. 60--88, dec 2017.

\bibitem{Srivastava2014}
\BIBentryALTinterwordspacing
N.~Srivastava, G.~Hinton, A.~Krizhevsky, I.~Sutskever, and R.~Salakhutdinov,
  ``{Dropout: A Simple Way to Prevent Neural Networks from Overfitting},''
  \emph{Journal of Machine Learning Research}, vol.~15, pp. 1929--1958, 2014.
  [Online]. Available: \url{http://jmlr.org/papers/v15/srivastava14a.html}
\BIBentrySTDinterwordspacing

\bibitem{Ghafoorian2016}
M.~Ghafoorian, N.~Karssemeijer, I.~W.~M. van Uden, F.-E. de~Leeuw, T.~Heskes,
  E.~Marchiori, and B.~Platel, ``{Automated detection of white matter
  hyperintensities of all sizes in cerebral small vessel disease},''
  \emph{Medical Physics}, vol.~43, no.~12, pp. 6246--6258, 2016.

\bibitem{DeBresser2018}
J.~de~Bresser, H.~J. Kuijf, K.~Zaanen, M.~A. Viergever, J.~Hendrikse, and G.~J.
  Biessels, ``{White matter hyperintensity shape and location feature analysis
  on brain MRI; proof of principle study in patients with diabetes},''
  \emph{Scientific Reports}, vol.~8, no.~1, p. 1893, dec 2018.

\end{thebibliography}

\clearpage\newpage\onecolumn\pagenumbering{gobble}
\appendices

\section{Example figures}
\label{app:images}
\begin{figure*}[!ht]%
\centering%
\includegraphics[width=3.5cm]{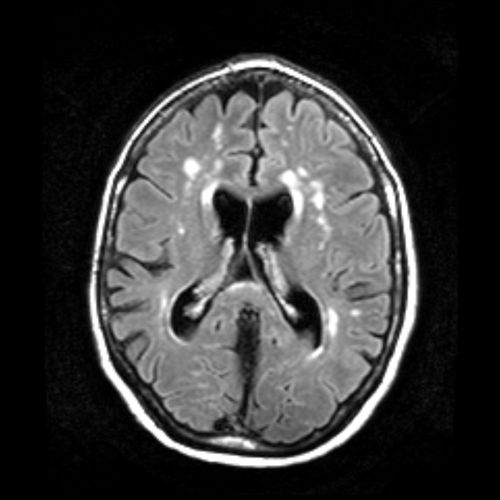}%
\includegraphics[width=3.5cm]{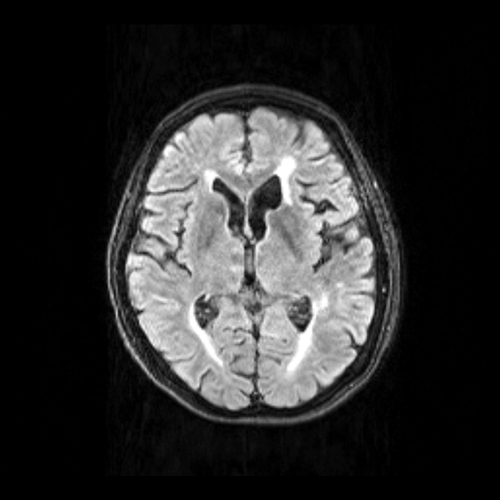}%
\includegraphics[width=3.5cm]{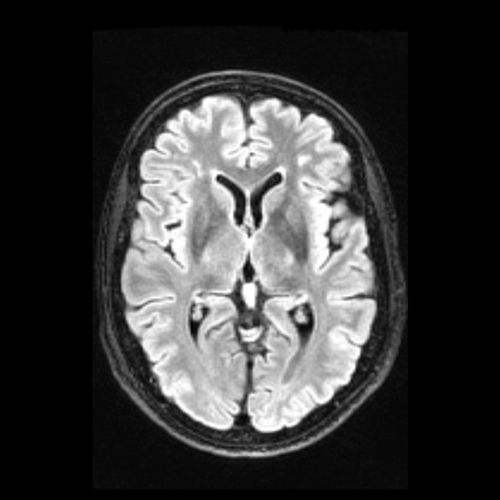}%
\includegraphics[width=3.5cm]{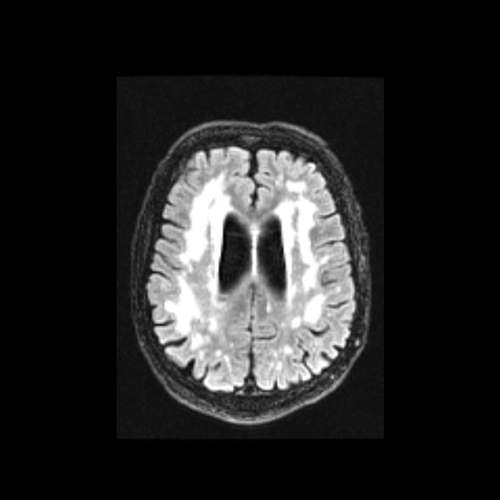}%
\includegraphics[width=3.5cm]{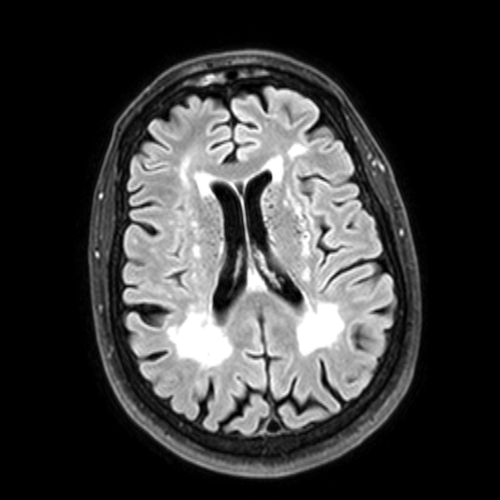}%
\caption{An example FLAIR image of each scanner used in this challenge. From left to right: UMC Utrecht 3~T Philips Achieva, NUHS Singpore 3~T Siemens TrioTim, VU Amsterdam 3~T GE Signa HDxt, VU Amsterdam 1.5~T GE Signa HDxt, and the VU Amsterdam 3~T Philips Ingenuity (PET/MR).}%
\label{app:extra_figures}%
\end{figure*}

Figure~\ref{app:extra_figures} shows an example FLAIR image of each scanner used in this challenge.

\clearpage\newpage
\section{Absolute of the percentage volume difference (AVD) metric}
Table~\ref{tab:results_avd} show the results on the original absolute of the percentage volume difference (AVD) metric. If the final ranking is computed using the AVD, instead of the log-transformed volume difference (lAVD), only minor differences occur in the ranking. The method of \cian{} has the best lAVD score, whereas \nlplogix{} has the best AVD score. The six methods that swap position ranked already relatively close to each other. This also highlights the benefits of a relative ranking method, since the relative rankings are much closer to each other than the absolute positions in the ranking.

\begin{table*}[!ht]
\renewcommand{\arraystretch}{1.3}
\caption{Mean performance and 95~\% confidence intervals of each participating method on the absolute of the percentage volume difference (AVD) metric. Bold indicates that a method has the best score. Methods are sorted based on their final ranking in case AVD would have been used instead of the log-transformed volume difference (lAVD). The symbols between brackets indicate whether a team is ranked on the same position ($-$), lower ($\vee$), or higher ($\wedge$) compared to the AVD-based ranking; with the new position indicated as well. The bottom rows include the results of the Simultaneous Truth And Performance Level Estimation (STAPLE) algorithm applied on all methods or on the top~4 ranking methods and observers O3 and O4, together with the ranking if these results would have been included. Note that O3 and O4 segmented the sixty training images. }
\label{tab:results_avd}
 \begin{center}
\begin{tabular}{rllllll}
\hline
\textbf{\#} & \textbf{Team} & \textbf{AVD (\%)} & \textbf{Ranking AVD} & \textbf{Ranking lAVD} \\
\hline\hline
 1 & sysu\_media  &  21.88 (18.53 -   25.90)  & 0.0076 & 0.0068 ($-$ 1)      \\
 2 & cian         &  21.72 (17.62 -   26.32)  & 0.0366 & 0.0357 ($-$ 2)      \\
 3 & nlp\_logix   &  \textbf{18.37 (15.39 -   21.53)} & 0.0485 & 0.0512 ($-$ 3)      \\
 4 & nic-vicorob  &  28.54 (22.13 -   36.67)  & 0.0735 & 0.0767 ($-$ 4)      \\
 5 & k2           &  19.08 (15.63 -   22.67)  & 0.1368 & 0.1420 ($-$ 5)      \\
 6 & lrde         &  21.71 (17.96 -   25.43)  & 0.1635 & 0.1746 ($\vee$ 7)   \\
 7 & misp         &  21.36 (16.80 -   26.43)  & 0.1659 & 0.1719 ($\wedge$ 6) \\
 8 & ipmi-bern    &  19.92 (16.11 -   24.18)  & 0.2498 & 0.2527 ($\vee$ 9)   \\
 9 & nih\_cidi    & 196.38 (21.96 -  536.97)  & 0.2697 & 0.2348 ($\wedge$ 8) \\
10 & scan         &  34.67 (25.45 -   46.37)  & 0.2762 & 0.2810 ($-$ 10)     \\
11 & achilles     &  24.41 (20.00 -   29.83)  & 0.2962 & 0.3032 ($-$ 11)     \\
12 & skkumedneuro &  58.54 (30.47 -  105.38)  & 0.3492 & 0.3588 ($-$ 12)     \\
13 & tignet       &  86.22 (65.05 -  111.15)  & 0.3802 & 0.3982 ($-$ 13)     \\
14 & tig          &  34.34 (29.27 -   39.23)  & 0.3858 & 0.4031 ($-$ 14)     \\
15 & knight       &  39.99 (29.11 -   54.15)  & 0.4159 & 0.4269 ($-$ 15)     \\
16 & upc\_dlmi    & 208.49 (101.36 - 366.18)  & 0.4337 & 0.4296 ($-$ 16)     \\
17 & nist         & 109.98 (70.39 -  159.42)  & 0.4747 & 0.4917 ($-$ 17)     \\
18 & text\_class  & 146.64 (92.07 -  215.39)  & 0.5725 & 0.5830 ($\vee$ 19)  \\
19 & neuro.ml     & 614.05 (330.65 -  954.28) & 0.5960 & 0.5349 ($\wedge$ 18)\\
20 & hadi         & 828.61 (517.02 - 1205.50) & 0.8886 & 0.8886 ($-$ 20)     \\
\hline
4 & STAPLE (all)   &  54.87 (33.54 - 85.10)  \\
2 & STAPLE (top 4) &  19.14 (15.53 - 23.25)  \\
\hline
5 & O3 & \textbf{17.27 (13.17 - 22.15)}  \\
4 & O4 & 18.78 (14.14 - 24.48) \\
\hline
\end{tabular}
\end{center}
\end{table*}

\clearpage\newpage
\section{Summaries of results}\label{app:summaries}
Detailed summaries of all results per participant are given in the following appendices. All figures show the performance of each method on the five scanners described in Section~\ref{sec:mri} for the following five criteria: (1) the Dice Similarity Coefficient (DSC), (2) a modified Hausdorff distance (95th percentile; H95), (3) the absolute log-transformed volume difference (lAVD), (4) the sensitivity for detecting individual lesions (recall), and (5) F1-score for individual lesions (F1). Next to that are two columns that show spatial maps of the false negative rate (left) and false positive rate (right).

The figures are presented in the order of the final ranking, as shown in Table~\ref{tab:results_ranking}.

\begin{figure*}[!h]%
\centering%
\includegraphics[height=19cm]{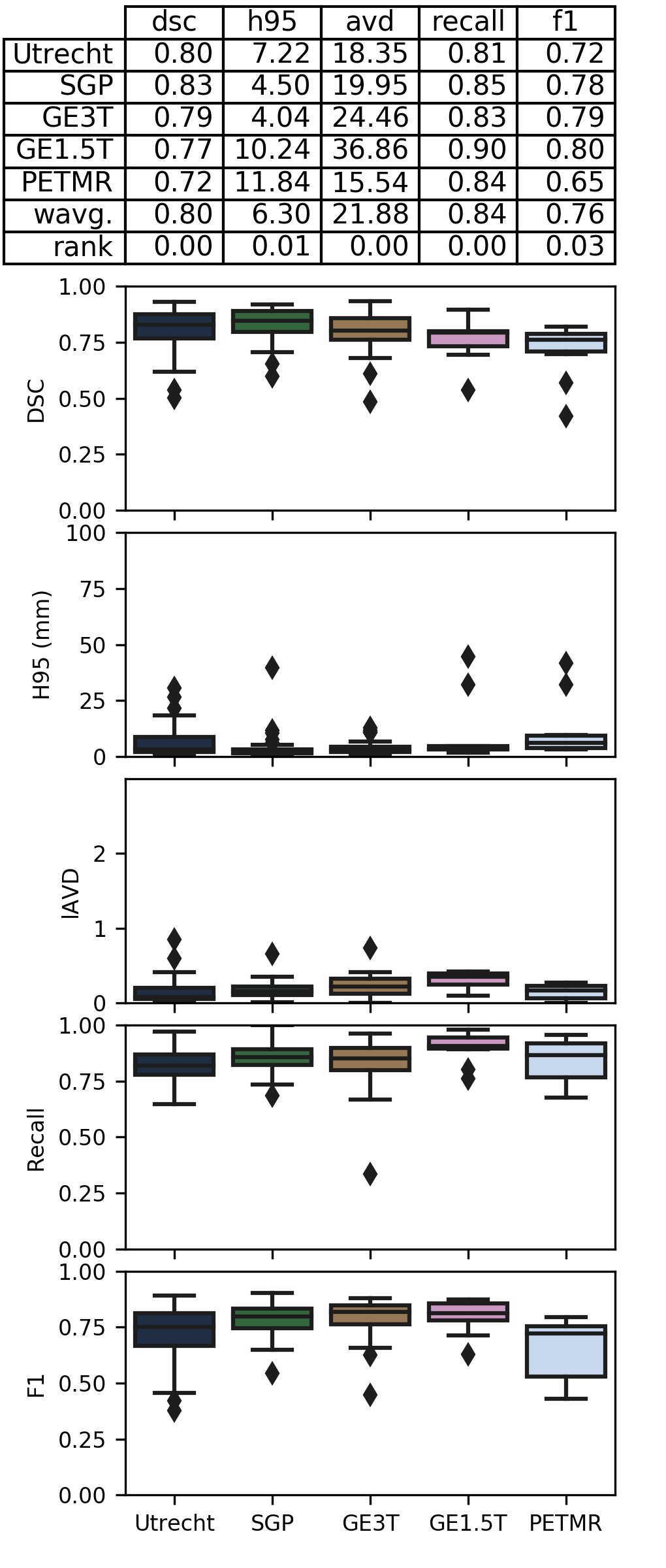}%
\includegraphics[height=19cm]{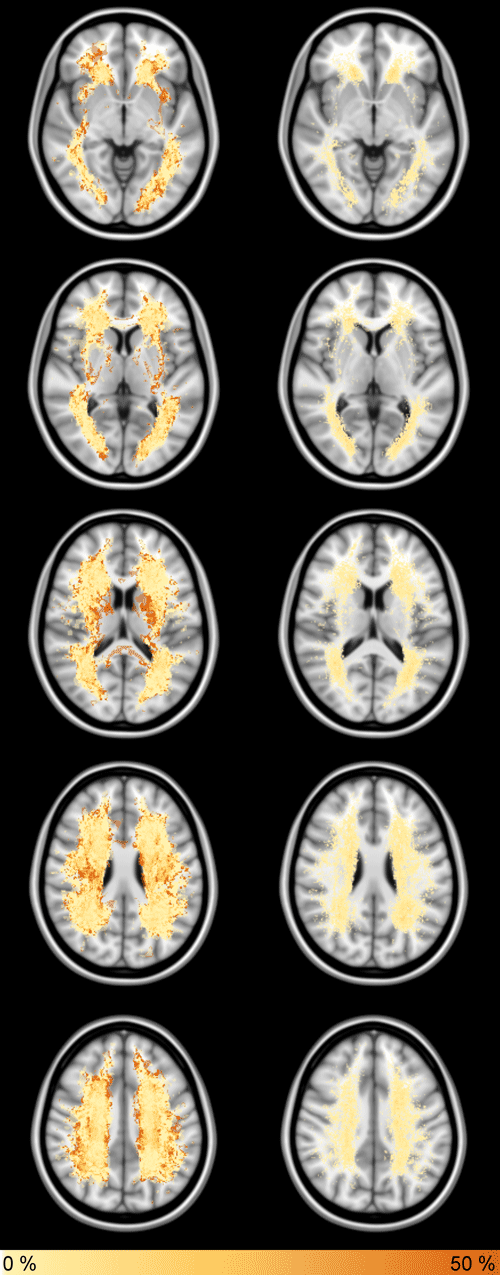}%
\caption{Detailed results of sysu\_media. The two columns on the right show the false negative rate (left) and false positive rate (right). }%
\label{app:sysu_media}%
\end{figure*}

\begin{figure*}[!h]%
\centering%
\includegraphics[height=19cm]{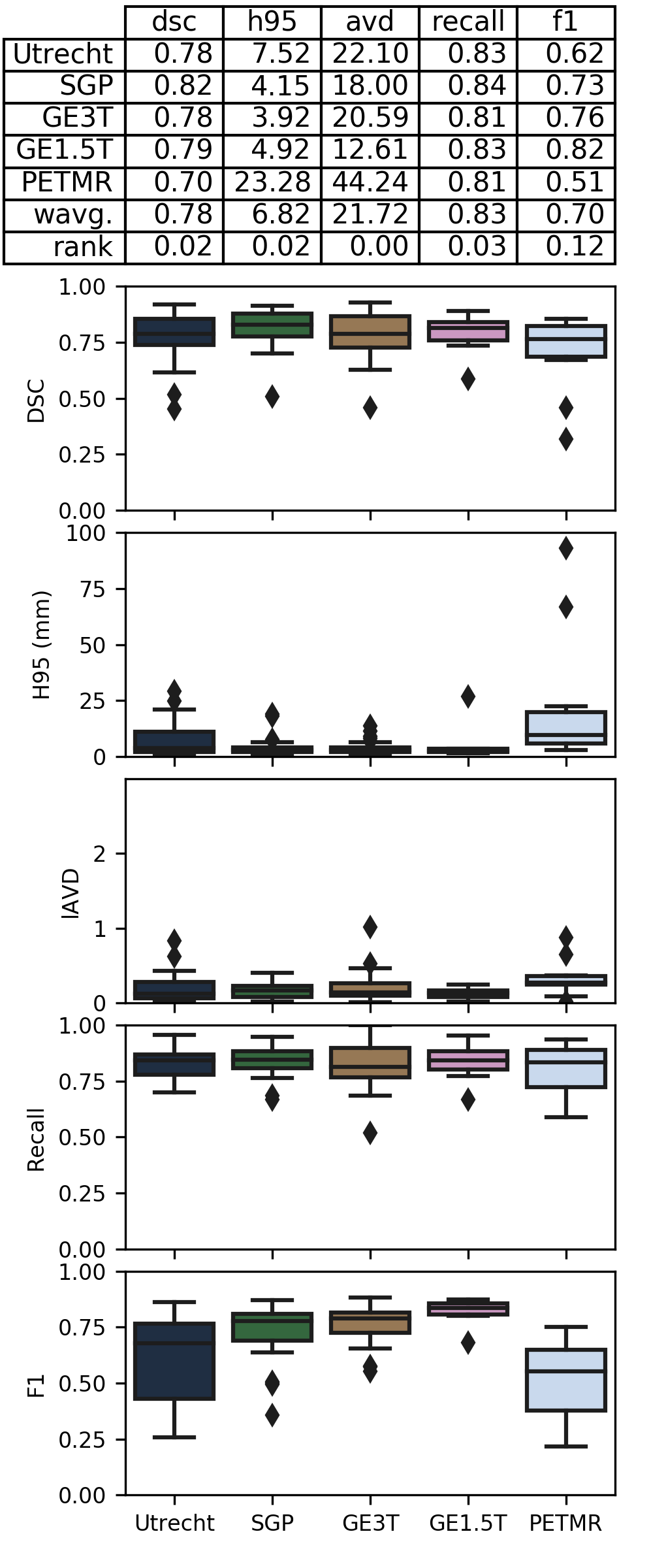}%
\includegraphics[height=19cm]{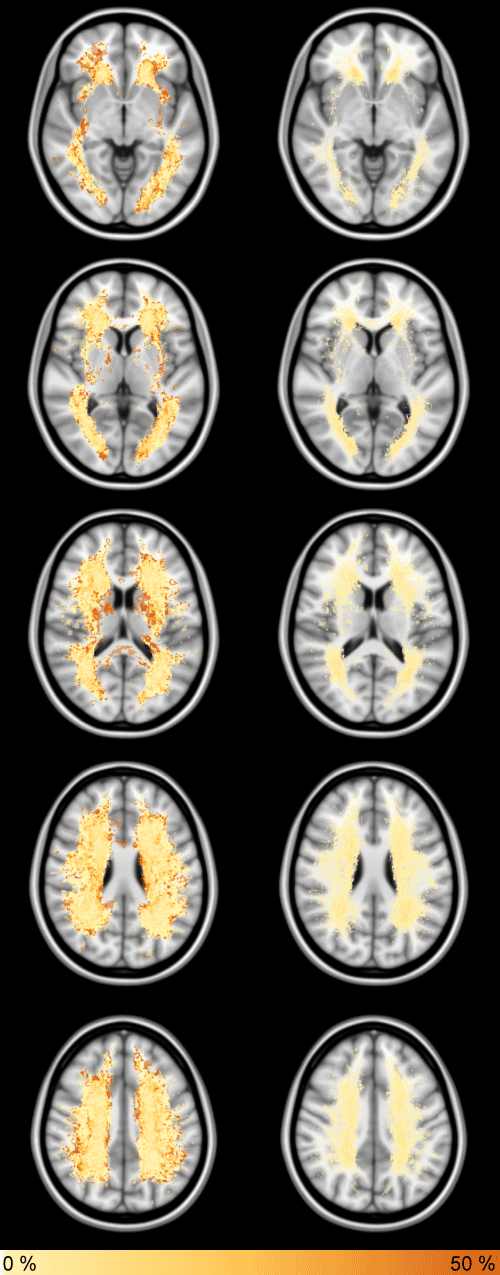}%
\caption{Detailed results of cian. The two columns on the right show the false negative rate (left) and false positive rate (right).}%
\label{app:cian}%
\end{figure*}

\begin{figure*}[!h]%
\centering%
\includegraphics[height=19cm]{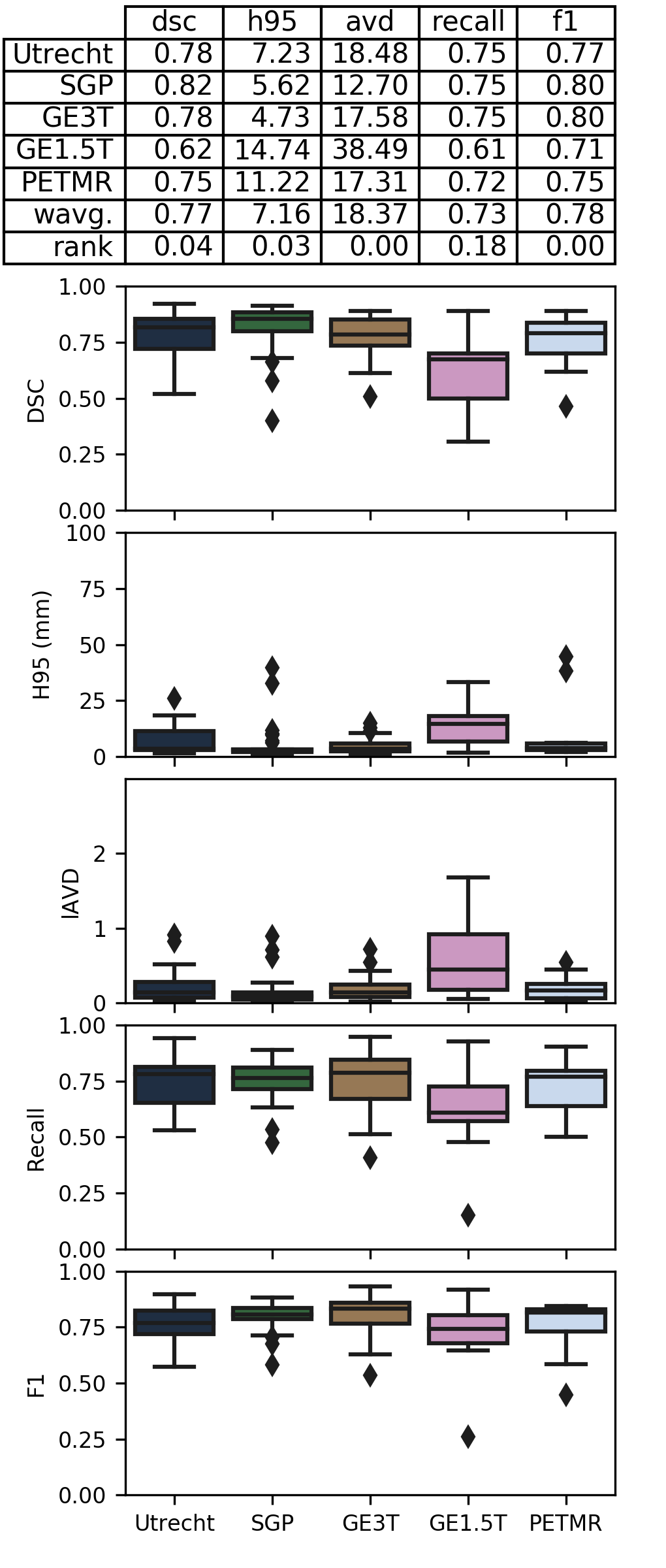}%
\includegraphics[height=19cm]{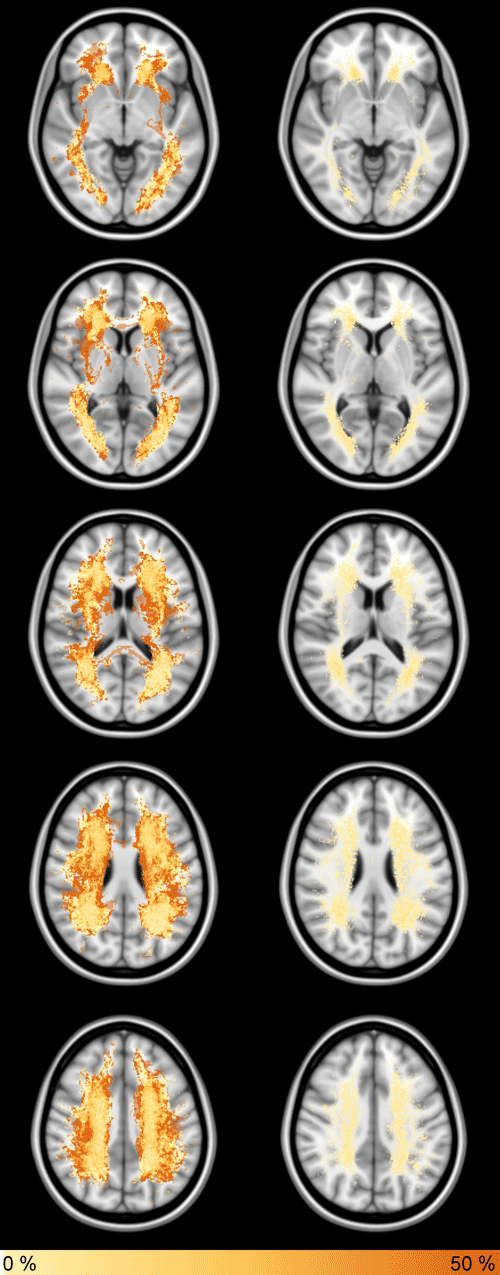}%
\caption{Detailed results of nlp\_logix. The two columns on the right show the false negative rate (left) and false positive rate (right).}%
\label{app:nlp_logix}%
\end{figure*}

\begin{figure*}[!h]%
\centering%
\includegraphics[height=19cm]{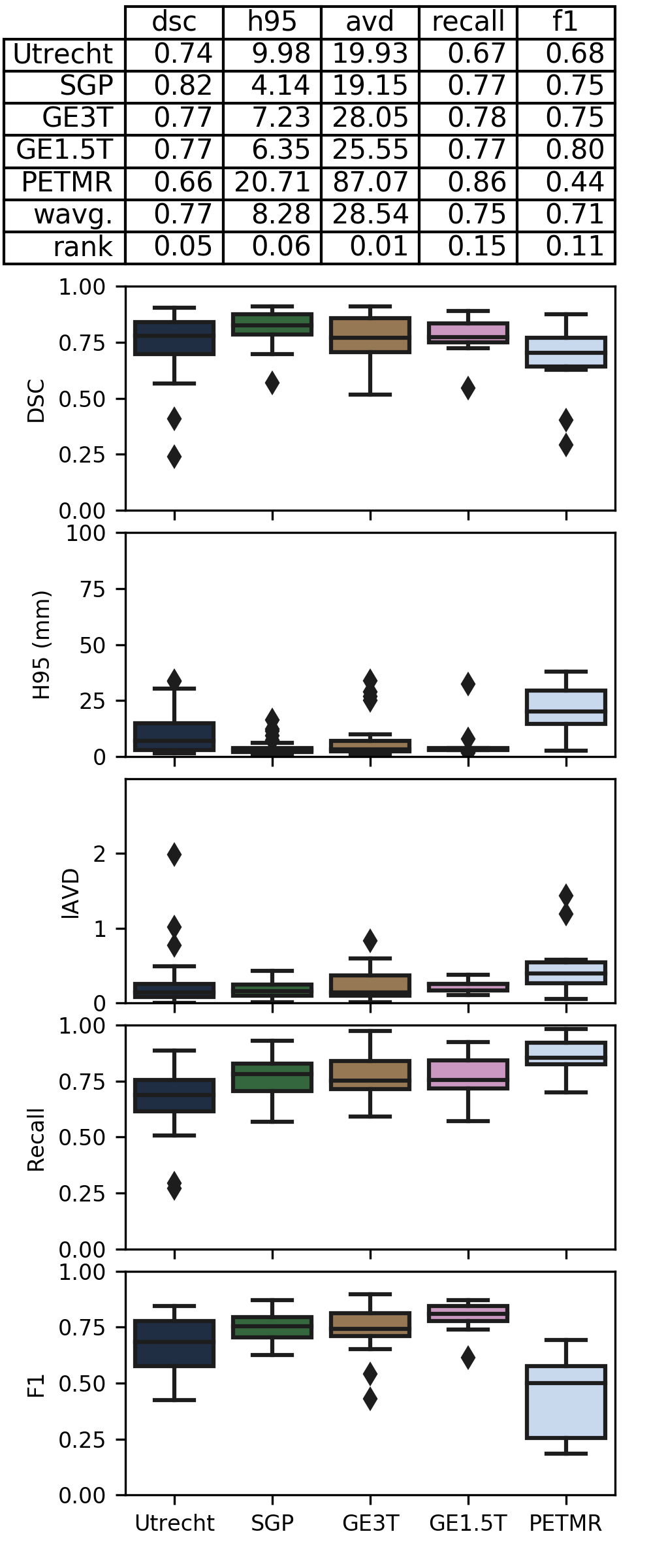}%
\includegraphics[height=19cm]{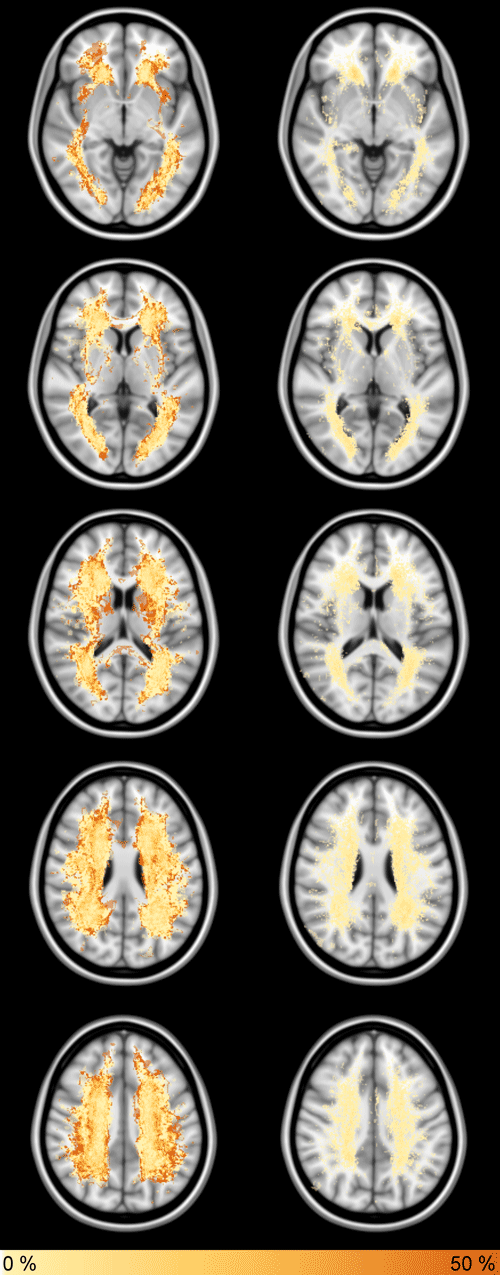}%
\caption{Detailed results of nic-vicorob. The two columns on the right show the false negative rate (left) and false positive rate (right).}%
\label{app:nic-vicorob}%
\end{figure*}

\begin{figure*}[!h]%
\centering%
\includegraphics[height=19cm]{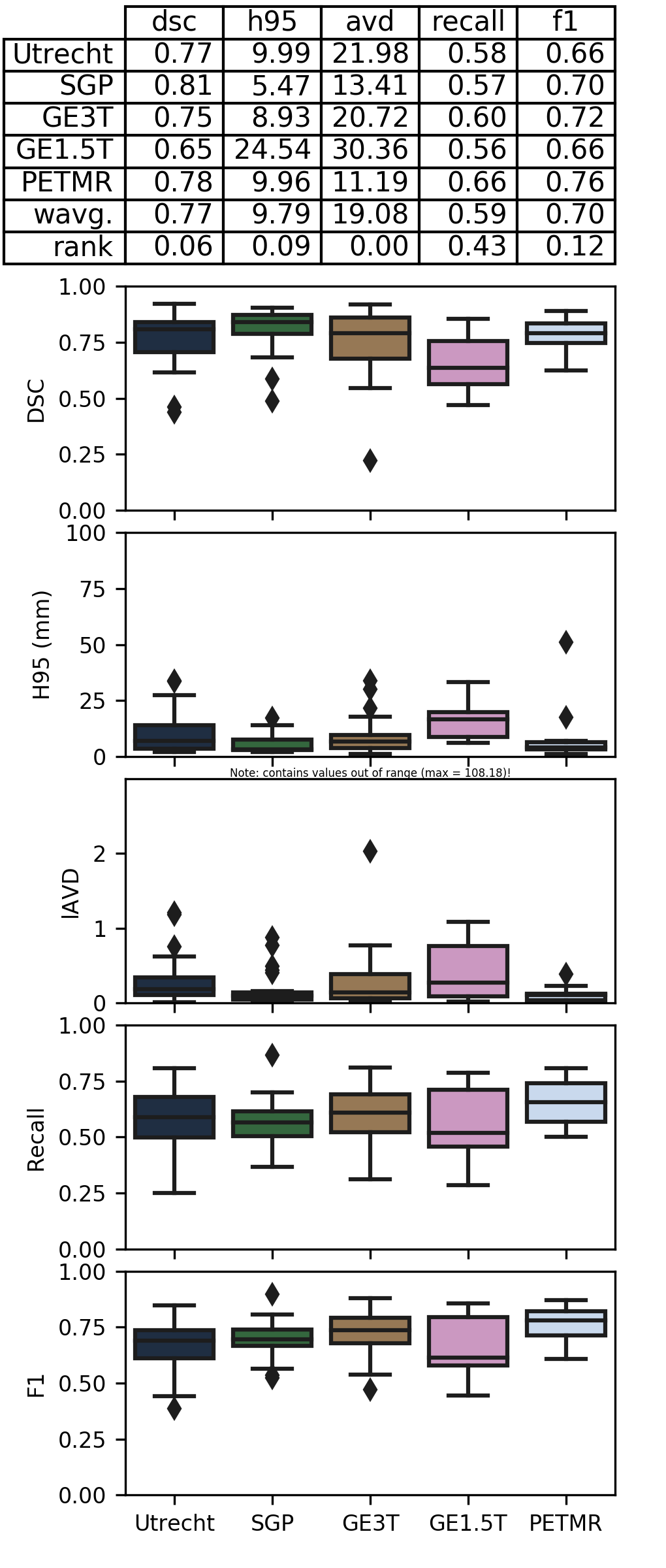}%
\includegraphics[height=19cm]{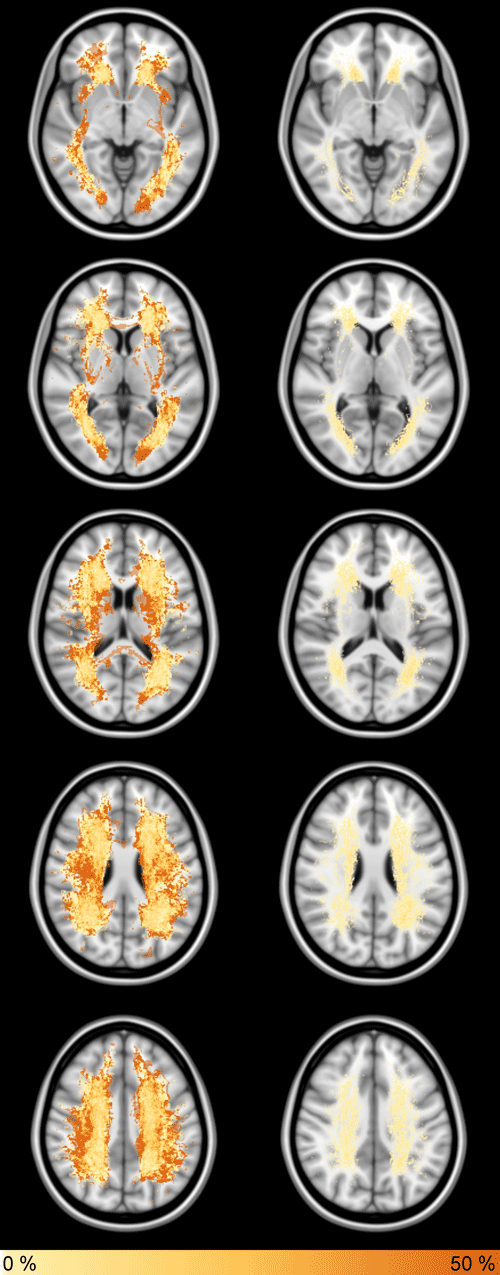}%
\caption{Detailed results of k2. The two columns on the right show the false negative rate (left) and false positive rate (right). Note: the H95 boxplot contains values out of range (max = 108.18~mm).}%
\label{app:k2}%
\end{figure*}

\begin{figure*}[!h]%
\centering%
\includegraphics[height=19cm]{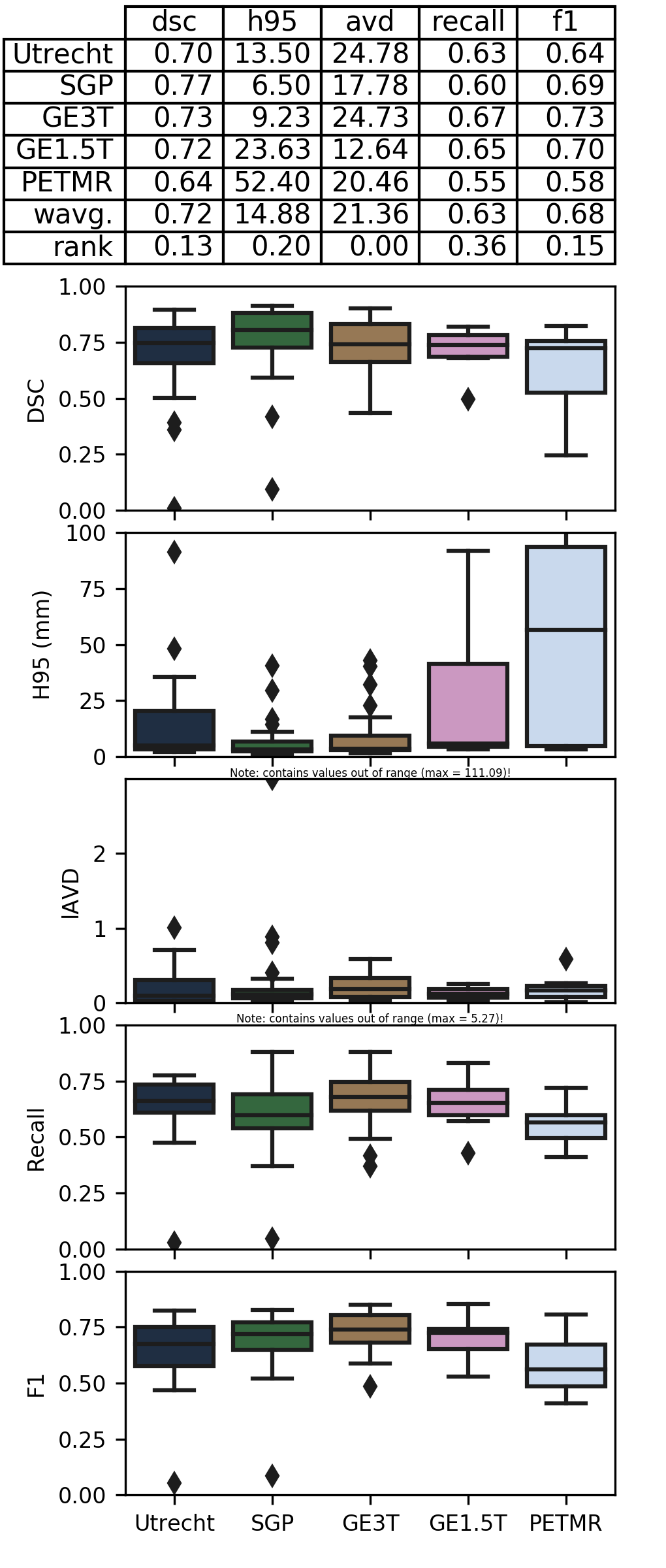}%
\includegraphics[height=19cm]{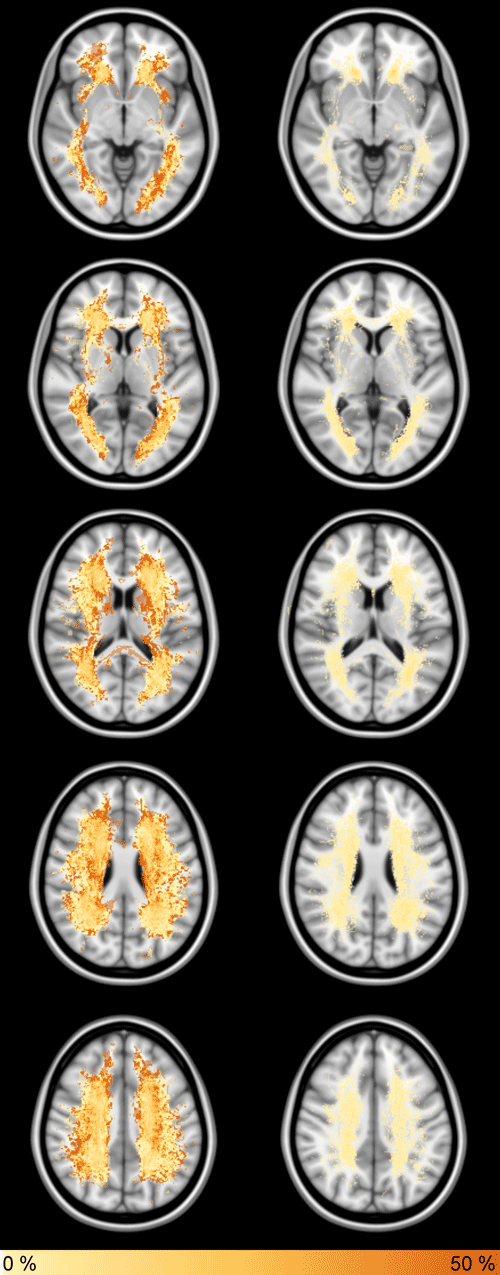}%
\caption{Detailed results of misp. The two columns on the right show the false negative rate (left) and false positive rate (right). Note: the H95 and lAVD boxplots contain values out of range (max H95 = 111.09~mm; max lAVD = 5.27).}%
\label{app:misp}%
\end{figure*}

\begin{figure*}[!h]%
\centering%
\includegraphics[height=19cm]{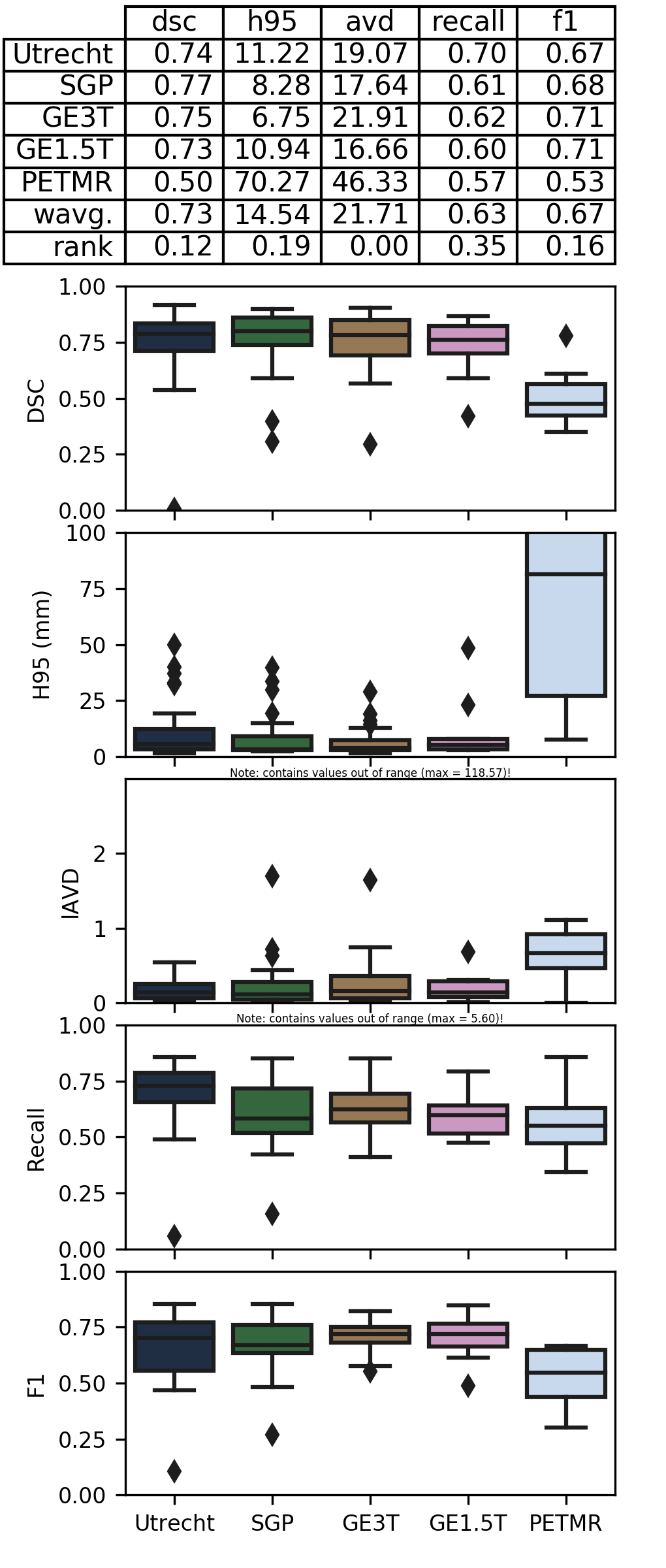}%
\includegraphics[height=19cm]{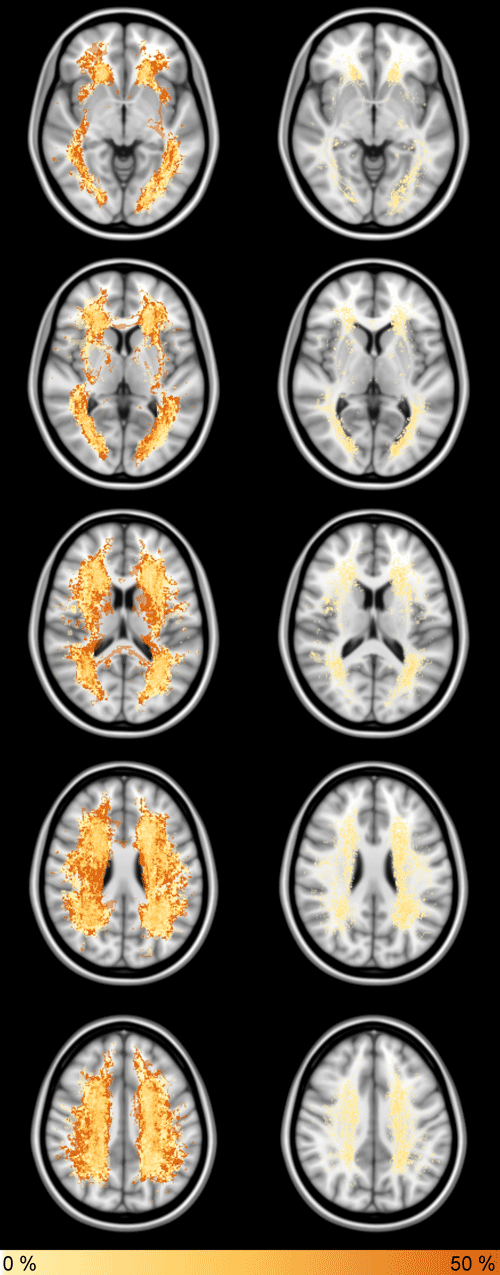}%
\caption{Detailed results of lrde. The two columns on the right show the false negative rate (left) and false positive rate (right). Note: the H95 and lAVD boxplots contain values out of range (max H95 = 118.57~mm; max lAVD = 5.60).}%
\label{app:lrde}%
\end{figure*}

\begin{figure*}[!h]%
\centering%
\includegraphics[height=19cm]{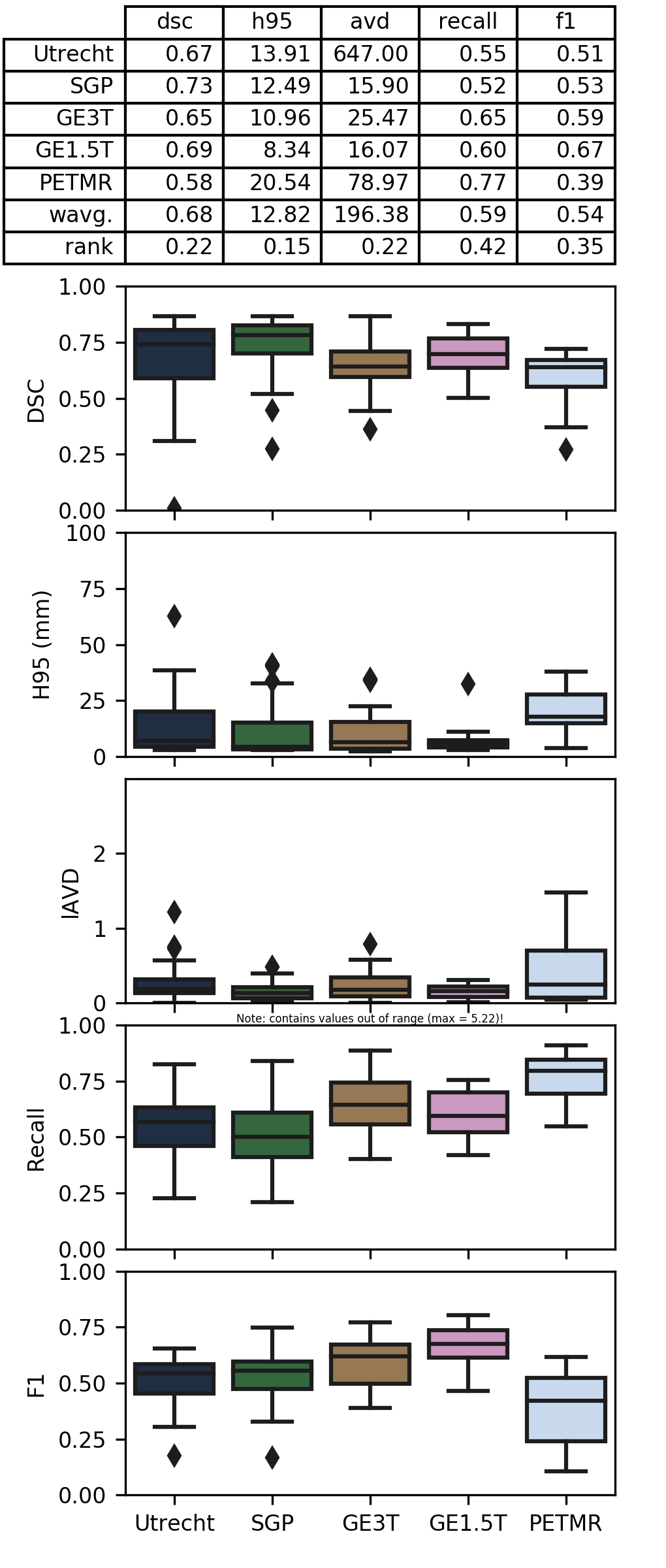}%
\includegraphics[height=19cm]{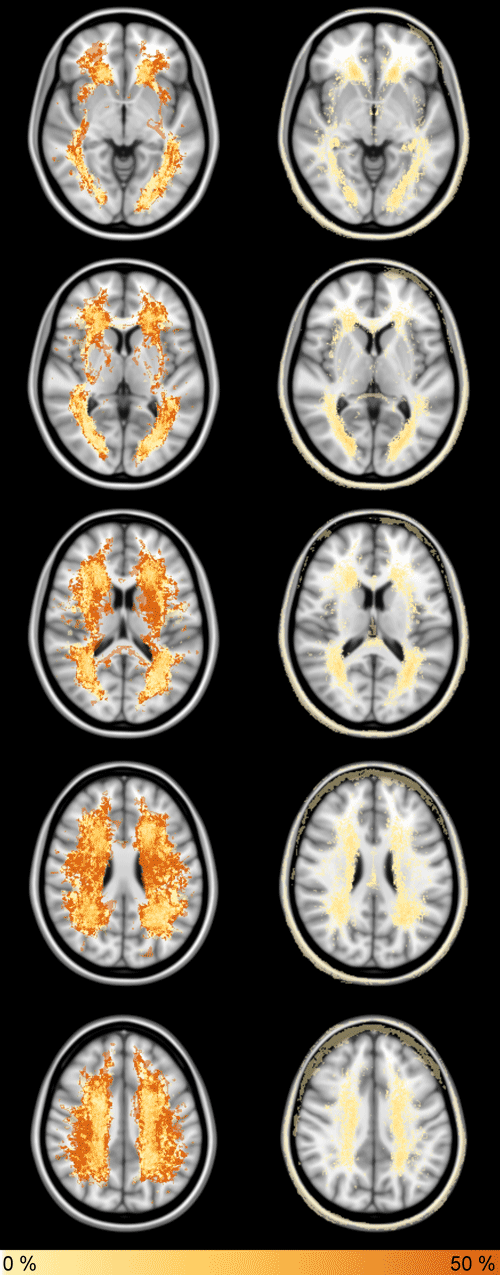}%
\caption{Detailed results of nih\_cidi. The two columns on the right show the false negative rate (left) and false positive rate (right). Note: the lAVD boxplot contains values out of range (max = 5.22).}%
\label{app:nih_cidi}%
\end{figure*}

\begin{figure*}[!h]%
\centering%
\includegraphics[height=19cm]{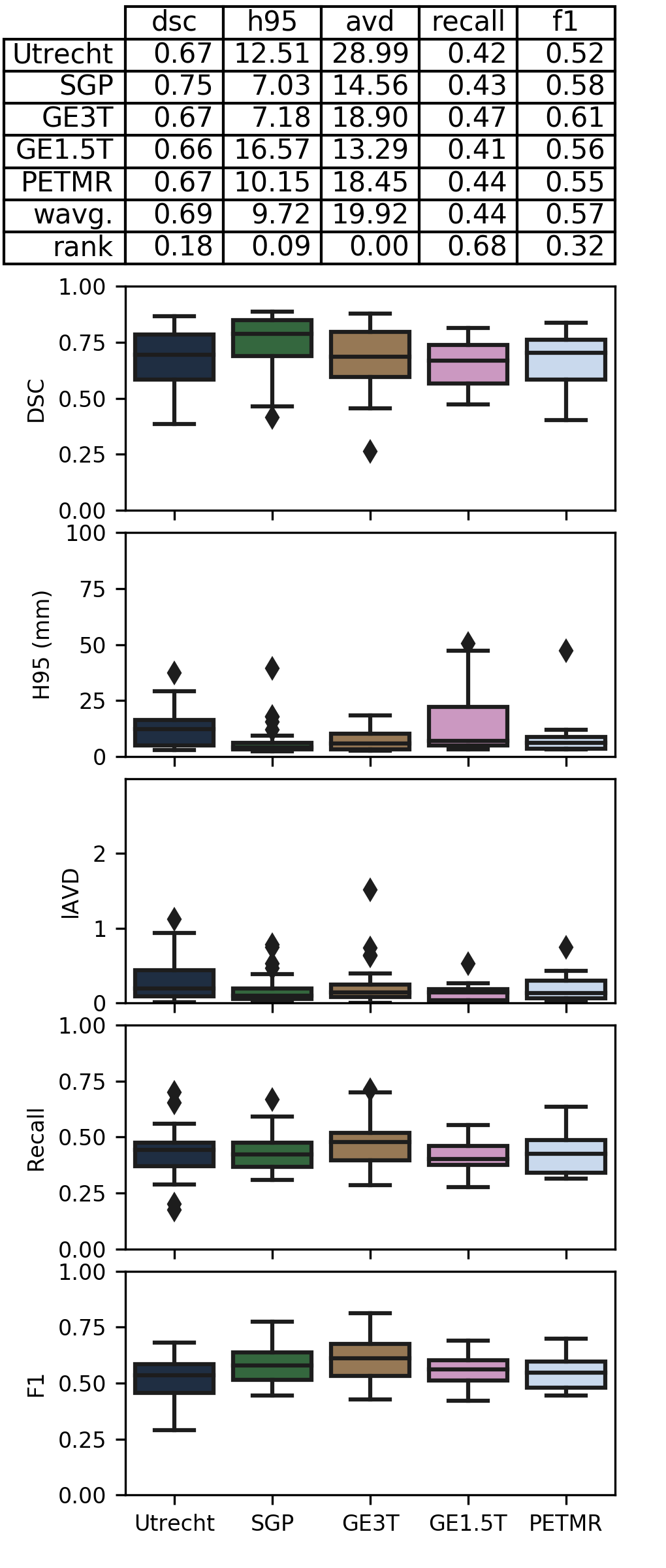}%
\includegraphics[height=19cm]{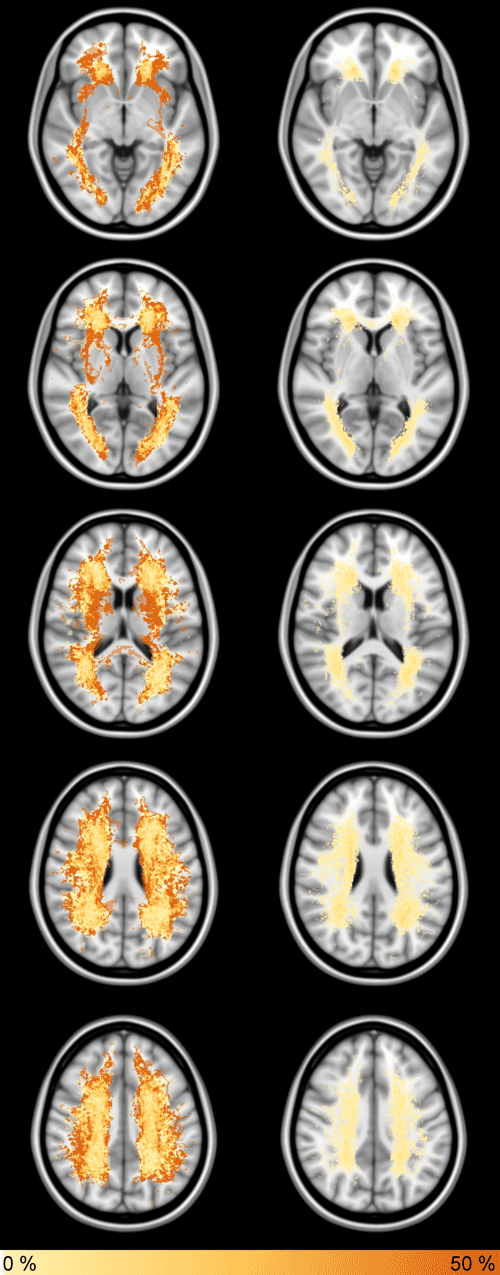}%
\caption{Detailed results of ipmi-bern. The two columns on the right show the false negative rate (left) and false positive rate (right).}%
\label{app:ipmi-bern}%
\end{figure*}

\begin{figure*}[!h]%
\centering%
\includegraphics[height=19cm]{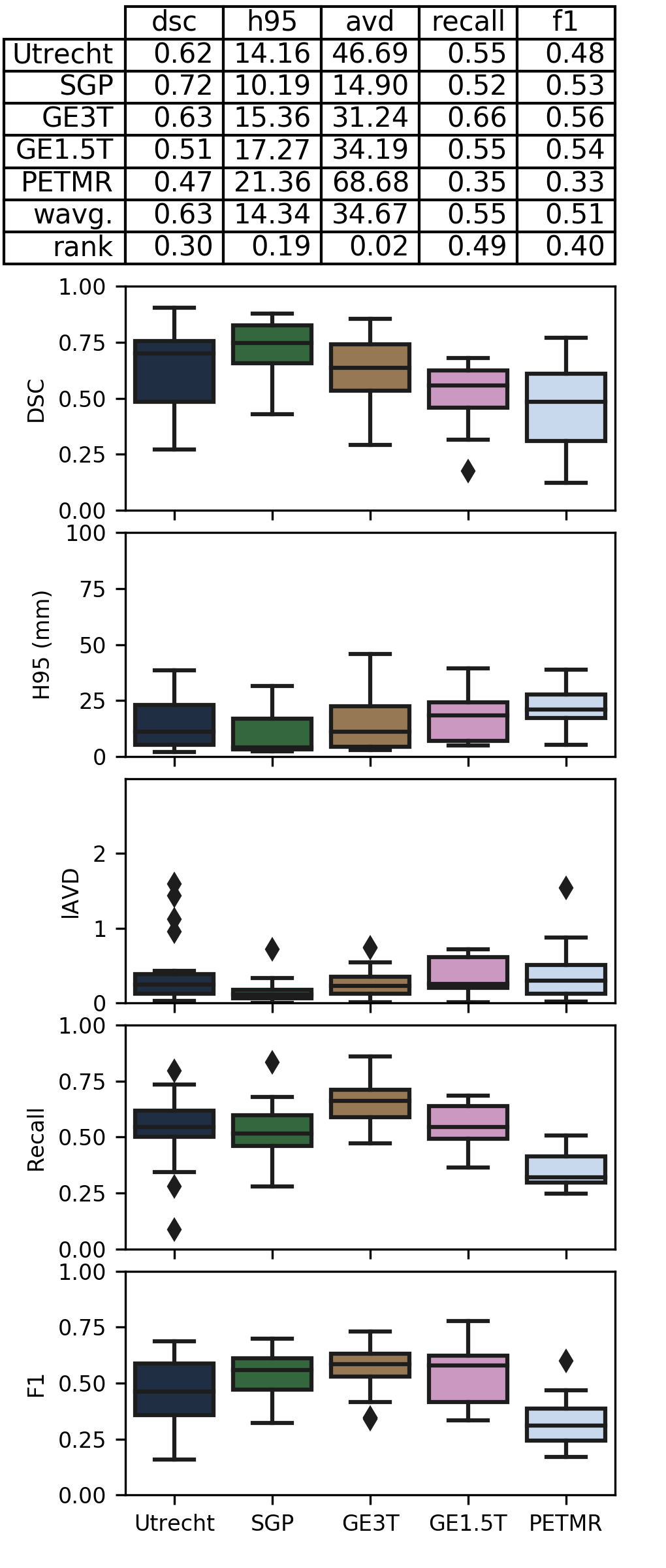}%
\includegraphics[height=19cm]{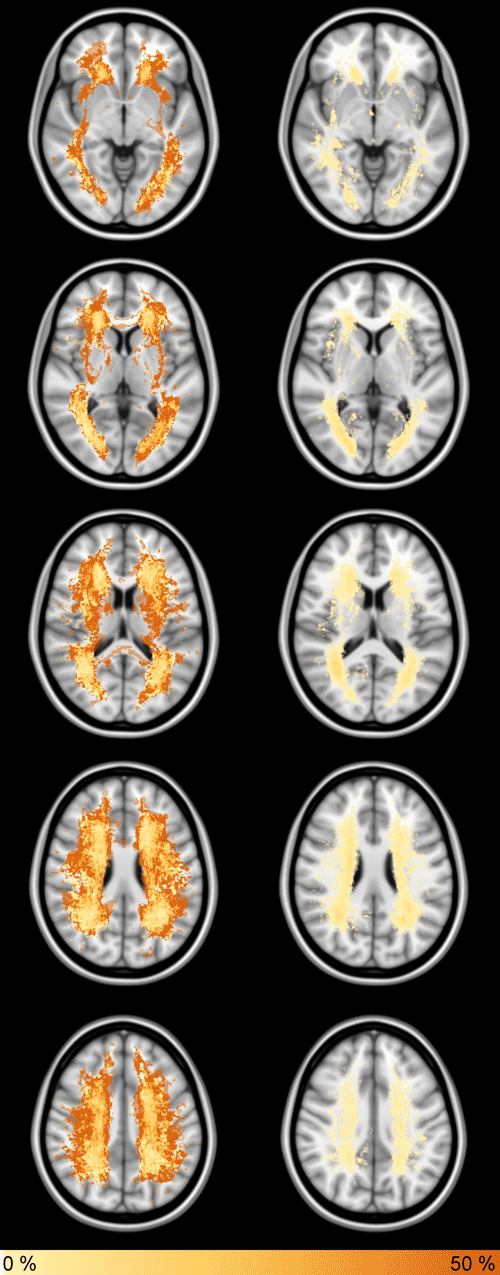}%
\caption{Detailed results of scan. The two columns on the right show the false negative rate (left) and false positive rate (right).}%
\label{app:scan}%
\end{figure*}

\begin{figure*}[!h]%
\centering%
\includegraphics[height=19cm]{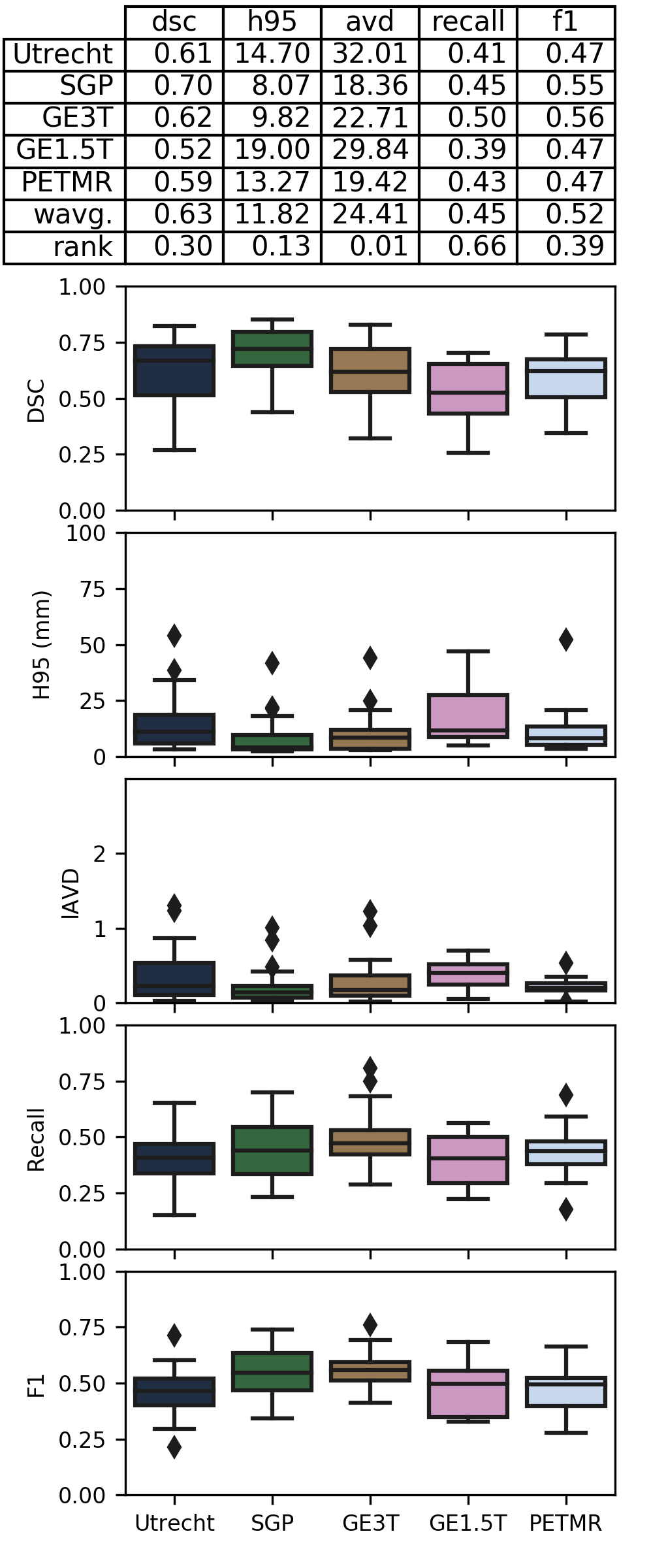}%
\includegraphics[height=19cm]{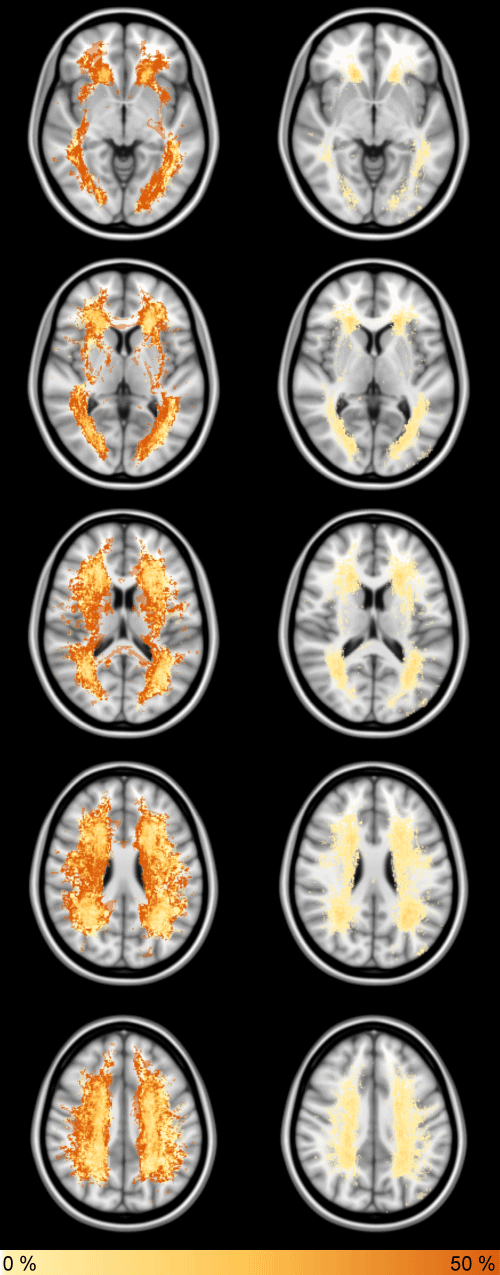}%
\caption{Detailed results of achilles. The two columns on the right show the false negative rate (left) and false positive rate (right).}%
\label{app:achilles}%
\end{figure*}

\begin{figure*}[!h]%
\centering%
\includegraphics[height=19cm]{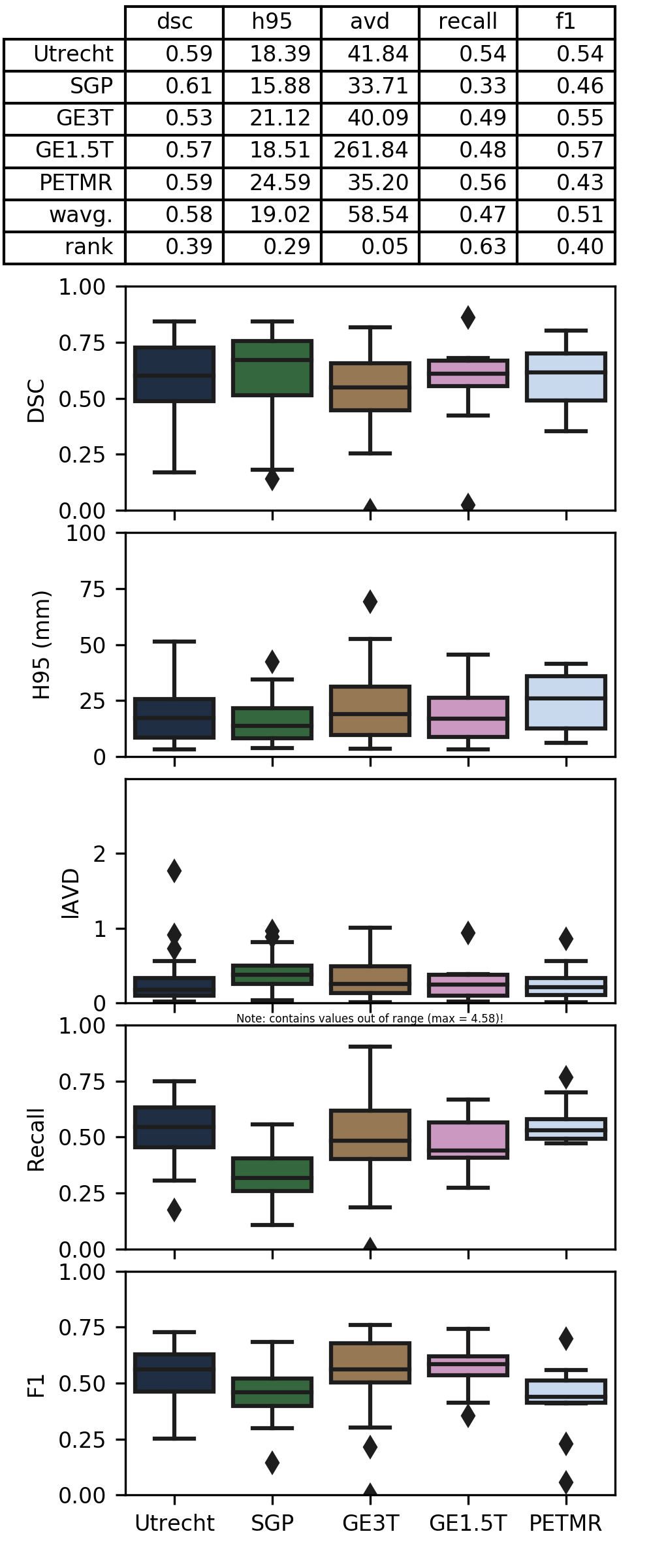}%
\includegraphics[height=19cm]{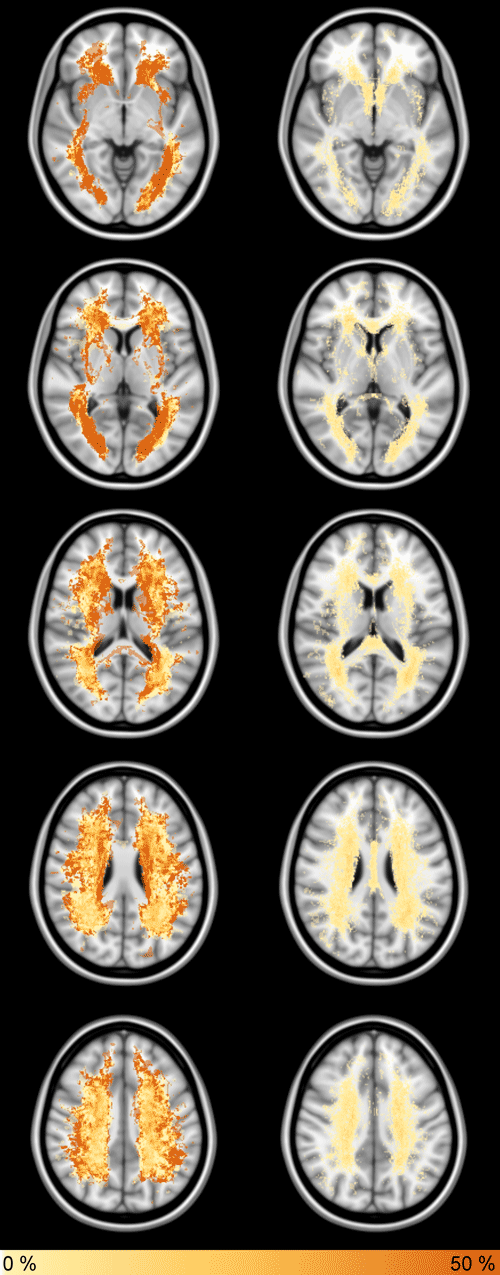}%
\caption{Detailed results of skkumedneuro. The two columns on the right show the false negative rate (left) and false positive rate (right). Note: the lAVD boxplot contains values out of range (max = 4.58).}%
\label{app:skkumedneuro}%
\end{figure*}

\begin{figure*}[!h]%
\centering%
\includegraphics[height=19cm]{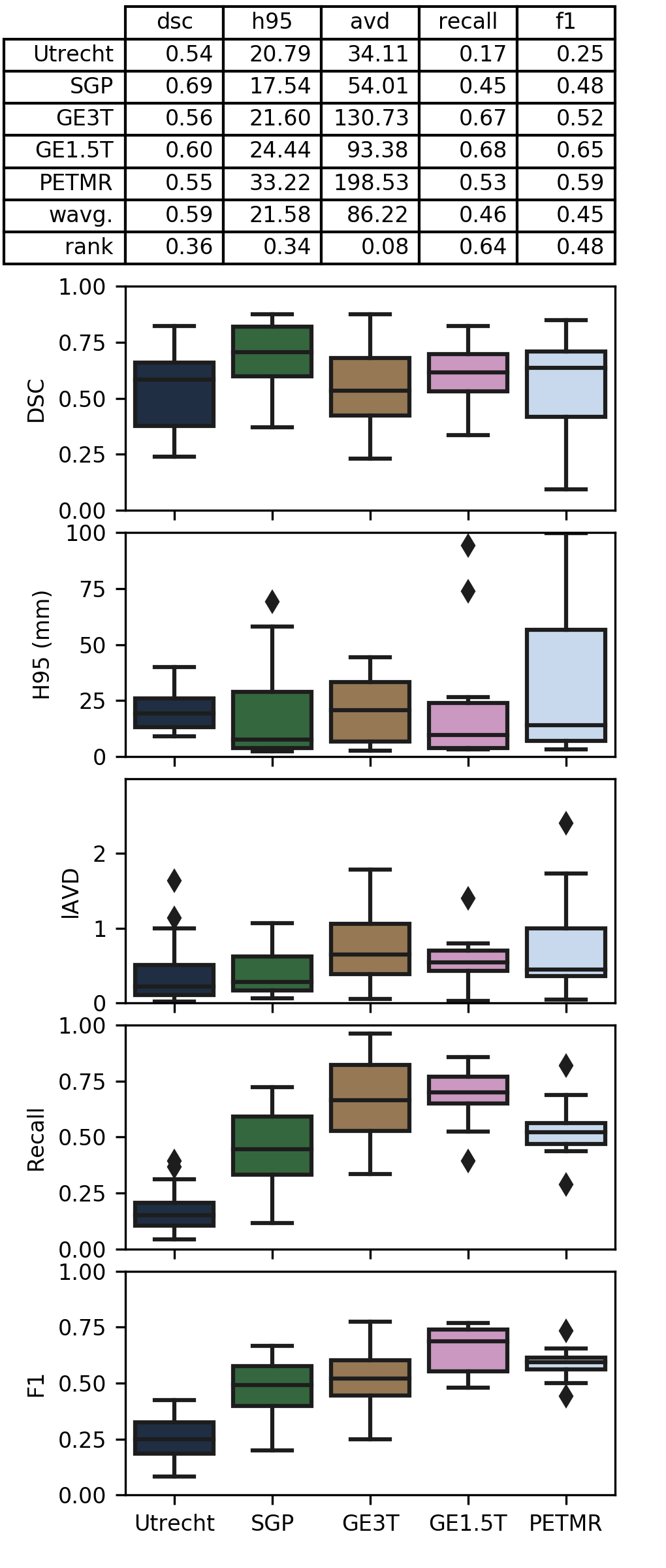}%
\includegraphics[height=19cm]{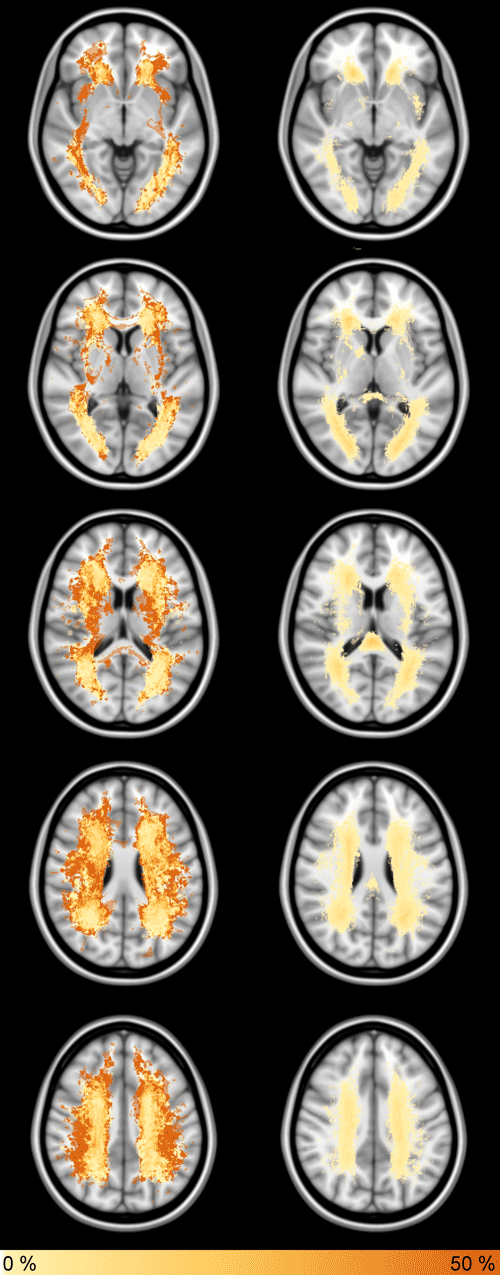}%
\caption{Detailed results of tignet. The two columns on the right show the false negative rate (left) and false positive rate (right).}%
\label{app:tignet}%
\end{figure*}

\begin{figure*}[!h]%
\centering%
\includegraphics[height=19cm]{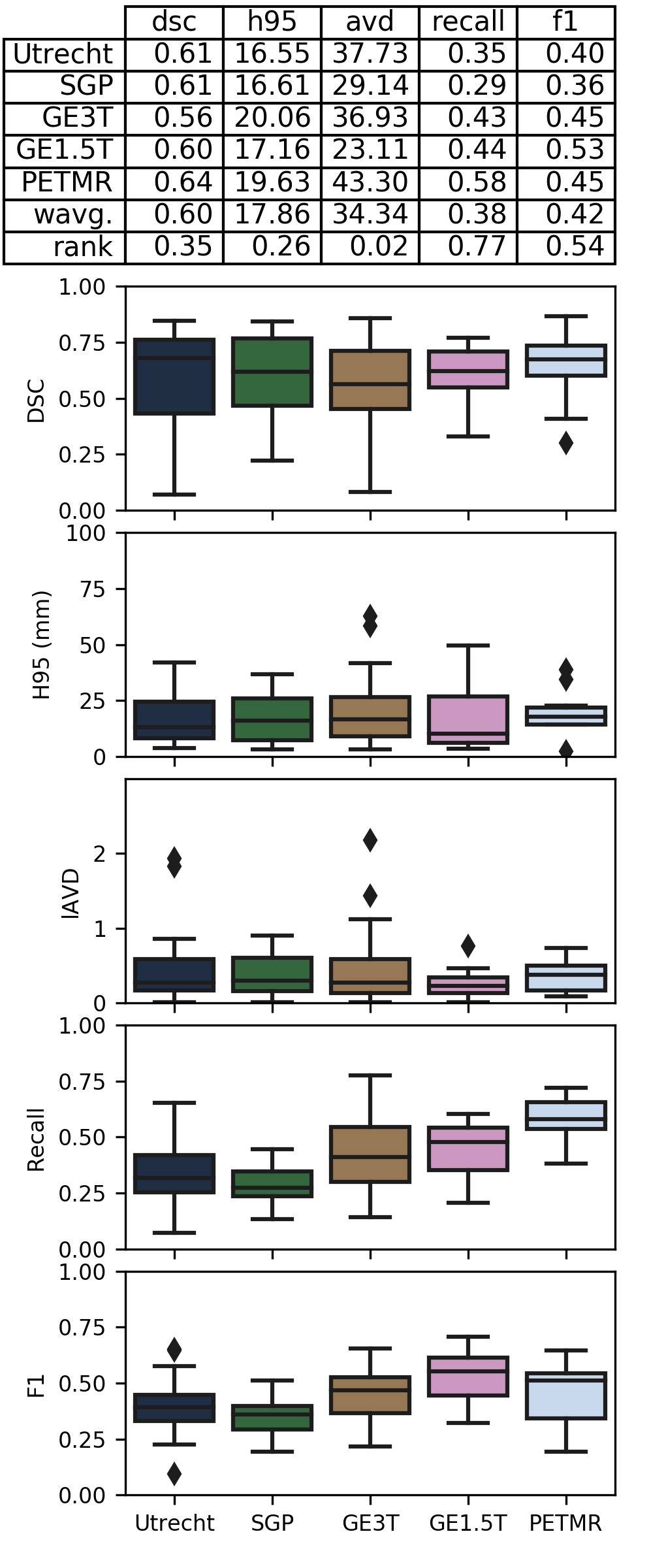}%
\includegraphics[height=19cm]{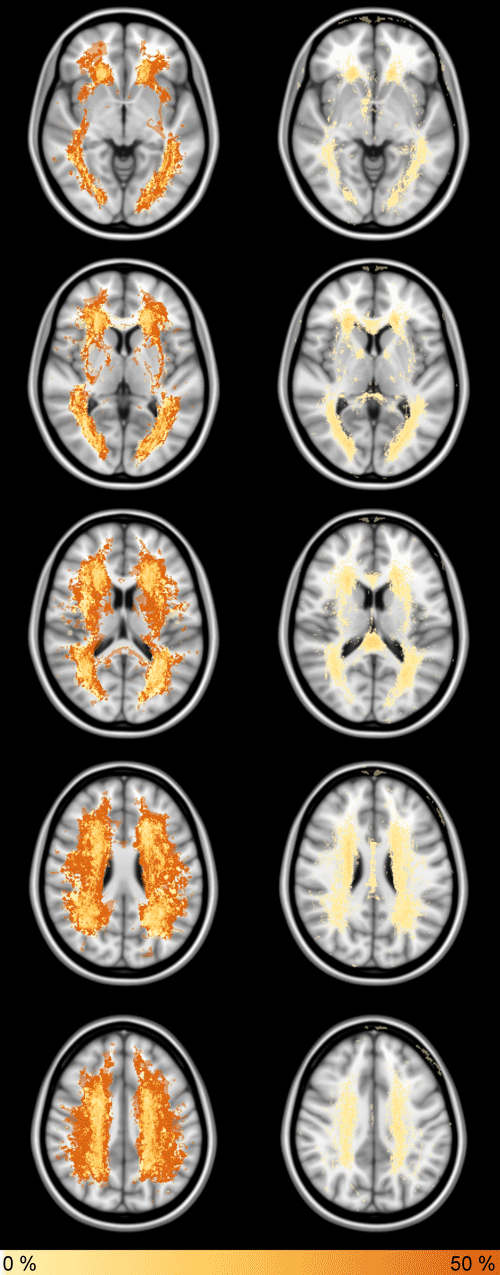}%
\caption{Detailed results of tig. The two columns on the right show the false negative rate (left) and false positive rate (right).}%
\label{app:tig}%
\end{figure*}

\begin{figure*}[!h]%
\centering%
\includegraphics[height=19cm]{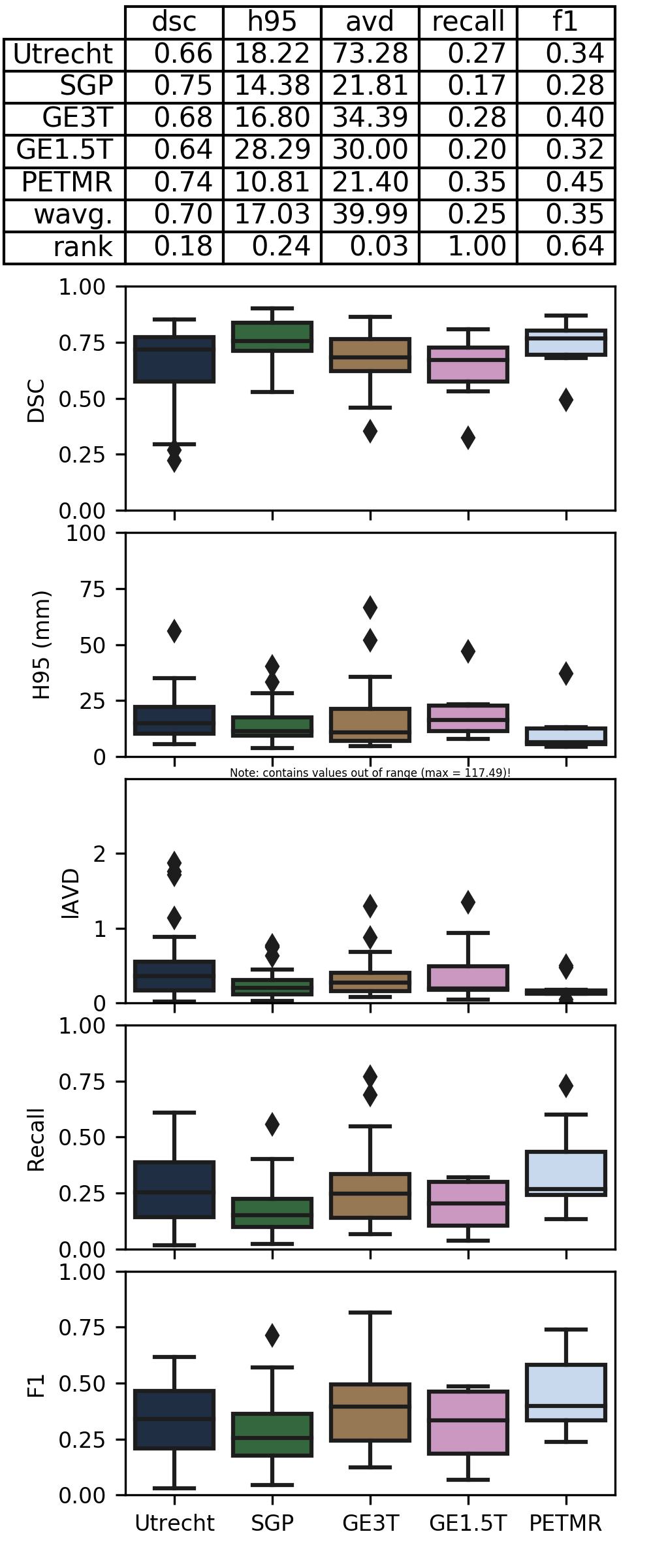}%
\includegraphics[height=19cm]{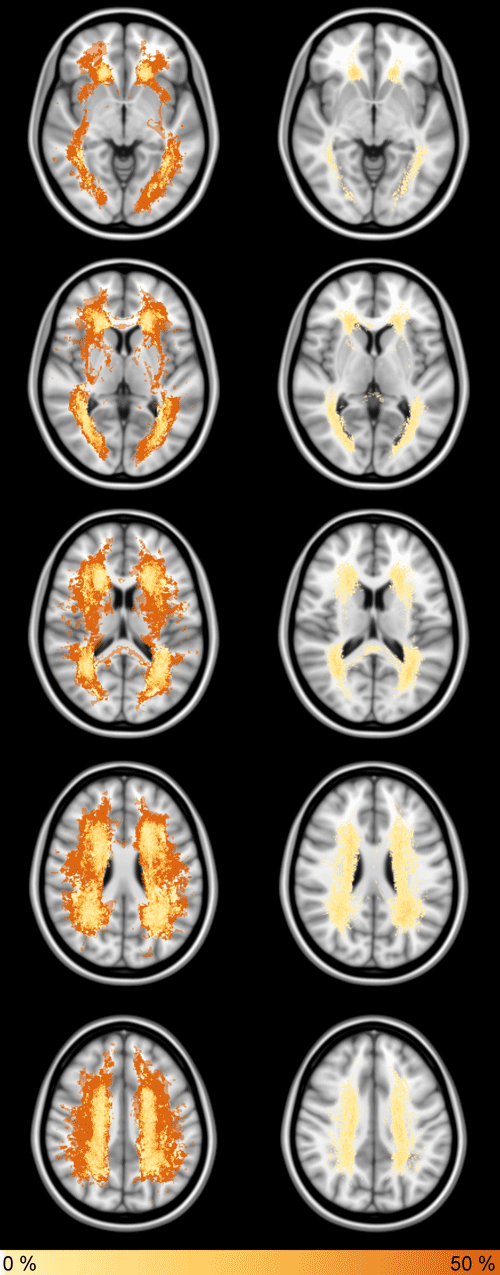}%
\caption{Detailed results of knight. The two columns on the right show the false negative rate (left) and false positive rate (right). Note: the H95 boxplot contains values out of range (max = 117.49~mm).}%
\label{app:knight}%
\end{figure*}

\begin{figure*}[!h]%
\centering%
\includegraphics[height=19cm]{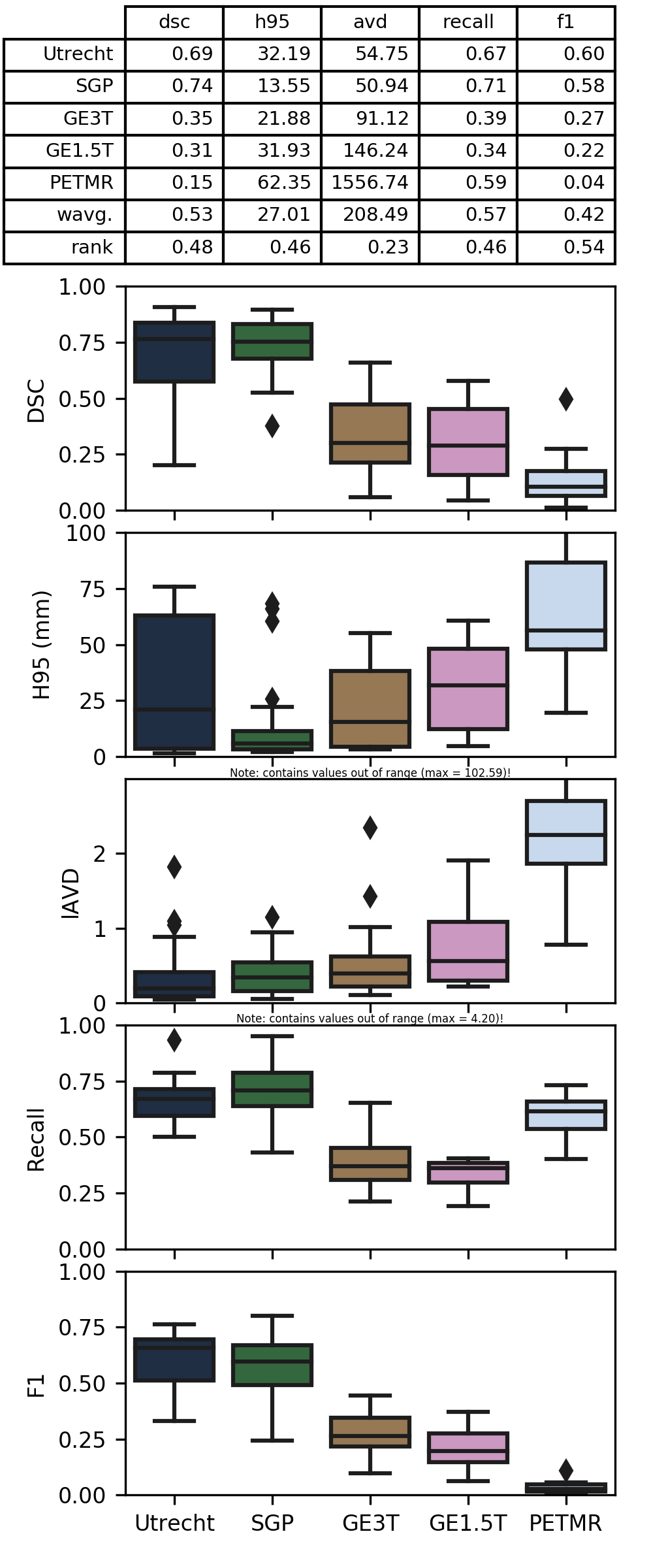}%
\includegraphics[height=19cm]{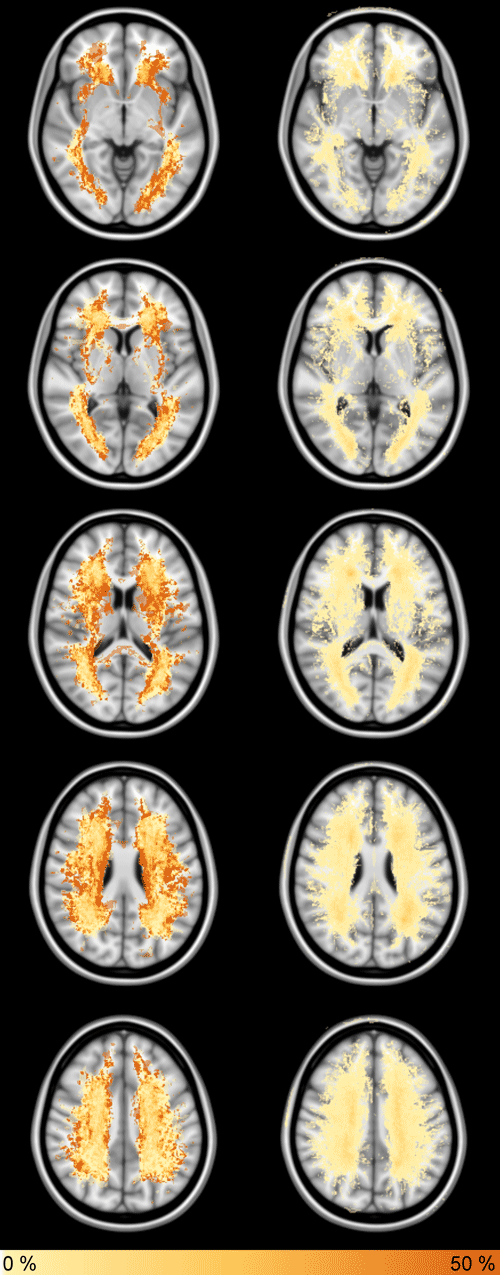}%
\caption{Detailed results of upc\_dlmi. The two columns on the right show the false negative rate (left) and false positive rate (right). Note: the H95 and lAVD boxplots contain values out of range (max H95 = 102.59~mm; max lAVD = 4.20).}%
\label{app:upc_dlmi}%
\end{figure*}

\begin{figure*}[!h]%
\centering%
\includegraphics[height=19cm]{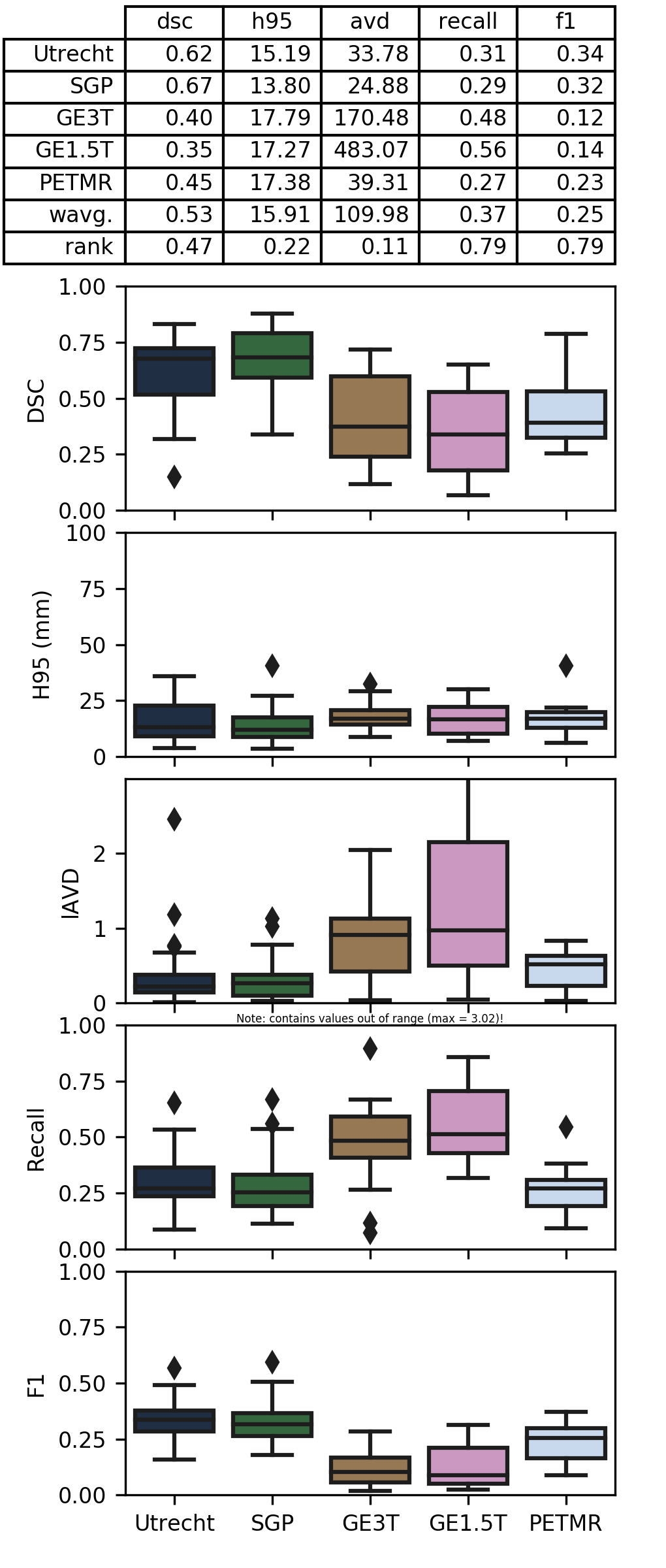}%
\includegraphics[height=19cm]{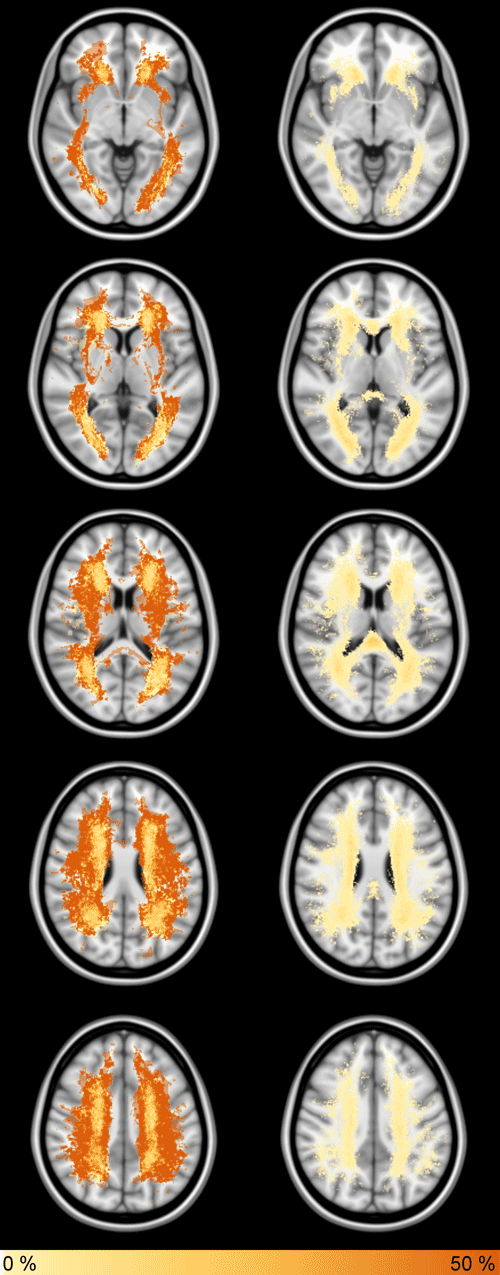}%
\caption{Detailed results of nist. The two columns on the right show the false negative rate (left) and false positive rate (right). Note: the lAVD boxplot contains values out of range (max = 3.02).}%
\label{app:nist}%
\end{figure*}

\begin{figure*}[!h]%
\centering%
\includegraphics[height=19cm]{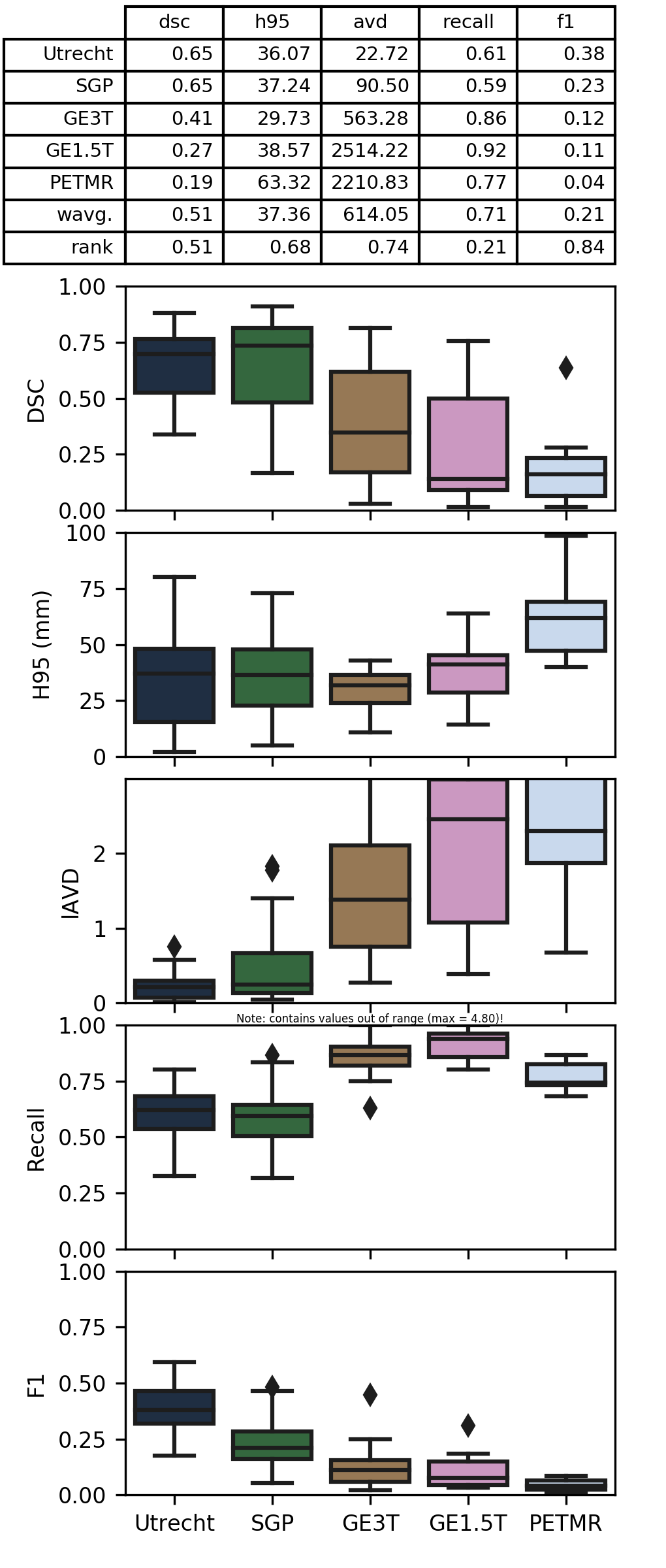}%
\includegraphics[height=19cm]{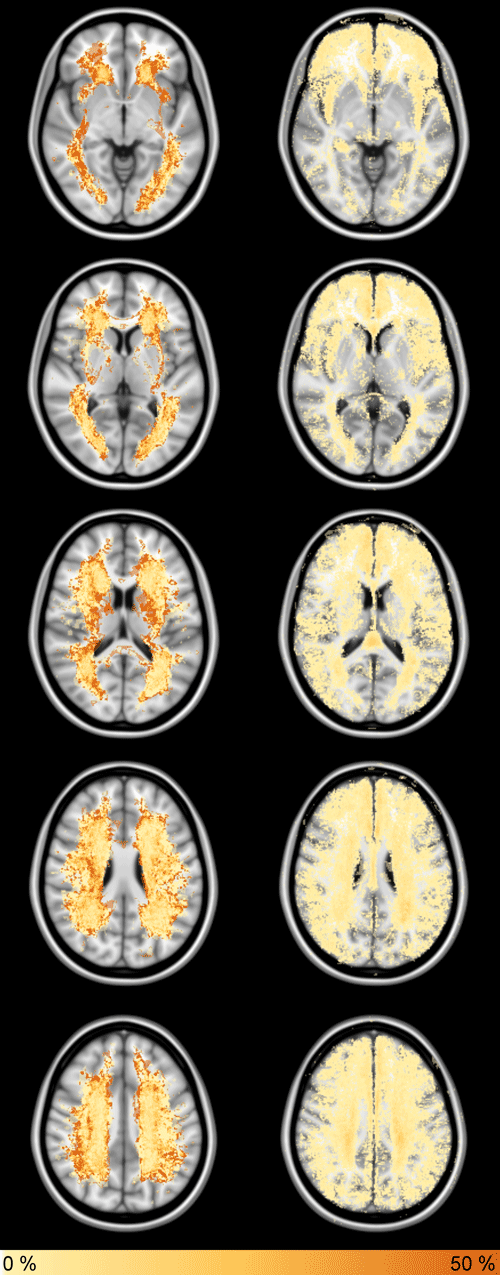}%
\caption{Detailed results of neuro.ml. The two columns on the right show the false negative rate (left) and false positive rate (right). Note: the lAVD boxplot contains values out of range (max = 4.80).}%
\label{app:neuro.ml}%
\end{figure*}

\begin{figure*}[!h]%
\centering%
\includegraphics[height=19cm]{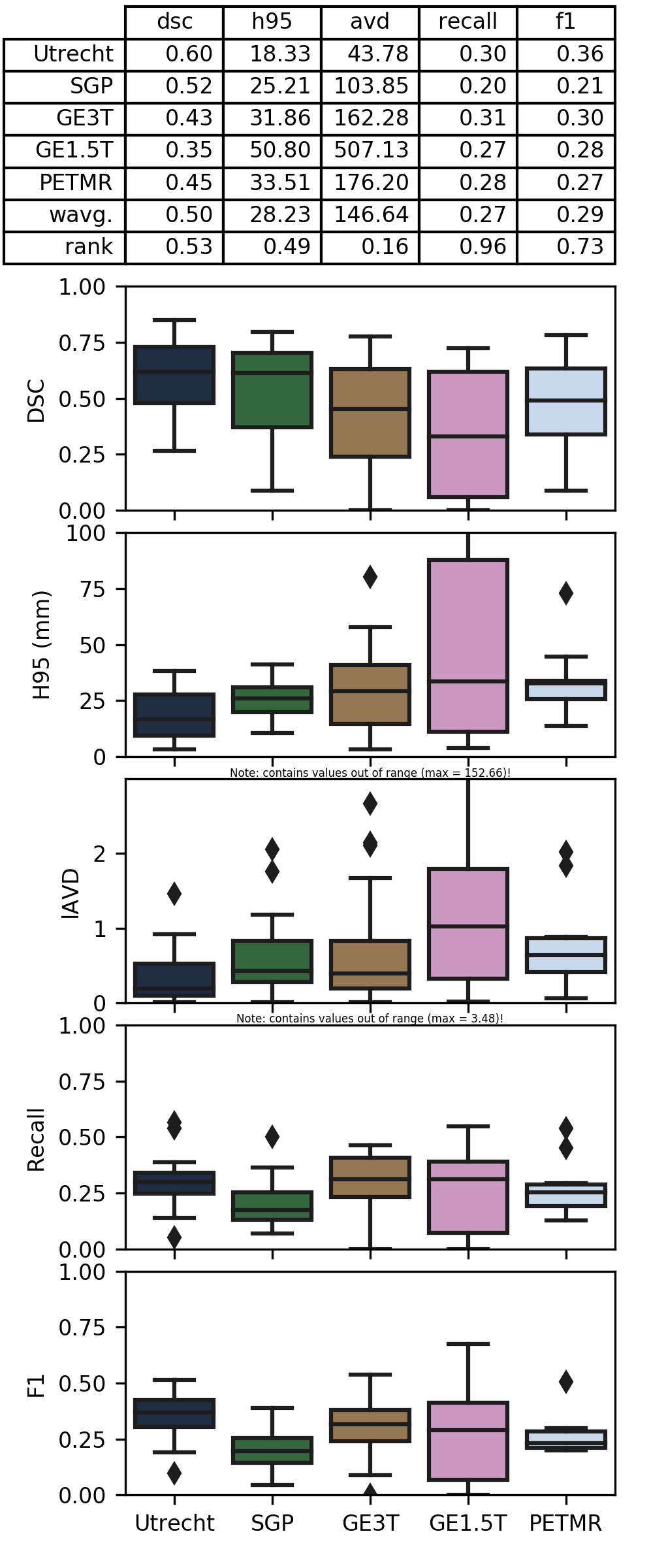}%
\includegraphics[height=19cm]{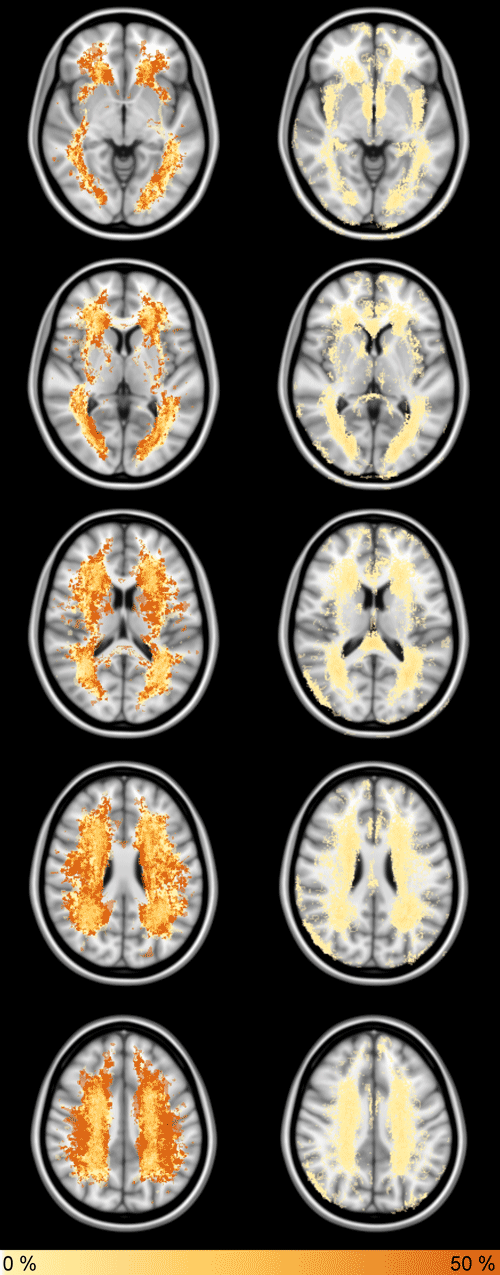}%
\caption{Detailed results of text\_class. The two columns on the right show the false negative rate (left) and false positive rate (right). Note: the H95 and lAVD boxplots contain values out of range (max H95 = 152.66~mm; max lAVD = 3.48).}%
\label{app:text_class}%
\end{figure*}

\begin{figure*}[!h]%
\centering%
\includegraphics[height=19cm]{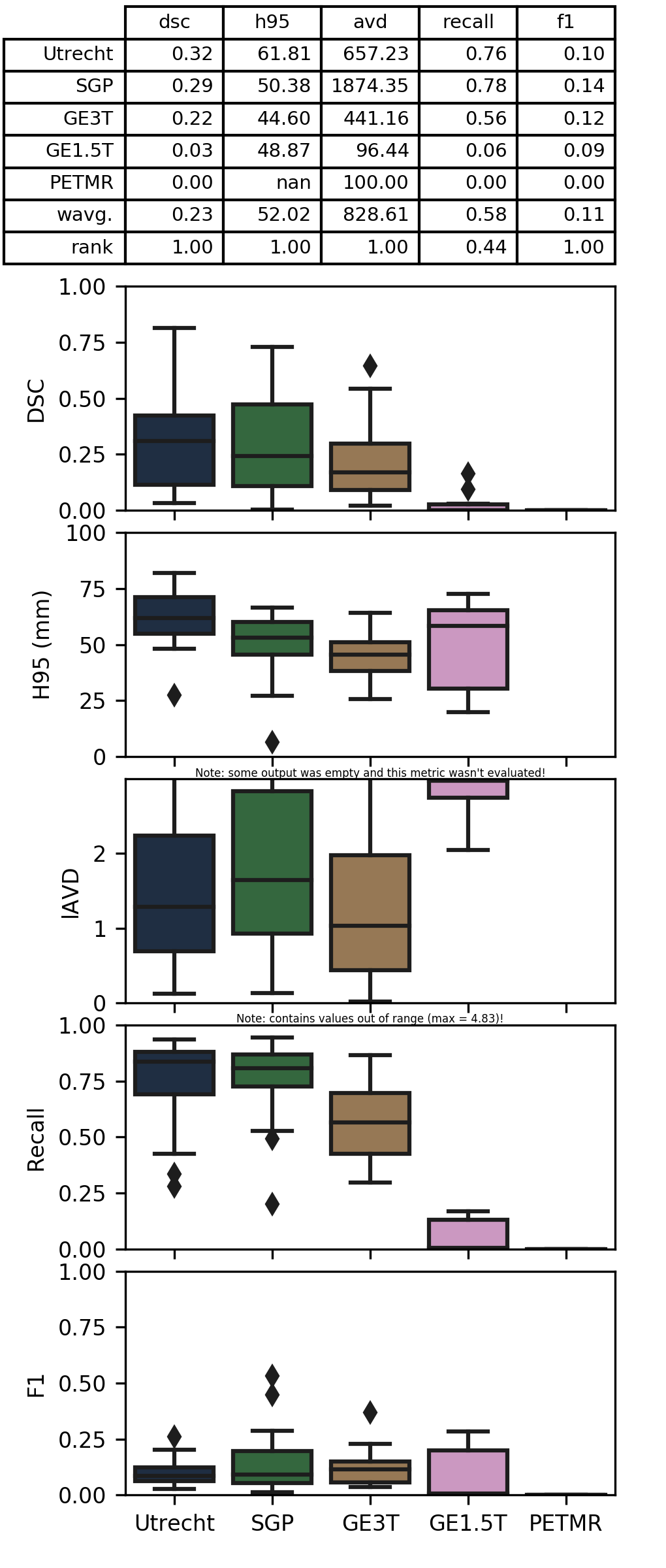}%
\includegraphics[height=19cm]{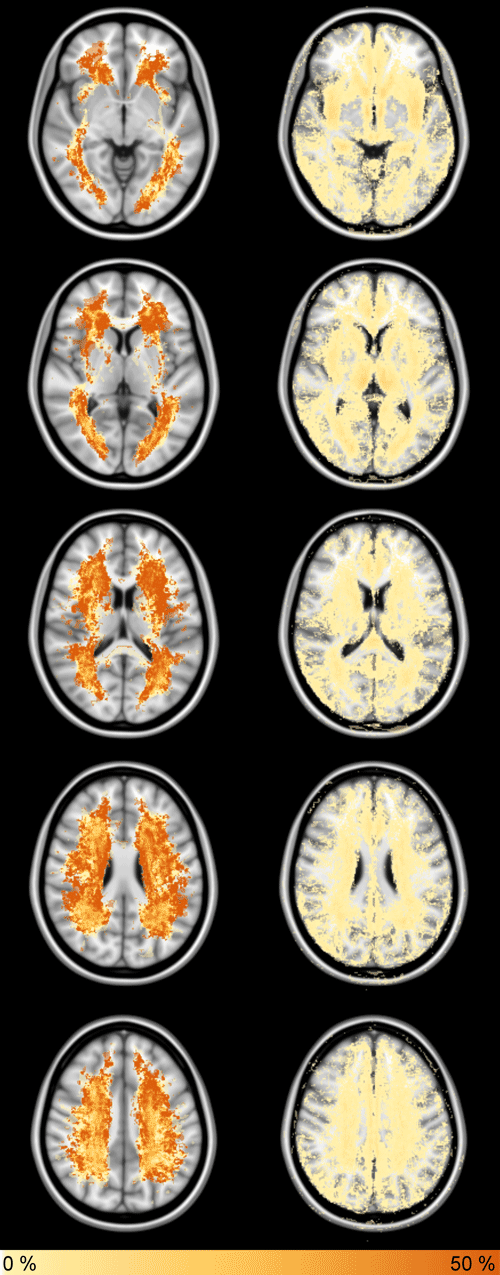}%
\caption{Detailed results of hadi. The two columns on the right show the false negative rate (left) and false positive rate (right). Note: some output was empty and the H95 and lAVD were not evaluated, and the lAVD boxplot contains values out of range (max = 4.83).}%
\label{app:hadi}%
\end{figure*}

\begin{figure*}[!h]%
\centering%
\includegraphics[height=19cm]{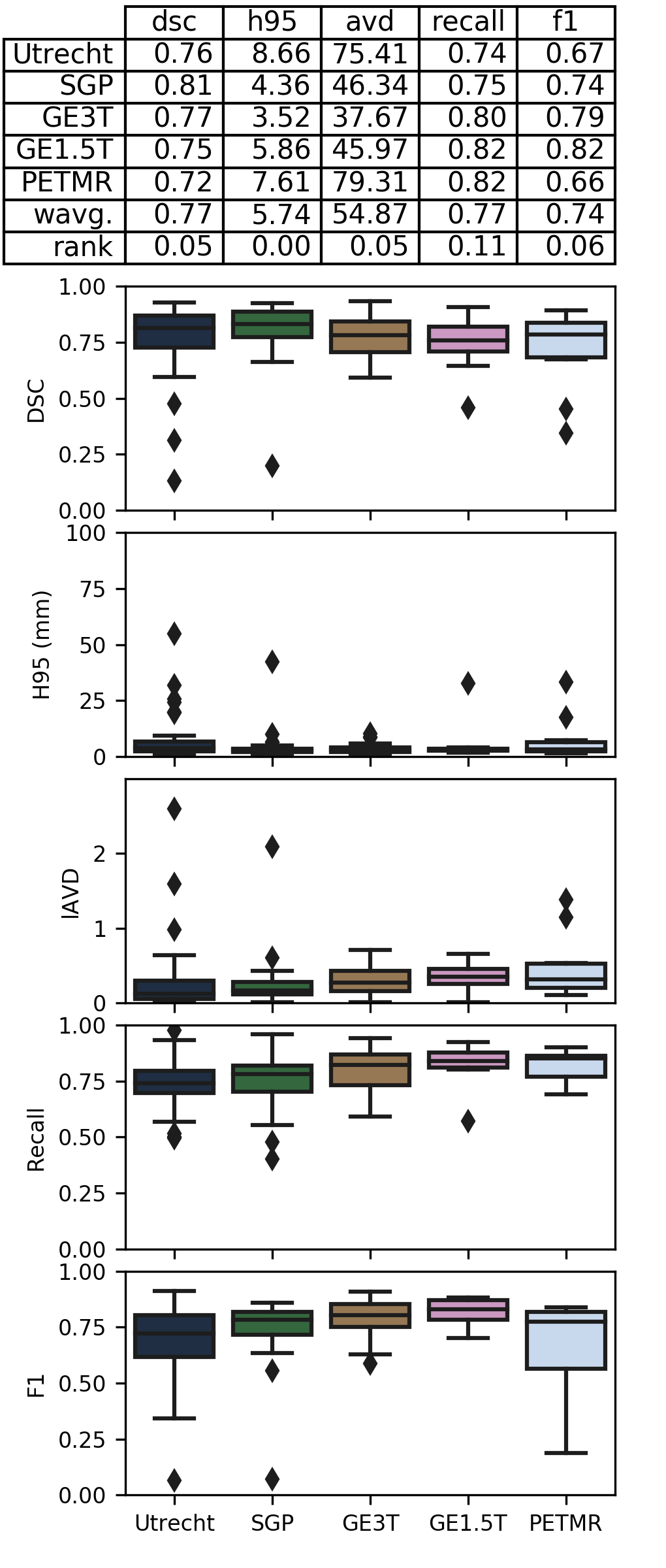}%
\includegraphics[height=19cm]{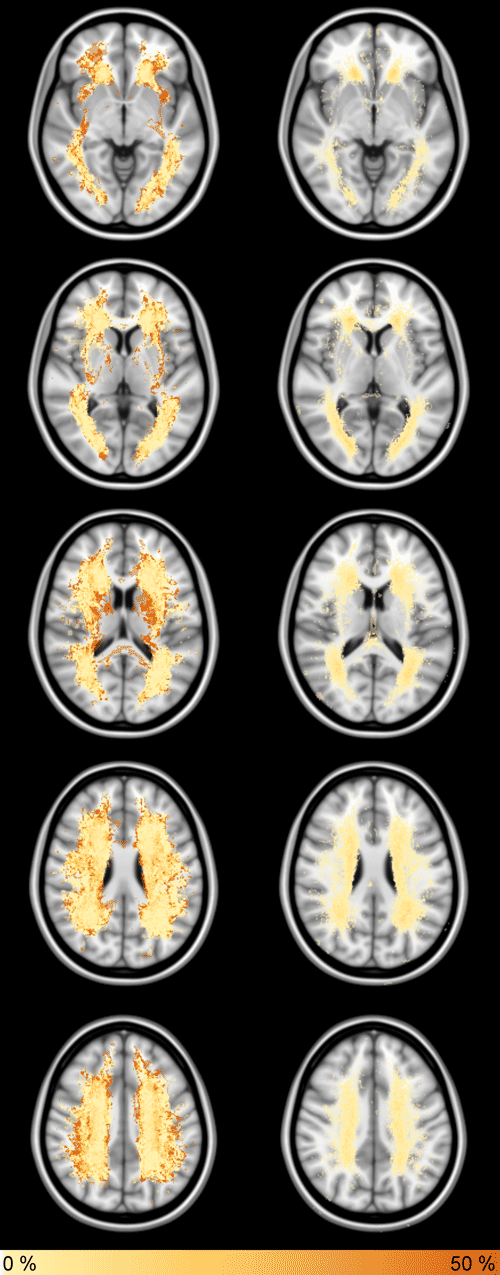}%
\caption{Detailed results of STAPLE applied on all methods. The two columns on the right show the false negative rate (left) and false positive rate (right).}%
\label{app:staple}%
\end{figure*}

\end{document}